\documentclass[default,iicol]{sn-jnl}

\jyear{2022}%

\theoremstyle{thmstyleone}%
\theoremstyle{thmstyletwo}%

\theoremstyle{thmstylethree}%
\usepackage{overpic}
\usepackage[none]{hyphenat}
\newcommand{\figref}[1]{Fig.~\ref{#1}}
\newcommand{\tabref}[1]{Table~\ref{#1}}
\newcommand{\secref}[1]{Sec.~\ref{#1}}
\newcommand{\Rev}[1]{\textcolor{black}{#1}}

\def\ie{\emph{i.e.}}
\def\eg{\emph{e.g.}}
\def\etc{\emph{etc}}
\def\etal{{\em et al.}}
\def\ourmodel{FSGAN}
\def\ourdataset{FS2K}
\newcommand\blfootnote[1]{%
  \begingroup
  \renewcommand\thefootnote{}\footnote{#1}%
  \addtocounter{footnote}{-1}%
  \endgroup
}

\newcommand\datasetNum{2,104}
\newcommand\datasetNumTrain{1,058}
\newcommand\datasetNumTest{1,046}
\newcommand\datasetNumSrcOne{1,529}
\newcommand\datasetNumSrcTwo{98}
\newcommand\datasetNumSrcThree{477}

\begin{document}

\title[Facial-Sketch Synthesis]{\textbf{\Rev{Facial-Sketch} Synthesis: A New Challenge}}


\author[1]{\fnm{Deng-Ping} \sur{Fan}}

\author[2]{\fnm{Ziling} \sur{Huang}$^\dagger$}

\author[3]{\fnm{Peng} \sur{Zheng}$^\dagger$}

\author[4]{\fnm{Hong} \sur{Liu}$^*$}

\author[3]{\fnm{Xuebin} \sur{Qin}$^*$}

\author[1]{\fnm{Luc Van} \sur{Gool}}


\affil[1]{\orgdiv{Computer Vision Lab}, \orgname{ETH Zürich}, \orgaddress{\city{Zürich},  \country{Switzerland}}}

\affil[2]{\orgdiv{Information and Communication Engineering}, \orgname{University of Tokyo}, \orgaddress{\city{Tokyo},  \country{Japan}}}

\affil[3]{\orgdiv{Computer Vision}, \orgname{MBZUAI}, \orgaddress{\city{Abu Dhabi},  \country{UAE}}}

\affil[4]{\orgdiv{Digital Content and Media Sciences Research Division}, \orgname{NII}, \orgaddress{\city{Tokyo},  \country{Japan}}}


\abstract{
This paper aims to conduct a comprehensive study on facial-sketch synthesis (FSS). 
However, due to the high costs of obtaining hand-drawn sketch datasets, there is a lack of a complete benchmark for assessing the development of FSS algorithms over the last decade. 
We first introduce a high-quality dataset for FSS, named \textbf{\ourdataset}, which consists of \datasetNum~image-sketch pairs spanning three types of sketch styles, image backgrounds, lighting conditions, skin colors, and facial attributes. 
\ourdataset~differs from previous FSS datasets in difficulty, diversity, and scalability and should thus facilitate the progress of FSS research. 
Second, we present the largest-scale FSS \Rev{investigation} by reviewing \Rev{$89$} classic methods, including \Rev{$25$} handcrafted feature-based facial-sketch synthesis approaches, \Rev{$29$} general translation methods, and $35$ image-to-sketch approaches. 
In addition, we elaborate comprehensive experiments on the existing $19$ cutting-edge models.
Third, we present a simple baseline for FSS, named \textbf{\ourmodel}. With only two straightforward components, \ie, facial-aware masking and style-vector expansion, our \ourmodel~surpasses the performance of all previous state-of-the-art models on the proposed \ourdataset~dataset
by a large margin. 
Finally, we conclude with lessons learned over the past years and point out several unsolved challenges. 
Our code is available at \url{https://github.com/DengPingFan/FSGAN}.
}




\maketitle

\section{Introduction}
Facial-sketch synthesis (FSS)\blfootnote{$\dagger$ Contributed equally. $*$ Corresponding authors.} aims to generate grayscale sketches from RGB images of human faces (image-to-sketch, I2S) or the other way around (sketch-to-image, S2I)~\cite{wang2008face,yi2019apdrawinggan}. FSS is commonly used by law enforcement or used in surveillance to assist in face recognition and retrieval, based on a sketch drawing from an eyewitness~\cite{wang2008face}. 
Entertainment is also used in mobile apps, such as TikTok and Facebook. In addition, it is an attractive topic in digital entertainment~\cite{koshimizu1999kansei}. Research into FSS has achieved significant progress over the past decade. 

Different from other face-related datasets, such as those for face recognition~\cite{kumar2009attribute,du2015robust,lu2008novel}, face detection~\cite{jain2010fddb}, face key-points detection~\cite{zhang2014facial}, face alignment~\cite{bulat2017far}, and face synthesis~\cite{sun2021multi}, 
which can be manually labelled by annotators with limited training, 
face sketch datasets are much more difficult to obtain because only professional artists can produce high-quality \Rev{references}. 
\begin{table*}[t!]
  \footnotesize
  \renewcommand{\arraystretch}{1.0}
  \renewcommand{\tabcolsep}{2.1mm}
  \caption{ 
  \Rev{Comparison with other FSS datasets.}
  }\label{tab:FSS-Dataset}
  \begin{tabular}{r|ccrrrrrrc}
  \hline
    \textbf{Dataset} & Year & Pub. & Total & Train & Test & Att. & Public & Paired & Resolution \\
   \hline
   CUFS~\cite{wang2008face}  &  2009 & TPAMI & 606   &  306 & 300 & $\times$ & $\surd$ & $\surd$ & \Rev{200 $\times$ 250}\\
   IIIT-Delhi~\cite{bhatt2010matching} & 2010 & BTAS & 231 & 58 & 173 &$\times$ & $\times$ &$\surd$ & \Rev{-} \\
   CUFSF~\cite{zhang2011coupled} & 2011 & CVPR & 1,194 &  500 & 694 & $\times$ & $\surd$ & $\surd$ & \Rev{779.62$\pm$15.05 $\times$ 812.10$\pm$13.92}\\
   VIPSL~\cite{wang2011face,gao2012face} & 2011 & TCSVT & 1,000 & 100 & 900 
   &$\times$ & $\times$ &$\surd$ & \Rev{-} \\
   DisneyPortrait~\cite{berger2013style} & 2013 & TOG & 672 & - & - & $\times$ & $\times$ &$\surd$ & \Rev{-} \\
   UPDG~\cite{YiLLR20}  &  2020 & CVPR & 952  &  798 & 154 & $\times$ & $\times$  & $\times$ & \Rev{-} \\
   APDrawing~\cite{yi2020line} & 2020 & TPAMI  &140 & 70 & 70 & $\times$ & $\surd$ & $\surd$ & \Rev{512 $\times$ 512} \\
  \hline
  \textbf{\ourdataset~(Ours)}  &  2022 & MIR & \textbf{\datasetNum} & \textbf{\datasetNumTrain} & \textbf{\datasetNumTest} &$\surd$ & $\surd$ & $\surd$ & \Rev{299.74$\pm$95.07 $\times$ 273.56$\pm$38.67}\\ 
  \hline
  \end{tabular}
  \begin{tablenotes}
     \item[*]~\Rev{Att. = Attributes. In \cite{zhang2018multidomain} and \cite{zhu2020knowledge}, CUFS is divided into 268 and 338 images for training and testing. For image resolution, we provide the width and height as $W_{avg} \pm W_{std}$ and $H_{avg} \pm G_{std}$, respectively. $W_{avg}$ and $W_{std}$ denote the mean value and standard deviation, respectively.}
  \end{tablenotes}
\end{table*}
Due to the high costs of obtaining professional sketches, existing image-sketch datasets~\cite{wang2008face,yi2019apdrawinggan,yi2020line} are relatively small with limited diversity. 
This dataset shortage has limited the development, especially for data-hungry deep learning models.

In addition, how to evaluate FSS remains an open question. 
Structural similarity (SSIM)~\cite{wang2004image} is one of the most widely used metrics for evaluating image quality, so it is also typically used to assess the performance of S2I models. 
Nevertheless, the characteristics of facial sketches are very different from RGB-based facial images, which makes it challenging to apply the current evaluation metrics to I2S tasks.
Therefore, a new objective and quantitative metric, which is also highly consistent with human assessment, is needed for benchmarking the FSS task. 

Moreover, due to the lack of \textit{high-quality datasets} and \textit{proper evaluation metrics}, different FSS models (\eg,~\cite{wang2008face,yi2019apdrawinggan}) are usually built and tested on diverse training datasets\footnote{Because they want to learn a different style of sketches.} and with different evaluation methods. 
Hence, it is not easy to provide fair and comprehensive comparisons. 
Furthermore, many cutting-edge transformation models (\eg,  
CycleGAN~\cite{zhu2017unpaired}, 
UNIT~\cite{liu2017unsupervised},
Pix2pixHD~\cite{wang2018pix2pixhd},
SPADE~\cite{park2019semantic},
DSMAP~\cite{chang2020domain},
NICE-GAN~\cite{Chen_2020_CVPR}, and DRIT++~\cite{DRIT_plus})
designed for related image-to-image transfer tasks could potentially be employed in FSS tasks.
However, as mentioned above, these models lack performance evaluations for the FSS task because of the shortage of datasets and evaluation metrics. 
Therefore, thorough comparisons and assessments of FSS-related models on a standard FSS dataset with unified evaluation metrics are long overdue.
To this end, we have introduced and maintained an online paper list (\url{https://github.com/DengPingFan/FaceSketch-Awesome-List}) to track the progress of this fast-developing field. 

\subsection{Contributions}
Our goal is to solve the discussed issues (\ie, limited datasets, metrics, and benchmarks) and further contribute to a new challenge for the FSS community. 
The main contributions are as follows:
\begin{itemize}
\item [1)] \textbf{FSS Dataset.}
We build a new high-quality FSS dataset, termed \textbf{\ourdataset}. It is the largest (see \tabref{tab:FSS-Dataset}) publicly released FSS dataset,\footnote{Establishing an FSS dataset drawn by professional artists is more challenging than other face datasets, \eg, face attribute datasets~\cite{liu2015deep}, which is why the largest existing FSS dataset, \ie, CUFSF~\cite{zhang2011coupled}, has only $\sim$1K images in the past $13$ years. Although \ourdataset~is only $\sim$2 times larger than CUFSF, we still took one year to create such a high-quality dataset.} consisting of \datasetNum \ image-sketch pairs with a wide range of image backgrounds, skin patches, sketch styles, and lighting conditions. In addition, we also provide extra attributes, \eg, \textit{gender}, \textit{smile}, \textit{hair style}, \etc., to enable deep learning models to learn more details. 

\item [2)] \textbf{FSS Review and Benchmark.}
We conduct the largest-scale FSS study, reviewing \Rev{$89$} representative approaches, including \Rev{$25$} methods using handcrafted features, \Rev{$29$ models for the general transfer task}, and $35$ I2S transfer algorithms. 
Based on our \ourdataset, we adopt the SCOOT metric~\cite{fan2019scoot} and conduct a rigorous evaluation of $19$ state-of-the-art models from the perspective of content and style. 

\begin{figure*}[t!]
  \centering
  \includegraphics[width=.98\linewidth]{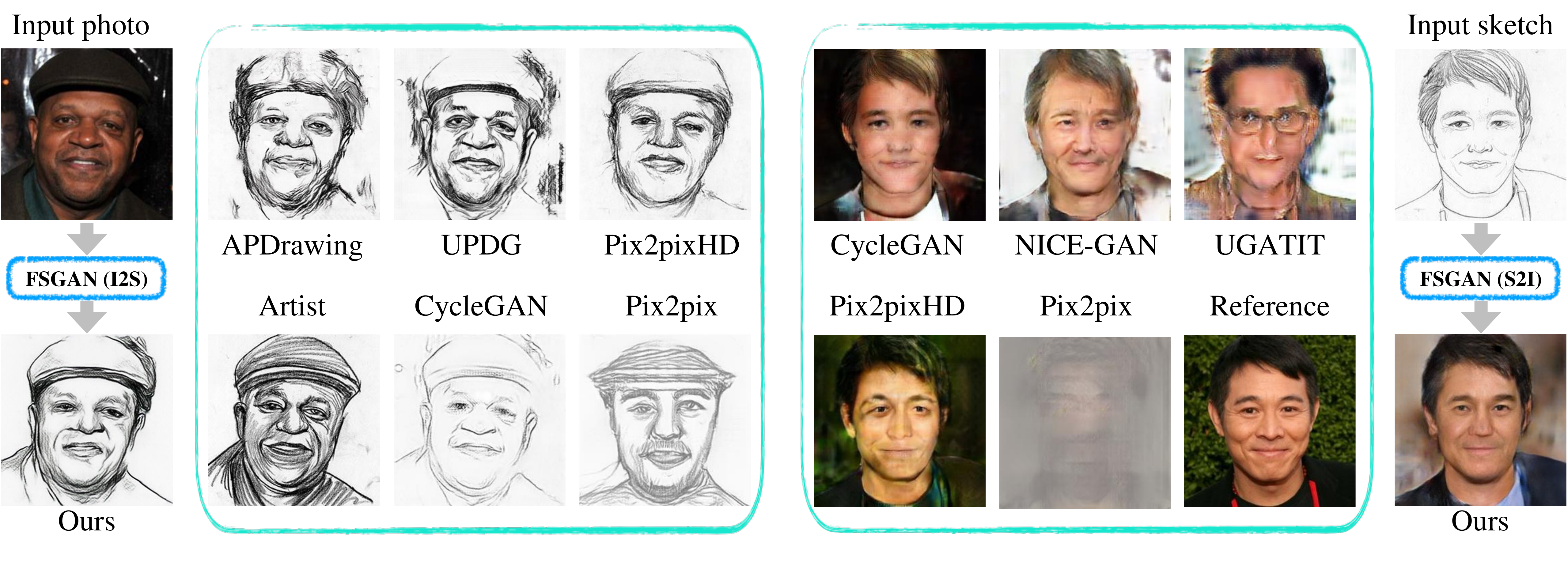}
  \caption{
  \textbf{Left:} Our \ourmodel~(I2S) learns from artist drawings and intelligently turns an input photo into a vivid face sketch. In contrast, the five cutting-edge style transfer approaches cannot obtain visually appealing results. Only UPDG~\cite{YiLLR20} and Pix2pixHD~\cite{wang2018pix2pixhd} perform relatively well, but they generate worse content and style than FSGAN.
  \textbf{Right:} Given a sketch, our \ourmodel~(S2I) can also transform the input into a vivid facial photo. Meanwhile, the results from the five representative deep learning models are either structurally damaged (\ie, CycleGAN~\cite{zhu2017unpaired}, NICE-GAN~\cite{Chen_2020_CVPR}, and UGATIT~\cite{kim2019u}) or blurry (\ie, Pix2pix~\cite{isola2017image}).
  More results can be found in \figref{fig:Bench-1}-\ref{fig:Bench-4}.
  }\label{fig:BaselineResults}
\end{figure*}

\item [3)] \textbf{FSS Baseline.} 
We design an efficient GAN-based baseline, 
termed \textbf{\ourmodel}, which consists of two simple core components, \ie, facial-aware masking and style-vector expansion. The former is utilized to restore details of the facial components, while the latter is adopted to learn different face styles.
\ourmodel~serves as a unified baseline model for both I2S and S2I tasks \Rev{(\figref{fig:BaselineResults})} on our newly built \ourdataset~dataset.
Our \Rev{project} is available at \url{https://github.com/DengPingFan/FSGAN}.

\item [4)] \textbf{Discussions and Future Directions.}
In addition to an overall performance assessment, we also conduct an attribute-level evaluation, present detailed discussions, and explore some promising future directions.
\end{itemize}

\section{Related Works}\label{sec:RelatedWorks}
\begin{figure*}[t!]
  \centering
  \includegraphics[width=.97\linewidth]{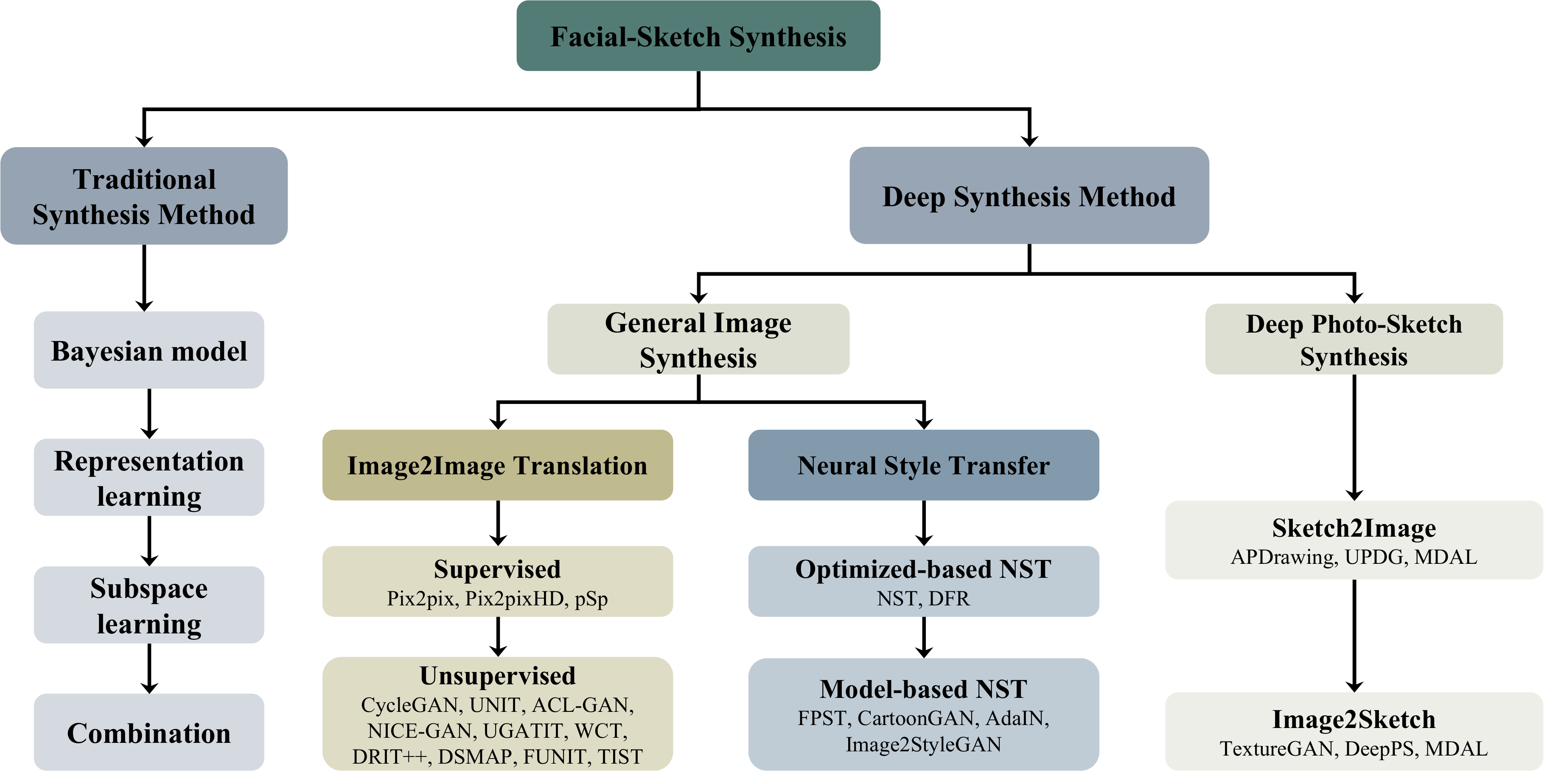}
  \caption{
   \Rev{A taxonomy of facial-sketch synthesis and the representative methods.}
  }\label{fig:relate_work}
\end{figure*}

This section first conducts a complete literature review of the existing FSS datasets. Then, in the second part, we discuss the taxonomy of facial-sketch synthesis and highlight particularly innovative and successful approaches for this task, including traditional facial synthesis, \Rev{image-to-image translation, neural style transfer, and deep photo-sketch synthesis.} The taxonomy of facial-sketch synthesis is shown in  \figref{fig:relate_work}.
A summary of the models, including their key innovations, datasets, code links, and citation information, can be found in \tabref{tab:related_work} and \tabref{tab:related_work3}.

\subsection{Dataset}
We outline four classical datasets for the FSS task, \ie, CUFS \cite{wang2008face}, \Rev{IIIT-D \cite{bhatt2010matching}, CUFSF \cite{zhang2011coupled}, VIPSL \cite{gao2012face}}, and three portrait sketching datasets \cite{berger2013style,YiLLR20,yi2020line}, which are the basis for building most FSS models \cite{peng2018face}. 

\textbf{CUFS}~\cite{wang2008face} is one of the earliest and most commonly used datasets. It contains $606$ photo-sketch pairs, which include $123$ samples from the AR face database \cite{martinez1998ar}, $188$ samples from the CUHK student database, and $295$ samples from the XM2VTS database \cite{messer1999xm2vtsdb}. 
A sketch drawn by an artist and a corresponding photo are provided for each sample. Each photo is taken in a frontal pose under normal lighting conditions and maintains a neutral expression. 
All three sub-databases use solid backgrounds, \eg, cyan, white, blue, \etc.
However, real-world scenes are complex and diverse, and it is difficult to guarantee that photos will be captured in such a fixed environment. 
Besides, the sketches in this dataset were created by the same artist, so they are of limited style.

\textbf{CUFSF} \cite{zhang2011coupled} is a commonly used database for assessing the performance of FSS models. 
It contains $1,194$ photo-sketch pairs, collected from the FERET database \cite{phillips2000feret}. An artist drew all sketches after viewing the corresponding photo. CUFSF has a similar photo collection environment to CUFS but is more challenging. Because the photos in the dataset undergo illumination changes, each face has low contrast with the background, and each sketch contains exaggerated shapes.

\begin{figure*}[t!]
  \centering
  \includegraphics[width=.95\linewidth]{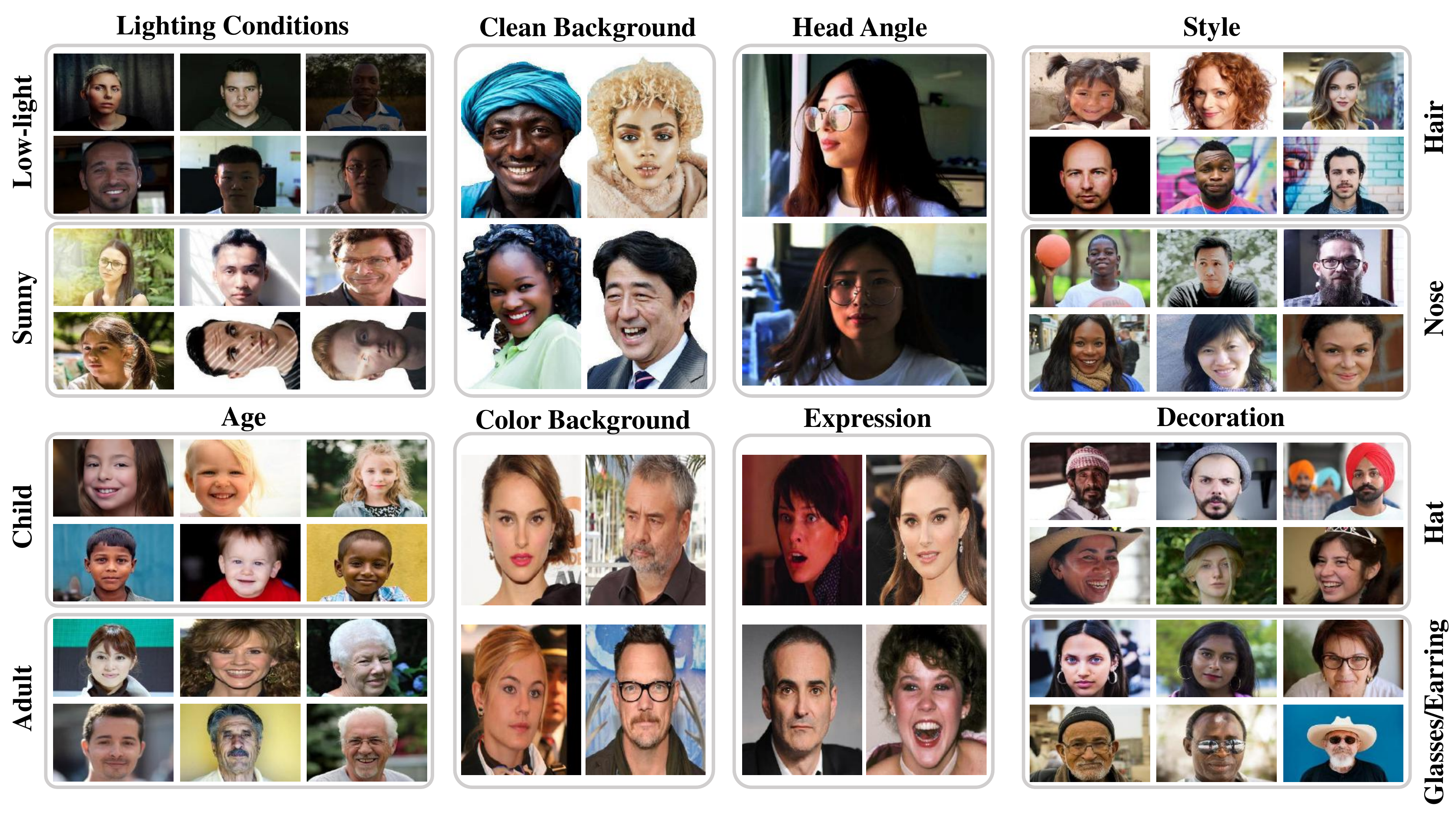}
  \vspace{-5pt}
  \caption{Representative image samples from our \ourdataset. The collected images depict diverse scenes according to different selection criteria, such as various lighting conditions (\ie, low-light, sunny), ages (\ie, child or adult), backgrounds (\ie, clean or colored), head angles, facial expressions (\eg, serious, smiling, and laughing), hair styles (\eg, black, blonde, long, and short), and accessories (\ie, hat or earrings). 
  }\label{fig:RepresentativeExample}
\end{figure*}

\textbf{VIPSL} \cite{gao2012face} contains $200$ face photos collected from the FRAV2D \cite{serrano2007influence}, FERET \cite{phillips2000feret}, and Indian face databases \cite{gao2012face}.
Unlike CUFS and CUFSF, VIPSL has five sketches for each face, drawn by five artists with different styles, while viewing the same photo under the same conditions as CUFS.

\textbf{IIIT-Delhi} \cite{bhatt2010matching,bhatt2012memetically} consists of three
types of sketch databases, including a viewed sketch database, a semi-forensic sketch database, and a forensic sketch database. 
All photos are derived from the CUHK student database and IIIT-Delhi Sketch database \cite{bhatt2010matching}.
The first viewed sketch database contains $238$ sketch-digital image pairs, with all sketches drawn by the professional artist based on a given photo.
The second sub-database has $140$ sketch-face image pairs, where all the sketches are drawn by memory after the artist has observed the corresponding photo.
The third forensic sketch database consists of $190$ sketches that a sketch artist draws according to the description of an eyewitness based on their recollection of a crime scene.
IIIT-Delhi contains multiple styles of sketch portraits, making it more challenging. 
However, obtaining forensic sketches is tricky since they are usually derived from law enforcement.

\textbf{Portrait Sketching Dataset.} Yi~\etal~\cite{yi2020line,YiLLR20} provided two datasets that simulate artistic portrait drawing (APDrawing). 
The first dataset \cite{yi2020line} contains $140$ pairs of face photos and corresponding sketch portraits drawn by a single portrait artist.
This was later extended to a larger dataset in \cite{YiLLR20}, with $952$ face photos and $625$  portrait sketches.
Of the collected photos, $220$ are from three famous painters, and the remaining $212$ photos are from a photography website.\footnote{\url{https://vectorportal.com/}}
It is worth noting that the photos and portraits in this dataset are not paired.
Disney Research published a portrait dataset \cite{berger2013style} composed of $24$ faces from the face database \cite{minear2004lifespan} and $672$ sketches 
from seven artists under four levels of abstraction.
Besides, they also provided each stroke as a transparent bitmap to be used later to create new sketches.

Unlike existing datasets, we provide a more challenging, high-quality, and attribute-annotated dataset, which is currently the largest FSS dataset. 
The new dataset contains $\datasetNum$ pairs of photos and sketches, $\datasetNumTrain$ used for model training, and the remaining for evaluation. 
The strengths of our \ourdataset~include multiple drawing styles, highly accurate alignment between sketches and photos, multiple attribute information, complex backgrounds, \etc. 
Detailed comparisons of the datasets are shown in \tabref{tab:FSS-Dataset}. 

\subsection{Traditional Facial Synthesis}
Researchers have used heuristic image transformations to interactively or automatically synthesize facial sketches~\cite{nishino1999linguistic,iwashita1999expressive,li1997extraction,tominaga1997facial,koshimizu1999kansei,brennan1982caricature} in the early years.
However, these methods tend to generate artificial and inexpressive sketches that lack artistic style.
Therefore, in recent years, more attention has been focused on learning-based facial synthesis schemes, whose taxonomy is shown in \figref{fig:relate_work}.
These can be categorized into Bayesian inference models, representation learning models, subspace learning models, \etc.

\subsubsection{Bayesian Inference Models}
Bayesian inference exploits evidence to update the states of the sketch components over probability models, which has been widely used in FSS \cite{Wang2013IJCV}. 
In \cite{chen2001example}, Chen \etal~first introduced an example-based facial-sketch synthesis system that uses a non-parametric sampling algorithm to learn subtle sketch styles. 
Later, the embedded hidden Markov model \cite{nefian1999face} was used to model the non-linear relationships in photo-sketch pairs, followed by a selective ensemble strategy to generate facial sketches \cite{gao2008face}.
Wang and Tang \cite{wang2008face} followed a similar idea but considered face structures across different scales, using a multi-scale Markov Random Field (MRF) to build the relationships between photo-sketch pairs. 
Xu \etal~\cite{xu2008hierarchical} proposed a hierarchical compositional model that considers the regularity and structural variation of faces.
These methods have made significant progress in generating sketches, but they only consider simple controlled conditions, ignoring variations in lighting and pose. 
Zhang \etal~\cite{zhang2010lighting} addressed this issue by simultaneously considering patch matching, intensity compatibility, gradient compatibility, and shape priors, resulting in better visual effects.
However, MRF-based models have two main drawbacks: (1) they struggle to synthesize unseen facial information and (2) their optimization is NP-hard.
Zhou \etal~\cite{zhou2012markov} used Markov weight fields and cascaded decomposition to build a robust facial synthesis system, using a linear combination of candidate patches to approximate new sketch patches. 
Wang \etal~\cite{wang2013learnable} built a non-parametric model to transform a photograph into a portrait painting, where an MRF is used to enhance the spatial coherence of the style parameters, and an active shape model and a graph-cut model are used to learn the local information of facial features. 
Wang \etal~\cite{wang2013transductive} presented a transductive learning method to synthesize facial sketches, which employs an on-the-fly optimization process to minimize the loss of the given test samples. 
Peng \etal~\cite{peng2015superpixel} designed a superpixel method built on the Markov model to improve the flexibility without dividing the photo into regular rectangular patches.
Then, they not only used the Markov network to model the relationships between image patches but also retained many visual aspects of the cues (such as edges) through multiple visual features~\cite{peng2015multiple}.

\subsubsection{Subspace Learning Models}
Subspace learning has been widely studied in the FSS task \cite{Wang2013IJCV}, which learns a low dimensional manifold space embedded in a high dimensional space \cite{abdi2010principal}. 
Tang and Wang \cite{tang2002face,tang2003face,tang2004face} proposed a series of example-based approaches based on the linear eigen-transformation method.
These methods are global linear systems, and they cannot fully explain the relationships between photo-sketch pairs because such a transformation is not a simple linear relationship.
Liu \etal~\cite{liu2005nonlinear} used the LLE to handle this problem, making photo and sketch patches have manifolds with similar local geometric shapes in two different image spaces.
However, pseudo-image generation and representation learning are divided into two independent processes, leading to sub-optimal results.
Huang and Wang \cite{huang2013coupled} proposed a joint learning framework, which contains domain-specific dictionary learning and subspace learning.

\subsubsection{Representation Learning Models}
Sparse coding and dictionary learning, \emph{a.k.a.} representation learning, are used for the FSS 
task \cite{Wang2013IJCV}. 
Ji \etal~\cite{ji2011local} demonstrated that personalized features are not effectively captured through the synthesis process. As such, several works \cite{ji2011local,chang2011face,zhang2011face} use different regression models, such as k-NN \cite{ji2011local}, Lasso \cite{ji2011local}, multivariate output regression \cite{chang2011face}, and support vector regression \cite{zhang2011face}, to build the transformation between photos and sketches.
To improve the quality of the generated facial sketches, Wang \etal~\cite{wang2011face,gao2012face} used local linear embedding (LLE) \cite{roweis2000nonlinear} to estimate an initial sketch or photo and then introduced a sparse multi-dictionary representation model that can focus on high-frequency and detailed information.
However, most representation-based models assume that the same representations are shared by the source input and the target output, limiting a particular style's local structures in the synthesis process.
To relax this constraint, Wang \etal~\cite{wang2012semi} introduced a semi-coupled dictionary learning method, in which a linear transformation is used to bridge the gap between two different domain-specific representations. 
Gao \etal~\cite{gao2012face} also took a two-step algorithm \cite{zhang2011face} into consideration, presenting a selection scheme to generate the initial pseudo-images and introducing a sparse-representation-based enhancement (SRE) to synthesize sketches.

\subsubsection{Combination Models}
Recently, some works have explored combination models, which combine different machine learning models, \eg, combing Bayesian inference and subspace learning methods. 
Berger \etal~\cite{berger2013style} proposed a model to simulate the styles of the different artists and the process of abstraction, which can be used for facial-sketch synthesis.
Song \etal~\cite{song2014real} introduced a real-time FSS method, which first uses a k-NN algorithm to find the top-k similar local patches. Then a linear combination is used to compute the corresponding sketch image and image denoising technology is adopt to enhance the visual quality.
However, the model~\cite{song2014real} is still time-consuming due to the k-NN process, so Wang \etal~\cite{wang2018random} addressed this problem by replacing offline random sampling with an online scheme that is further combined with a recognition weight representation.
Most existing traditional methods are entirely dependent on the scale of the training data, so Zhang \etal~\cite{zhang2015robust} presented a robust model trained on a template stylistic sketch. The model includes representation learning, MRF, and a cascaded model.
Li \etal~\cite{li2017free} proposed a free-hand sketch synthesis method, combining a perceptual grouping model with a deformable stroke model. 
The work in \cite{li2017adaptive} introduces an adaptive learning method that combines representation learning and a Markov network.
Men \etal~\cite{men2018common} proposed a common framework for interactive texture transfer with structure guidance. Their model implements the synthesis process dynamically using multiple channels, including structure extraction, structure propagation, and guided texture transfer. 

\subsection{General Image Synthesis}
\Rev{Deep facial-sketch synthesis belongs to the task of image generalization. Therefore, general image synthesis methods, such as image-to-image translation and neural style transfer, can also be used to generate facial sketches. We will overview various cutting-edge transformation models.}

\subsubsection{Image-to-Image Translation}
Image-to-image translation (I2I) \cite{saxena2021comparison} is a hot topic in computer vision and machine learning. The goal is to transform the input image from a source domain to a different target domain while retaining the intrinsic source content and transferring the extrinsic target style.
\Rev{Current I2I models are typically built on a generative adversarial network (GAN)~\cite{goodfellow2014generative}. They can be generally categorized into supervised and unsupervised I2Is.} 

\textbf{Supervised I2I.} 
\Rev{Supervised I2I uses aligned image pairs as the source and target domains to learn a transformation model that can convert the source image into the target image.
One representative I2I method is Pix2pix \cite{isola2017image}, which applies a conditional GAN (cGAN) \cite{mirza2014conditional} to the task.
The main difference from the original cGAN is that the generator in Pix2pix is a U-Net \cite{ronneberger2015u}. 
However, Wang \etal~\cite{wang2018pix2pixhd} observed that the adversarial training in Pix2pix is unstable, preventing the model from generating high-resolution images. 
Therefore, they extended the original Pix2pix with a new feature matching loss, which can generate high-resolution images of size $2048 \times 1024$.
Zhu \etal~\cite{zhu2017toward} proposed the BicycleGAN, which includes a conditional VAE and a conditional latent regression GAN,  to resolve the collapse problem and achieve improved performance.
Furthermore, to reduce the loss of semantic information in the Pix2pixHD model \cite{wang2018pix2pixhd}, Park \etal~\cite{park2019semantic} introduced a SPADE-based generator, which adds spatially-adaptive normalization into the generator of Pix2pixHD so as to enhance the semantic information throughout the network.}

\textbf{Unsupervised I2I.}
\Rev{Collecting paired data is not practical because it is labor-intensive.
Therefore, several unsupervised I2I models have been proposed to train two different generative networks under the constraint of a cycle-consistency loss. If we convert a zebra image to a horse image and then back to a zebra image, we should get the same input image back.
Examples include CycleGAN \cite{zhu2017unpaired}, DiscoGAN \cite{kim2017learning}, and DualGAN \cite{yi2017dualgan}.
Later, Liu \etal~\cite{liu2017unsupervised} proposed an unsupervised I2I model (UNIT), in which the same latent code in a shared latent feature space can represent image pairs in different domains.
Kim \etal~\cite{kim2019u} later proposed a novel attention module with a new normalization function, which they integrated into a GAN model to supervise texture and shape variations flexibly. 
By rethinking the standard GAN model, Chen \etal~\cite{Chen_2020_CVPR} proposed a NICE-GAN with the key idea of coupling discriminators and encoders, \ie, reusing the discriminator parameters for encoding the input.
Zhao \etal~\cite{zhao2020unpaired} proposed ACL-GAN, which utilizes a new adversarial consistency loss instead of a cyclic loss to emphasize the commonality between the source and target domains.
To improve the content representation ability, Chang \etal~\cite{chang2020domain} proposed DSMAP to leverage the relationship between content and style. 
Specifically, the model maps content features from a shared domain-invariance feature space into two separate domain-specific features.
Furthermore, DRIT++ \cite{DRIT_plus} uses two image generators, two content encoders, a content discriminator, two attribute encoders, and two domain discriminators to embed an image into a domain-invariant content space and a domain-specific attribute space. 
Besides, Jiang \etal~\cite{jiang2020tsit} proposed two-stream I2I translation (TSIT) to learn both semantic structural features and stylistic features and then fuse the feature maps of the content and style in a coarse-to-fine manner.
More recently, Zhang \etal~\cite{zhang2020cross} proposed a CoCosNet for exemplar-based image translation, which contains two sub-networks. The first embeds the inputs from different domains into a feature domain that depends on the semantic correspondence. Meanwhile, the second uses a series of denormalization blocks to progressively synthesize the target images.
Zhou \etal~further extended CoCosNet with full-resolution semantic correspondence learning \cite{zhou2021cocosnet}, with the main difference being the use of a regular and GRU-based propagation applied iteratively at each semantic level. 
More recently, Chen \etal~\cite{sofgan} proposed a SofGAN, which decouples the portrait feature into a geometric feature and a texture feature. These two features are then fed into two network branches. 
The first branch is a hyper network to decode the geometric feature into the weight of the SOF net that represents the semantic occupancy field (SOF) among 3D space. Then, a segmentation map is rendered via a ray-casting-marching scheme using the output features of the SOF net.
The second branch is a texture transformation of each semantic region using a GAN generator with a style code sampled from the texture space.
Finally, a novel Semantic Instance Wise (SIW) StyleGAN module is used to stylize the generated segmaps and output a photorealistic portrait regionally.
}

\subsubsection{Neural Style Transfer}
Neural style transfer (NST), which aims at generating visually appealing images via neural networks, has been introduced into the FSS task \cite{jing2019neural}.
Specifically, NST is used to render a content image in different styles. NST methods can be categorized into optimization-based methods and model-based methods.\footnote{Note that some related works belong to the general GAN-based model, such as CartoonGAN \cite{chen2018cartoongan} and pSp \cite{richardson2020encoding}. These GAN models can be used for either neural style transfer or image-to-image translation. Since we do not make a specific review of the generalized GAN model, we classified a few GAN models into the neural style transfer task as a quick overview of these methods.}

\textbf{Optimization-based methods.} 
\Rev{The online NST algorithm iteratively updates a given input image to match the desired CNN features, including the photo’s content and artistic style information.
Gatys \etal~\cite{gatys2015neural,gatys2016image} made the first contribution to this field, using a classical CNN (\ie, VGG \cite{simonyan2015very}) to render an image with famous painting styles. 
Besides, StyleGAN \cite{karras2019style} uses a latent space to maintain consistent results for image synthesis. However, it is challenging to achieve promising results under the given conditions.
Recently, Abdal \etal~\cite{abdal2019image2stylegan} integrated the classical NST \cite{gatys2015neural,gatys2016image} into the StyleGAN model, using NST to project the input image into the latent space defined in StyleGAN.
Then, Kotovenko \etal~\cite{kotovenko2021rethinking} further enhanced the classical NST \cite{gatys2015neural,gatys2016image} by optimizing parameterized brushstrokes, which is built on a simple differentiable rendering mechanism.}

\textbf{Model-based methods.}  
\Rev{Optimization-based online methods achieve satisfactory results, but there are still some limitations. 
One major drawback is the slow computational speed and high cost of online iterative optimization. 
To address this issue, several works introduce a feed-forward network to mimic the optimization objective of style transfer \cite{jing2019neural}.}

\Rev{End-to-end models can be divided into those that design a basic deep neural architecture and those that introduce a new loss function.
For basic architectures, Johnson \etal~\cite{johnson2016perceptual} took advantage of the benefits of the neural network and optimization-based NST model and proposed a method for training a feed-forward network using a new perceptual loss. 
TextureNet \cite{ulyanov2016texture} follows a similar idea but with different neural network architecture.
Both \cite{johnson2016perceptual} and \cite{ulyanov2016texture} are real-time style transfer methods.
Chen and Schmidt \cite{chen2016fast} introduced a style swap operation to exchange the patches with visual context and those with style,  further formulating a new optimization objective that aims to learn an inverse neural network for arbitrary style transfer.
In terms of methods based on the loss function, CartoonGAN \cite{chen2018cartoongan} was presented to transfer real-world photos into cartoon-style images. It consists of two novel loss functions designed to preserve clear edge information and cope with the stylistic difference between photos and cartoons.}

\begin{table*}[t!]
  \tiny
  \renewcommand{\arraystretch}{1.0}
  \renewcommand{\tabcolsep}{0.03mm}
  \caption{
  Summary of popular related works. These can be categorized into \Rev{three} types: \textit{Traditional Facial Synthesis}, \textit{General Image Synthesis}, and \textit{Deep Image-to-Sketch Synthesis}.
  }
  \label{tab:related_work}
  \begin{tabular}{|c|r|c|c|c|c|c|c|r|}
  \toprule
  \#  & Model  & Publ.   & Year  & Code    & Components  & Dataset  & Assist. & Cite. \\\midrule
  \multicolumn{9}{c}{Traditional Facial Synthesis}\\\midrule
  1& \href{EFSGNS}{https://ieeexplore.ieee.org/stamp/stamp.jsp?arnumber=937657}~\cite{chen2001example} & ICCV   & 2001 & -  & Active Shape Model, Non-parametric Sampling  & E  & - & \href{160}{https://scholar.google.com/scholar?cites=8041449582823287565&as_sdt=2005&sciodt=0,5&hl=en}
  \\
  2& \href{Nonlinear}{https://ieeexplore.ieee.org/stamp/stamp.jsp?arnumber=1467376}~\cite{liu2005nonlinear} & CVPR   & 2005  & -  &Local Linear Preserving at patch levels & Y  & - & \href{398}{https://scholar.google.com/scholar?cites=353123226030332105&as_sdt=2005&sciodt=0,5&hl=en} 
  \\
  3& \href{E-HMM}{https://ieeexplore.ieee.org/stamp/stamp.jsp?tp=&arnumber=4453838}~\cite{gao2008face} & TCSVT   & 2008 & -  &Embedded Hidden Markov Model, Selective Ensemble  & Y  & - & \href{165}{https://scholar.google.com/scholar?cites=5355877645583956851&as_sdt=2005&sciodt=0,5&hl=en} 
  \\
  4& \href{HCM}{https://ieeexplore.ieee.org/stamp/stamp.jsp?arnumber=4468712}~\cite{xu2008hierarchical} & PAMI   & 2008  & -  & Graph, Minimum Description Length & C, D, BW, E   & - & \href{93}{https://scholar.google.com/scholar?cites=14427331241847078873&as_sdt=2005&sciodt=0,5&hl=en} 
  \\
  5& \href{MRF}{https://ieeexplore.ieee.org/stamp/stamp.jsp?arnumber=4624272}~\cite{wang2008face} & PAMI   & 2009  & \href{Code}{https://github.com/ClaireXie/face2sketch} & Multi-scale Markov Random Fields  & Y  & - & \href{872}{https://scholar.google.com/scholar?cites=15538075060384594671&as_sdt=2005&sciodt=0,5&hl=en} 
  \\
  6& \href{LPR}{https://link.springer.com/content/pdf/10.1007/978-3-642-15567-3_31.pdf}~\cite{zhang2010lighting} & ECCV   & 2010  & -  & Local Evidence Function, Patch Matching, Shape Prior, MRF & Y  & - & \href{120}{https://scholar.google.com/scholar?cites=15659024496359119515&as_sdt=2005&sciodt=0,5&hl=en} 
  \\
  7& \href{LRM}{https://ieeexplore.ieee.org/stamp/stamp.jsp?arnumber=6005598}~\cite{ji2011local} & ICIG   & 2011  & -  &Local Regression, kNN & Y  & - & \href{19}{https://scholar.google.com/scholar?cites=377831356568380590&as_sdt=2005&sciodt=0,5&hl=en} 
  \\
  8& \href{MOR}{https://link.springer.com/content/pdf/10.1007/978-3-642-21602-2_60.pdf}~\cite{chang2011face} & HCII   & 2011  & -  & Multivariate Output Regression  & Y  & - & \href{22}{https://scholar.google.com/scholar?cites=4197252680864291982&as_sdt=2005&sciodt=0,5&hl=en} 
  \\
  9& \href{MDSR}{https://ieeexplore.ieee.org/stamp/stamp.jsp?arnumber=6005537&casa_token=zXeSJgZIGMwAAAAA:b5wfFaY-amoSK4ky3CjP0tracOTblcT6nyR1yhu-11q5JlEDc2PDmlc185KeE0WR1NpKTC-Na0Y}~\cite{wang2011face} & ICIG   & 2011  & -  & LLE, Dictionary Learning, Sparse Representation & Y, BX  & - & \href{55}{https://scholar.google.com/scholar?cites=7825893782012456874&as_sdt=2005&sciodt=0,5&hl=en}
  \\ 
  10& \href{SVR}{https://ieeexplore.ieee.org/stamp/stamp.jsp?arnumber=6115625&casa_token=0sRb_iz3R1cAAAAA:vPM_MXtB-XAtFaJESvj9iL3GQpUL8MC9Db452m2sDmgkttLYy9i_9RFgh5YwEsUG5G1I6tYZ2Mk}~\cite{zhang2011face} & ICIP   & 2011 & -  &Support Vector Regression & Y, BX  & - & \href{41}{https://scholar.google.com/scholar?cites=74867895239307705&as_sdt=2005&sciodt=0,5&hl=en}
  \\
  11& \href{SCDL}{https://ieeexplore.ieee.org/stamp/stamp.jsp?arnumber=6247930}~\cite{wang2012semi} & CVPR   & 2012  & -  &Sparse Coding, Semi-coupled Dictionary Learning & Y  & - & \href{613}{https://scholar.google.com/scholar?cites=14604596917736816261&as_sdt=2005&sciodt=0,5&hl=en} 
  \\
  12& \href{MWF}{https://ieeexplore.ieee.org/stamp/stamp.jsp?arnumber=6247788}~\cite{zhou2012markov} & CVPR   & 2012  & -  &Markov Weight Fields, Cascade Decomposition  & Y, E  & - & \href{173}{https://scholar.google.com/scholar?cites=9432620588376576790&as_sdt=2005&sciodt=0,5&hl=en} 
  \\
  13& \href{SR}{https://ieeexplore.ieee.org/stamp/stamp.jsp?arnumber=6196209&casa_token=KdRDYYSo_8oAAAAA:1ghHPU8zINMF93h3g60NcCc-GtMYcpb9ZCyQTbRnqr15LVPpDVHMB3wEp2nvxXB5_OuKb5W9faI&tag=1}~\cite{gao2012face} & TCSVT   & 2012  & -  &Sparse Neighbor Selection, Sparse-Representation Enhance  & Y, BX  & - & \href{185}{https://scholar.google.com/scholar?cites=1745189352710028792&as_sdt=2005&sciodt=0,5&hl=en}
  \\
  14& \href{SAPS}{https://dl.acm.org/doi/pdf/10.1145/2461912.2461964}~\cite{berger2013style} & TOG   & 2013  & -  &Edge Detection, Shape Deformation & B  & - & \href{116}{https://scholar.google.com/scholar?cites=3956967327189690251&as_sdt=2005&sciodt=0,5&hl=en} 
  \\
  15& \href{FESM}{http://www.bmva.org/bmvc/2013/Papers/paper0036/paper0036.pdf}~\cite{wang2013learnable} & BMVC   & 2013  & -  &Markov Random Field, Graph-cut  & E  & - & \href{22}{https://scholar.google.com/scholar?cites=4299542421620235268&as_sdt=2005&sciodt=0,5&hl=en} 
  \\
  16& \href{Transductive}{https://ieeexplore.ieee.org/stamp/stamp.jsp?arnumber=6515363&casa_token=UKf4wGwzzF4AAAAA:2v1KEPUy9mNkRrnnqwb1Xg59oXFrb6k2okQu1sCIS_mLcuoo3Rk6GkMvBZOeAh8eJ_jkZnMbYBQ}~\cite{wang2013transductive} & TNNLS   & 2013  & -  &Probabilistic graph model,Transductive Learning  & Y, CU & - & \href{167}{https://scholar.google.com/scholar?cites=7035900011903306800&as_sdt=2005&sciodt=0,5&hl=en} 
  \\
  17& \href{CDFSL}{https://openaccess.thecvf.com/content_iccv_2013/papers/Huang_Coupled_Dictionary_and_2013_ICCV_paper.pdf}~\cite{huang2013coupled} & ICCV   & 2013   & -  &Coupled Dictionary and Feature Space Learning  & Y  & - & \href{177}{https://scholar.google.com/scholar?cites=3027861659369696314&as_sdt=2005&sciodt=0,5&hl=en} 
  \\
  18& \href{REB}{https://link.springer.com/content/pdf/10.1007/978-3-319-10599-4_51.pdf}~\cite{song2014real} & ECCV   & 2014  & \href{Project}{https://ybsong00.github.io/eccv14/index.html}  & kNN, Linear Estimation, Sketch Denoising  & Y, D  & - & \href{124}{https://scholar.google.com/scholar?cites=1750317384921147661&as_sdt=2005&sciodt=0,5&hl=en} 
  \\
  19& \href{RobustStyle}{https://ieeexplore.ieee.org/stamp/stamp.jsp?arnumber=7331298&casa_token=LZUyh-5HaEoAAAAA:HsbMl8S4ANVCtjdbxVFm8SIXkMKYFzl0tlo5aXMmHI07wKSR1sBb9x2dHouBhXyWVB80znJNeQI&tag=1}~\cite{zhang2015robust} & TIP   & 2015 & -  &Sparse Representation, Multi-scale Selection  & Y, E  & - & \href{49}{https://scholar.google.com/scholar?cites=5699532639801537132&as_sdt=2005&sciodt=0,5&hl=en} 
  \\
  20& \href{SPP}{https://ieeexplore.ieee.org/stamp/stamp.jsp?arnumber=7335623}~\cite{peng2015superpixel} & TCSVT   & 2015  & \href{Project}{http://chunleipeng.com/TCSVT2015_SFSPS.html} &Superpixels, Markov Networks  & Y, CU, BY  & - & \href{45}{https://scholar.google.com/scholar?cites=13175172138055396386&as_sdt=2005&sciodt=0,5&hl=en}
  \\
  21& \href{MR}{https://ieeexplore.ieee.org/stamp/stamp.jsp?arnumber=7244234&casa_token=YfhbbBVTuuAAAAAA:J1bNf__dieipZqVMF1QeM1hW66kevNR-MMfQE962iaolK9O3bp7od-HpHgipdcVhU_2peHEfxzc}~\cite{peng2015multiple} & TNNLS   & 2016  & -  &Markov Networks, Edge Enhancement, Alternating Opt. & Y, BY  & - & \href{107}{https://scholar.google.com/scholar?cites=11603015024067356011&as_sdt=2005&sciodt=0,5&hl=en}
  \\
  22& \href{DSM}{https://link.springer.com/article/10.1007/s11263-016-0963-9}~\cite{li2017free} & IJCV   & 2017  & \href{Project}{https://panly099.github.io/skSyn.html} &Perceptual Grouping, Deformable Stroke Model & A, B  & - &  \href{37}{https://scholar.google.com/scholar?cites=3629012578125742096&as_sdt=2005&sciodt=0,5&hl=en} 
  \\
  23& \href{AR}{https://www.sciencedirect.com/science/article/pii/S0925231217310032?casa_token=YjZ9GqyReLkAAAAA:1FagV2Dic2wtTyBeQMZGqFgKom2h4p2SmrLPX6xku8FKhdGNHNNjyZkUiMufeV9VpsSiHY3FmuY}~\cite{li2017adaptive} & NC   & 2017  & -  &Adaptive Representation, Markov Networks  & Y  & - & \href{10}{https://scholar.google.com/scholar?cites=2984190597586859265&as_sdt=2005&sciodt=0,5&hl=en} 
  \\
  24& \href{RS}{https://arxiv.org/pdf/1701.01911.pdf}~\cite{wang2018random} & PR   & 2018& -  & Offline Random Sampling, Locality Constraint  & Y, CU  & - & \href{96}{https://scholar.google.com/scholar?cites=11195808932529361329&as_sdt=2005&sciodt=0,5&hl=en} 
  \\
  25& \href{CFITT}{https://openaccess.thecvf.com/content_cvpr_2018/papers/Men_A_Common_Framework_CVPR_2018_paper.pdf}~\cite{men2018common} &CVPR   & 2018  & \href{Github}{https://github.com/menyifang/CFITT} &PatchMatch, Guided Texture Transfer   & E  & Sm. & \href{19}{https://scholar.google.com/scholar?cites=7628289378450750388&as_sdt=2005&sciodt=0,5&hl=en} 
  \\
  \midrule
  \multicolumn{9}{c}{General Image Synthesis}\\\midrule
  26& \href{NST}{https://openaccess.thecvf.com/content_cvpr_2016/papers/Gatys_Image_Style_Transfer_CVPR_2016_paper.pdf}~\cite{gatys2015neural,gatys2016image} & CVPR   & 2016   & \href{Github}{https://github.com/kaishengtai/neuralart}  & Parametric Texture Mode, Representation Inversion  & E  & - & \href{3853}{https://scholar.google.com/scholar?cites=15430064963552939126&as_sdt=2005&sciodt=0,5&hl=en} 
  \\
  27& \href{FNS}{https://arxiv.org/pdf/1603.08155.pdf\%7C}~\cite{johnson2016perceptual} & ECCV   & 2016 & \href{Github}{https://github.com/jcjohnson/fast-neural-style}  & Image Transformation and Loss Network, Perceptual Loss  & F  & - & \href{7038}{https://scholar.google.com/scholar?cites=5132755018694140583&as_sdt=2005&sciodt=0,5&hl=en} 
  \\ 
  28& \href{TextureNet}{http://proceedings.mlr.press/v48/ulyanov16.pdf}~\cite{ulyanov2016texture} &ICML   & 2016  & \href{Github}{https://github.com/DmitryUlyanov/texture_nets}  & Generator Network, Descriptor Network,   &E  & - & \href{813}{https://scholar.google.com/scholar?cites=5452588382099665760&as_sdt=2005&sciodt=0,5&hl=en} 
  \\
  29& \href{FPST}{https://arxiv.org/abs/1612.04337}~\cite{chen2016fast} &NeurIPSW   & 2016  & \href{Github}{https://github.com/rtqichen/style-swap}  & CNN, Style Swap, Inverse Network  & F, P  & - & \href{285} {https://scholar.google.com/scholar?cites=11584469925299163156&as_sdt=2005&sciodt=0,5&hl=en}
  \\
  30& \href{CIN}{https://arxiv.org/pdf/1610.07629.pdf}~\cite{dumoulin2016learned} &ICLR   & 2017  & \href{Github}{https://github.com/magenta/magenta/tree/master/magenta/models/image_stylization}& Conditional Instance Normalization & G, E  & - & \href{838}{https://scholar.google.com/scholar?cites=7122040962029266183&as_sdt=2005&sciodt=0,5&hl=en} 
  \\
  31& \href{ITN}{https://openaccess.thecvf.com/content_cvpr_2017/papers/Ulyanov_Improved_Texture_Networks_CVPR_2017_paper.pdf}~\cite{ulyanov2017improved} &CVPR   & 2017  & \href{Github}{https://github.com/DmitryUlyanov/texture_nets}&  Instance Normalization, Julesz Generator Network & E  & - & \href{546}{https://scholar.google.com/scholar?cites=15121556168732029899&as_sdt=2005&sciodt=0,5&hl=en} 
  \\
  32& \href{AdaIN}{https://openaccess.thecvf.com/content_ICCV_2017/papers/Huang_Arbitrary_Style_Transfer_ICCV_2017_paper.pdf}~\cite{huang2017arbitrary} &ICCV   & 2017  & \href{Github}{https://github.com/xunhuang1995/AdaIN-style}  &Adaptive Instance Normalization  & F, P  & - & \href{2123}{https://scholar.google.com/scholar?cites=6462913724934880335&as_sdt=2005&sciodt=0,5&hl=en} 
  \\
  33& \href{WCT}{https://arxiv.org/pdf/1705.08086.pdf}~\cite{WCT-NIPS-2017} &NeurIPS   & 2017  & \href{Github}{https://github.com/Yijunmaverick/UniversalStyleTransfer} & Multi-level Stylization, Whitening and Coloring Transforms   & F, L  & - & \href{578}{https://scholar.google.com/scholar?cites=7001062204457348357&as_sdt=2005&sciodt=0,5&hl=en} 
  \\
  34& \href{CartoonGAN}{https://openaccess.thecvf.com/content_cvpr_2018/papers/Chen_CartoonGAN_Generative_Adversarial_CVPR_2018_paper.pdf}~\cite{chen2018cartoongan} & CVPR  &2018  & \href{Github}{https://github.com/znxlwm/pytorch-CartoonGAN}  & GAN, Semantic Content Loss, Edge-promoting Loss &E &- & \href{227}{https://scholar.google.com/scholar?cites=9106316219708164634&as_sdt=2005&sciodt=0,5&hl=en} 
  \\
  35& \href{I2SGAN}{https://openaccess.thecvf.com/content_ICCV_2019/papers/Abdal_Image2StyleGAN_How_to_Embed_Images_Into_the_StyleGAN_Latent_Space_ICCV_2019_paper.pdf}~\cite{abdal2019image2stylegan} &CVPR   & 2019  & \href{Github}{https://github.com/zaidbhat1234/Image2StyleGAN} & StyleGAN, Embedding  & AC, BU  & - & \href{389}{https://scholar.google.com/scholar?cites=10160113996096332863&as_sdt=2005&sciodt=0,5&hl=en} 
  \\
  36& \href{RST}{https://arxiv.org/pdf/2103.17185.pdf}~\cite{kotovenko2021rethinking} &CVPR   & 2021  & \href{Github}{https://github.com/CompVis/brushstroke-parameterized-style-transfer} & Differentiable Renderer, Brushstrokes Parameterization & E  & - & \href{10}{https://scholar.google.com/scholar?cites=12590451920139992694&as_sdt=2005&sciodt=0,5&hl=en} 
  \\
  37& \href{pSp}{https://openaccess.thecvf.com/content/CVPR2021/papers/Richardson_Encoding_in_Style_A_StyleGAN_Encoder_for_Image-to-Image_Translation_CVPR_2021_paper.pdf}~\cite{richardson2020encoding} & CVPR   &2021  & \href{Github}{https://github.com/eladrich/pixel2style2pixel}  & StyleGAN, Disentangled Latent Feature, Map2Style &AC, BU & - & \href{194}{https://scholar.google.com/scholar?cites=12264250297849199750&as_sdt=2005&sciodt=0,5&hl=en} 
  \\
  38& \href{Pix2pix}{https://openaccess.thecvf.com/content_cvpr_2017/papers/Isola_Image-To-Image_Translation_With_CVPR_2017_paper.pdf}~\cite{isola2017image} & CVPR   & 2017 & \href{Github}{https://github.com/junyanz/pytorch-CycleGAN-and-pix2pix}  & Generator with Skip, PatchGAN  & A, G, Q, R, S, U, BZ  & - & \href{13244}{https://scholar.google.com/scholar?cites=16757839449706651543&as_sdt=2005&sciodt=0,5&hl=en} 
  \\
  39& \href{CycleGAN}{https://openaccess.thecvf.com/content_ICCV_2017/papers/Zhu_Unpaired_Image-To-Image_Translation_ICCV_2017_paper.pdf}~\cite{zhu2017unpaired} & ICCV   & 2017  & \href{Github}{https://github.com/junyanz/pytorch-CycleGAN-and-pix2pix}  & Map Functions and Discriminators, Cycle Consistency Loss  & A, G, Q, R, S, U, AV, AW  & - & \href{12734}{https://scholar.google.com/scholar?cites=18396328236259959400&as_sdt=2005&sciodt=0,5&hl=en} \\
  40& \href{DualGAN}{https://openaccess.thecvf.com/content_ICCV_2017/papers/Yi_DualGAN_Unsupervised_Dual_ICCV_2017_paper.pdf}~\cite{yi2017dualgan} & ICCV   &2017  & \href{Github}{https://github.com/togheppi/DualGAN}  & Trained in Closed Loop, Reconstruction Loss &R, U, Y, CU, BZ, E  & - & \href{1554}{https://scholar.google.com/scholar?cites=6550565919250210407&as_sdt=2005&sciodt=0,5&hl=en} 
  \\
  41& \href{DiscoGAN}{http://proceedings.mlr.press/v70/kim17a/kim17a.pdf}~\cite{kim2017learning} & ICML  & 2017 & \href{Github}{https://github.com/SKTBrain/DiscoGAN}  & GAN with a Reconstruction Loss &  CI, K, I, AH, S & - & \href{1714}{https://scholar.google.com/scholar?cites=463778412690777341&as_sdt=2005&sciodt=0,5&hl=en}
  \\
  42& \href{BicycleGAN}{https://papers.nips.cc/paper/2017/file/819f46e52c25763a55cc642422644317-Paper.pdf}~\cite{zhu2017toward} & NeurIPS   & 2017   & \href{Github}{https://github.com/junyanz/BicycleGAN}  & cVAE-GAN,  cLR-GAN & R, S, U, BZ & -& \href{1114}{https://scholar.google.com/scholar?cites=1935169713047603976&as_sdt=2005&sciodt=0,5&hl=en} 
  \\
  43& \href{UNIT}{https://papers.nips.cc/paper/2017/file/dc6a6489640ca02b0d42dabeb8e46bb7-Paper.pdf}~\cite{liu2017unsupervised} & NeurIPS   & 2017    & \href{Github}{https://github.com/mingyuliutw/UNIT}  & Common Latent Space, VAEs, Cycle-consistency, GAN  & G, I, Q, V, W, X, BI  & - & \href{2138}{https://scholar.google.com/scholar?cites=14169741715291172305&as_sdt=2005&sciodt=0,5&hl=en} 
  \\
  44& \href{Pix2pixHD}{https://openaccess.thecvf.com/content_cvpr_2018/papers/Wang_High-Resolution_Image_Synthesis_CVPR_2018_paper.pdf}~\cite{wang2018pix2pixhd} & CVPR  & 2018  & \href{Github}{https://github.com/NVIDIA/pix2pixHD}  &  Coarse-to-fine Generator, Multi-scale Discriminator  & Q, AD, AE, AF  & - & \href{2527}{https://scholar.google.com/scholar?cites=8637738140607437341&as_sdt=2005&sciodt=0,5&hl=en} 
  \\
  45& \href{MUNIT}{https://openaccess.thecvf.com/content_ECCV_2018/papers/Xun_Huang_Multimodal_Unsupervised_Image-to-image_ECCV_2018_paper.pdf}~\cite{huang2018multimodal} & ECCV   & 2018  & \href{Github}{https://github.com/NVlabs/MUNIT}  & Content/Style Encoder, AdaIN, Decoder  & A, S, AP, BI, E  & - & \href{1615}{https://scholar.google.com/scholar?cites=13317525907573308290&as_sdt=2005&sciodt=0,5&hl=en} 
  \\
  46& \href{SPADE}{https://openaccess.thecvf.com/content_CVPR_2019/papers/Park_Semantic_Image_Synthesis_With_Spatially-Adaptive_Normalization_CVPR_2019_paper.pdf}~\cite{park2019semantic} & CVPR   & 2019  & \href{Github}{https://github.com/NVlabs/SPADE}  &Spatially-Adaptive Normalization, Pix2pixHD & F, Q, AE, AR  & Sm. & \href{1362}{https://scholar.google.com/scholar?cites=12479535951654162053&as_sdt=2005&sciodt=0,5&hl=en}
  \\
  47& \href{U-GAT-IT}{https://openreview.net/forum?id=BJlZ5ySKPH}~\cite{kim2019u} & ICLR   &2020  & \href{Github}{https://github.com/znxlwm/UGATIT-pytorch}  & Attention map, Adaptive Layer-Instance Normalization & AU, AV, AW, AX  & -  & \href{248}{https://scholar.google.com/scholar?cites=10390912983102174696&as_sdt=2005&sciodt=0,5&hl=en} 
  \\
  48& \href{CoCosNet}{https://openaccess.thecvf.com/content_CVPR_2020/papers/Zhang_Cross-Domain_Correspondence_Learning_for_Exemplar-Based_Image_Translation_CVPR_2020_paper.pdf}~\cite{zhang2020cross} & CVPR   &2020  & \href{Github}{https://github.com/microsoft/CoCosNet}  & Cross-domain Correspondence, Translation Network &AE, AC, BK &- & \href{104}{https://scholar.google.com/scholar?cites=18376074441966684157&as_sdt=2005&sciodt=0,5&hl=en}
  \\
  49& \href{TSIT}{https://www.ecva.net/papers/eccv_2020/papers_ECCV/papers/123480205.pdf}~\cite{jiang2020tsit} & ECCV   & 2020  & \href{Github}{https://github.com/EndlessSora/TSIT}  & Multi-scale Feature Normalization, Two-stream Network & Q,AE, AP, AW, BH  & - & \href{34}{https://scholar.google.com/scholar?cites=6228459104800637542&as_sdt=2005&sciodt=0,5&hl=en} 
  \\
  50& \href{DSMAP}{https://www.ecva.net/papers/eccv_2020/papers_ECCV/papers/123530562.pdf}~\cite{chang2020domain} & ECCV   & 2020 &\href{Github}{https://github.com/acht7111020/DSMAP}  &Domain-specific Content Mappings & AQ, AW, AX & -& \href{13}{https://scholar.google.com/scholar?cites=12420146798094423721&as_sdt=2005&sciodt=0,5&hl=en}
  \\
  51& \href{ACL-GAN}{https://arxiv.org/pdf/2003.04858.pdf}~\cite{zhao2020unpaired} & ECCV   & 2020  &\href{Github}{https://github.com/hyperplane-lab/ACL-GAN}  &Adversarial Consistency Loss, MUNIT & I, AU & - & \href{29}{https://scholar.google.com/scholar?cites=4792912034124725763&as_sdt=2005&sciodt=0,5&hl=en}
  \\
  52&\href{DRIT++}{https://link.springer.com/article/10.1007/s11263-019-01284-z}~\cite{DRIT_plus} & IJCV   &2020  & \href{Github}{https://github.com/HsinYingLee/DRIT}  & Disentangled Representation with Cross-cycle Consistency & AP, AQ, AW, AX, I  & - & \href{218}{https://scholar.google.com/scholar?cites=14537810097975332786&as_sdt=2005&sciodt=0,5&hl=en} 
  \\
  53& \href{CoCosNet v2}{https://openaccess.thecvf.com/content/CVPR2021/papers/Zhou_CoCosNet_v2_Full-Resolution_Correspondence_Learning_for_Image_Translation_CVPR_2021_paper.pdf}~\cite{zhou2021cocosnet} & CVPR   &2021  & \href{Github}{https://github.com/microsoft/CoCosNet-v2}  & ConvGRU Module, Hierarchical Strategy, PatchMatch & AE & - & \href{32}{https://scholar.google.com/scholar?cites=11302776795234140037&as_sdt=2005&sciodt=0,5&hl=en}
  \\
  54& \href{SofGAN}{https://dl.acm.org/doi/abs/10.1145/3470848?casa_token=7Jw8pKRSaBoAAAAA:l66EiRNumrjIl54plcMB7Aela1ZA0CcmVDIrYuX7TXPcika6tdJH2Nf94un1pk3o3a3u0dV7p-eI1w}~\cite{sofgan} & TOG   &2022  & \href{Project}{https://apchenstu.github.io/sofgan/}  & SOF Net, StyleGAN, Style Mixing, SPADE & AC,BU,I & Bm.,Sm.,Attri. & \href{11}{https://scholar.google.com/scholar?cites=123812763911102651&as_sdt=2005&sciodt=0,5&hl=en}
  \\
  \midrule
  \end{tabular}
  \begin{tablenotes}
      \textbf{Publ.}: Publication information.
      \textbf{Year}: Publication year.
      \textbf{Code}: The link of the corresponding open resources.
      \textbf{Components}: The key components of each model. 
      \textbf{Dataset}:
      A = TU-Berlin Sketch Dataset~\cite{eitz2012humans}, 
      B = Disney Portrait Dataset~\cite{berger2013style},
      C = FERET~\cite{phillips2000feret}, 
      D = AR~\cite{martinez1998ar}, 
      E = Self-Collected, 
      F = MSCOCO~\cite{lin2014microsoft},
      G = ImageNet~\cite{ILSVRC15}, 
      I = CelebA~\cite{liu2015deep},
      L = DTD~\cite{cimpoi2014describing},
      P = Wikiart~\cite{wikiart},
      Q = Cityspace~\cite{cordts2016cityscapes},
      R = CMP Facades~\cite{tylevcek2013spatial},
      S = Edge2photo~\cite{zhu2016generative,yu2014fine},
      U = Day2night~\cite{laffont2014transient},
      V = MNIST~\cite{lecun1998gradient},
      Y = CUFS~\cite{wang2008face},
      Z = Caltech-200 Bird~\cite{wah2011caltech},
      AC = CelebAHQ~\cite{karras2017progressive},
      AD = NYU Indoor RGBD dataset~\cite{silberman2012indoor},
      AE = ADE20K~\cite{zhou2017scene},
      AK = QMUL-Shoe-Chair-V2~\cite{yu2017sketchx},
      AL = QuickDraw dataset~\cite{ha2017neural},
      AP = Yosemite~\cite{zhu2017unpaired},
      AQ = cat2dog~\cite{DRIT_plus},
      AR = Flickr Landscapes~\cite{park2019semantic},
      AS = APDrawing Dataset~\cite{yi2019apdrawinggan},
      AT = Anime Faces of Getchu~\cite{jin2017towards},
      AU = Selfie2anime~\cite{kim2019u},
      AV = hourse2zebra~\cite{zhu2017unpaired},
      AW = photo2vangogh~\cite{zhu2017unpaired},
      AX = photo2portrait~\cite{DRIT_plus},
      BH = Berkeley Deep Drive~\cite{xu2017end},
      BI = SYNTHIA dataset~\cite{ros2016synthia},
      BJ = UPDG~\cite{YiLLR20},
      BK = DeepFashion~\cite{liu2016deepfashion},
      BU = FFHQ~\cite{karras2019style},
      BV = DIV2K~\cite{agustsson2017ntire}, 
      BW = LHI~\cite{yao2007introduction},
      BX = VIPSL~\cite{wang2011face},
      BY = IIIT-Delhi~\cite{bhatt2010matching},
      BZ = Map2Aerial~\cite{isola2017image},
      CB = StandfordCars~\cite{krause20133d},
      CH = LSUN~\cite{yu2015lsun},
      CU = CUFSF~\cite{zhang2011coupled}.
      \textbf{Assist.}: Assistant Information, \eg, Bm.= Background map,  Sm.= Segmentation map, Fl. = Facial landmark, Sv. = Style vector, Cm. = Color map, Attri. = Facial Attribute, Km. = Keypoint map, Tp. = Texture patch. 
      \textbf{Cite.}: Google citation statistics are from \Rev{2022-05-21.}
  \end{tablenotes}
\end{table*}

\begin{table*}[t!]
  \centering
  \tiny
  \renewcommand{\arraystretch}{1.1}
  \renewcommand{\tabcolsep}{0.8mm}
  \caption{
  Summary of popular related works.
  Please refer to \tabref{tab:related_work} for more detailed descriptions.}\label{tab:related_work3}
  \begin{tabular}{|c|r|c|c|c|c|c|c|r|}
  \toprule
  \#  & Model    & Publ.   & Year  & Code    & Component  & Dataset  & Assist. & Cite. \\\midrule
  \multicolumn{9}{c}{Deep Image-to-Sketch Synthesis}\\\midrule
  55& \href{FCRL}{https://dl.acm.org/doi/pdf/10.1145/2671188.2749321}~\cite{zhang2015end} &ICMR   & 2015  & -  &Fully Convolutional Network  & Y  & - & \href{127}{https://scholar.google.com/scholar?cites=1531407909888772421&as_sdt=2005&sciodt=0,5&hl=en} 
  \\
  56& \href{DGFL}{https://www.ijcai.org/proceedings/2017/0500.pdf}~\cite{zhu2017deep} &IJCAI   & 2017  & -  & Deep CNNs, Graphic model  & Y  & - & \href{34}{https://scholar.google.com/scholar?cites=3224061884788235903&as_sdt=2005&sciodt=0,5&hl=en} \\
  57& \href{Scribbler}{https://openaccess.thecvf.com/content_cvpr_2017/papers/Sangkloy_Scribbler_Controlling_Deep_CVPR_2017_paper.pdf}~\cite{sangkloy2017scribbler} & CVPR   & 2017 & \href{Project}{http://scribbler.eye.gatech.edu/}  & Encoder-decoder with residual connections, GAN   & Y, E  & - & \href{427}{https://scholar.google.com/scholar?cites=13445531679657986508&as_sdt=2005&sciodt=0,5&hl=en} \\
  58& \href{FSSC2F}{https://www.aaai.org/ocs/index.php/AAAI/AAAI18/paper/viewFile/16088/16357}~\cite{zhang2018face} & AAAI   & 2018 & -  & U-Net, Probabilistic Graphic Model  & Y  & - & \href{11}{https://scholar.google.com/scholar?cites=5363649240397599909&as_sdt=2005&sciodt=0,5&hl=en} 
  \\
  59& \href{TextureGAN}{https://openaccess.thecvf.com/content_cvpr_2018/papers/Xian_TextureGAN_Controlling_Deep_CVPR_2018_paper.pdf}~\cite{xian2018texturegan} & CVPR   & 2018  & \href{Github}{https://github.com/janesjanes/Pytorch-TextureGAN}  &Local Texture Loss, VGG Loss, Scribbler & E, S & Bm. Tp.& \href{221}{https://scholar.google.com/scholar?cites=12073710899908047608&as_sdt=2005&sciodt=0,5&hl=en}
  \\
  60& \href{SCC-GAN}{https://openaccess.thecvf.com/content_cvpr_2018/papers/Song_Learning_to_Sketch_CVPR_2018_paper.pdf}~\cite{song2018learning} & CVPR  & 2018 & \href{Code}{https://github.com/MarkMoHR/sketch-photo2seq}  & Hybrid model, Shortcut Cycle Consistency &AK, AL & - & \href{76}{https://scholar.google.com/scholar?cites=11408073639701679346&as_sdt=2005&sciodt=0,5&hl=en} \\
  61& \href{ContextualGAN}{https://www.ecva.net/papers/eccv_2018/papers_ECCV/papers/Yongyi_Lu_Image_Generation_from_ECCV_2018_paper.pdf}~\cite{lu2018image} & ECCV   & 2018  & \href{Github}{https://github.com/elliottwu/sText2Image}  &Contextual Loss, Joint Representation, GAN & I, Z, CB  & - & \href{74}{https://scholar.google.com/scholar?cites=5891894641710536105&as_sdt=2005&sciodt=0,5&hl=en}
  \\
  62& \href{pGAN}{https://www.ijcai.org/Proceedings/2018/0162.pdf}~\cite{zhang2018robust} & IJCAI   & 2018  & \href{Github}{https://github.com/hujiecpp/pGAN}  & UNet, Parametric Sigmoid, CycleGAN  & Y, CU  & Bm. & \href{24}{https://scholar.google.com/scholar?cites=6220907250365599477&as_sdt=2005&sciodt=0,5&hl=en} 
  \\
  63& \href{MRNF}{https://www.ijcai.org/Proceedings/2018/0159.pdf}~\cite{zhang2018markov} & IJCAI  &2018  & -  & Markov Random Neural Fields  &Y & - & \href{16}{https://scholar.google.com/scholar?cites=12981894418039705757&as_sdt=2005&sciodt=0,5&hl=en} 
  \\
  64& \href{PS$^2$-MAN}{https://ieeexplore.ieee.org/stamp/stamp.jsp?arnumber=8373815}~\cite{wang2018high} & FG  & 2018  & \href{Github}{https://github.com/lidan1/PhotoSketchMAN}  & Multi-Adversarial Networks, CycleGAN  & Y, CU  & - & \href{98}{https://scholar.google.com/scholar?cites=16947639257088162630&as_sdt=2005&sciodt=0,5&hl=en} 
  \\
  65& \href{DualT}{https://ieeexplore.ieee.org/stamp/stamp.jsp?arnumber=8463611&casa_token=lE73UIxSV7kAAAAA:47fOd3CDY3bdlU_N8M8A3cm_X00RFIbd2w_dzl1v2IjAiJugFuDac5X9hBefOxHqms7TStn5uRw}~\cite{zhang2018dual} & TIP   & 2018  & -  & Deep Features, Intra- and Inter-Domain Transfer & Y  & - & \href{51}{https://scholar.google.com/scholar?cites=14917409916062677053&as_sdt=2005&sciodt=0,5&hl=en}
  \\
  66& \href{MDAL}{https://ieeexplore.ieee.org/stamp/stamp.jsp?arnumber=8478205}~\cite{zhang2018multidomain} & TNNLS   & 2018   & \href{Github}{https://github.com/hujiecpp/MDAL}  & Domain alignment, Interpreting by Reconstruction  & Y, CU  & - & \href{45}{https://scholar.google.com/scholar?cites=585586712531552492&as_sdt=2005&sciodt=0,5&hl=en} 
  \\
  67& \href{FAG-GAN}{https://ieeexplore.ieee.org/stamp/stamp.jsp?arnumber=8347106&casa_token=_qpn3TG098AAAAAA:-sC_3s5naE7js922NQ33yYky8We1aBAw-ZOkvuh2st1HTpMAyNmVd6smafekLUjkO4GtocH4RJU}~\cite{kazemi2018facial} & WACVW  &2018  & -  & Attribute Classification, Conditional CycleGAN &I, C & - & \href{30}{https://scholar.google.com/scholar?cites=4976044318850268497&as_sdt=2005&sciodt=0,5&hl=en} 
  \\
  68& \href{Geo-GAN}{https://ieeexplore.ieee.org/stamp/stamp.jsp?arnumber=8552937&casa_token=sUNxnSb6HIAAAAAA:ZjSXjtvJ5SaJir1iRymxXzJXCUuX-G5trXg5ilxNx6M9oowLrW0D28RFlrDw9JSRjwPPwmAYxrI}~\cite{kazemi2018unsupervised} & BIOSIG   & 2018 & \href{Github}{https://github.com/rt219/Unsupervised-Sketch-to-Photo-Synthesis}  & Geometry Discriminator, CycleGAN  &CU, C  & - & \href{17}{https://scholar.google.com/scholar?cites=10713827734602026169&as_sdt=2005&sciodt=0,5&hl=en}
  \\
  69& \href{PI-REC}{https://arxiv.org/abs/1903.10146}~\cite{you2019pirec} & arXiv   & 2019  & \href{Github}{https://github.com/youyuge34/PI-REC}  &Multi-stage synthesis, LSGAN, VGG Loss & A, I, S, AT  & Cm. & \href{18}{https://scholar.google.com/scholar?cites=8595999052891287495&as_sdt=2005&sciodt=0,5&hl=en} 
  \\
  70& \href{DLLRR}{https://ieeexplore.ieee.org/stamp/stamp.jsp?arnumber=8621606}~\cite{zhang2019deep} & TNNLS  &2019 & -  & Coupled Autoencoder, Low-rank Representation &Y &- & \href{27}{https://scholar.google.com/scholar?cites=9500808269173003338&as_sdt=2005&sciodt=0,5&hl=en} 
  \\
  71& \href{Col-cGAN}{https://ieeexplore.ieee.org/stamp/stamp.jsp?arnumber=8621611&casa_token=tTrQSWE9ezgAAAAA:kjrR4afX1LMZH85XSj1dPkbFZNbnIETmgYjC873DPeJwjoxNQos9dUk9fdne-UqFQt-4HihjVcM}~\cite{zhu2019deep} & TNNLS  &2019 & -  & Collaborative Loss, cGAN, Deep Collaborative Nets &Y, CU &- & \href{43}{https://scholar.google.com/scholar?cites=9843745136039326784&as_sdt=2005&sciodt=0,5&hl=en} 
  \\
  72& \href{CFSS}{https://ieeexplore.ieee.org/stamp/stamp.jsp?arnumber=8848856&casa_token=qErUkYVu4wAAAAAA:V-Xf2A5S90Aagx1LeQ4nsfujj5Ozlb84qaE324uCiDEA3dvnxnkXEnouYFIo-V73w8g0Gd-5Itw}~\cite{zhang2019cascaded} & TIP  &2019  & -  &  cGAN, VGG, Feature Selection  &Y &- & \href{14}{https://scholar.google.com/scholar?cites=14972025413823209047&as_sdt=2005&sciodt=0,5&hl=en} 
  \\
  73& \href{KT}{https://www.ijcai.org/Proceedings/2019/0147.pdf}~\cite{zhu2019face} & IJCAI  &2019  & -  & Knowledge Transfer, Teacher-Student Net &Y, CU &- & \href{16}{https://scholar.google.com/scholar?cites=5252562650469968619&as_sdt=2005&sciodt=0,5&hl=en} 
  \\
  74& \href{im2pencil}{https://openaccess.thecvf.com/content_CVPR_2019/papers/Li_Im2Pencil_Controllable_Pencil_Illustration_From_Photographs_CVPR_2019_paper.pdf}~\cite{li2019im2pencil} & CVPR   &2019 &
  \href{Github}{https://github.com/Yijunmaverick/Im2Pencil}  & Outline and Shading Branch Networks, Pix2pix & E  & Sv.  & \href{28}{https://scholar.google.com/scholar?cites=206645996815512577&as_sdt=2005&sciodt=0,5&hl=en} 
  \\
  75& \href{ISF}{https://openaccess.thecvf.com/content_ICCV_2019/papers/Ghosh_Interactive_Sketch__Fill_Multiclass_Sketch-to-Image_Translation_ICCV_2019_paper.pdf}~\cite{ghosh2019interactive} & ICCV   & 2019  & \href{Project}{https://arnabgho.github.io/iSketchNFill}  & Shape and Appearance Generators, Two-stage & S, AC, E  & - & \href{62}{https://scholar.google.com/scholar?cites=11610888272506196409&as_sdt=2005&sciodt=0,5&hl=en} \\
  76& \href{APDrawing}{https://openaccess.thecvf.com/content_CVPR_2019/papers/Yi_APDrawingGAN_Generating_Artistic_Portrait_Drawings_From_Face_Photos_With_Hierarchical_CVPR_2019_paper.pdf}~\cite{yi2019apdrawinggan} & CVPR   & 2019  & \href{Github}{https://github.com/yiranran/APDrawingGAN}  & Hierarchical GAN, DT Loss, Local Transfer Loss   & AS  & Fl., Bm., Sv.& \href{82}{https://scholar.google.com/scholar?cites=7553502268301774290&as_sdt=2005&sciodt=0,5&hl=en} 
  \\
  77& \href{APDrawing++}{https://ieeexplore.ieee.org/stamp/stamp.jsp?arnumber=9069416}~\cite{yi2020line} & TPAMI   & 2020  & \href{Github}{https://github.com/yiranran/APDrawingGAN2}  & APDrawing, Line Continuity Loss  & AS  & Fl., Bm., Sv.& \href{12}{https://scholar.google.com/scholar?cites=12934399838149256595&as_sdt=2005&sciodt=0,5&hl=en} 
  \\
  78& \href{UPDG}{https://openaccess.thecvf.com/content_CVPR_2020/papers/Yi_Unpaired_Portrait_Drawing_Generation_via_Asymmetric_Cycle_Mapping_CVPR_2020_paper.pdf}~\cite{YiLLR20} & CVPR   & 2020 & \href{Github}{https://github.com/yiranran/Unpaired-Portrait-Drawing}  & Asymmetric CycleGAN, Cycle-consistency Loss & BJ  & Fl., Bm., Sv.& \href{22}{https://scholar.google.com/scholar?cites=8699048799354653218&as_sdt=2005&sciodt=0,5&hl=en} 
  \\
  79& \href{WCR-GAN}{https://openaccess.thecvf.com/content_CVPR_2020/papers/Wang_Learning_to_Cartoonize_Using_White-Box_Cartoon_Representations_CVPR_2020_paper.pdf}~\cite{wang2020learning} & CVPR   & 2020  & \href{Github}{https://github.com/SystemErrorWang/White-box-Cartoonization}  &Cartoon Representation Learning, GAN & F, BU, BV, E & -& \href{29}{https://scholar.google.com/scholar?cites=17523362885919981753&as_sdt=2005&sciodt=0,5&hl=en}
  \\
  80& \href{EdgeGAN}{https://openaccess.thecvf.com/content_CVPR_2020/papers/Gao_SketchyCOCO_Image_Generation_From_Freehand_Scene_Sketches_CVPR_2020_paper.pdf}~\cite{gao2020sketchycoco} & CVPR   & 2020  & \href{Project}{https://mikexuq.github.io/test_building_pages/index.html}  & SketchyCOCO, Divide-and-Conquer strategy & F  & Attri. & \href{34}{https://scholar.google.com/scholar?cites=753124261281861742&as_sdt=2005&sciodt=0,5&hl=en}
  \\
  81& \href{DeepPS}{https://www.ecva.net/papers/eccv_2020/papers_ECCV/papers/123600596.pdf}~\cite{yang2020deep} & ECCV   & 2020   & \href{Github}{https://github.com/VITA-Group/DeepPS}  & Sketch Refinement with Dilations, Pix2pixHD & AC, I &-& \href{25}{https://scholar.google.com/scholar?cites=13660754683270069229&as_sdt=2005&sciodt=0,5&hl=en}
  \\
  82& \href{DeepFaceDrawing}{https://dl.acm.org/doi/abs/10.1145/3386569.3392386}~\cite{chen2020deepfacedrawing} & TOG   & 2020 & \href{Github}{https://github.com/franknb/Drawing-to-Face}  &Component Embedding, Feature Mapping, Image Synthesis & AC, E & Km. & \href{41}{https://scholar.google.com/scholar?cites=1509088381513797333&as_sdt=2005&sciodt=0,5&hl=en}
  \\
  83& \href{CA-GAN}{https://ieeexplore.ieee.org/stamp/stamp.jsp?arnumber=9025751&casa_token=bm5Z8VFLB7wAAAAA:QF5zKbk4iN55M5_TMVu4xvVUHp0k2xWpy2Q5P9VmLLDGF92TSwvdPk8F_Oegd0YwwZxb11PSNW8}~\cite{yu2020toward} & TC  & 2020 & \href{Github}{https://github.com/fei-hdu/ca-gan}  &  Composition/Appearance Encoder, P-Net, Stacked GAN &Y, CU &Fl. & \href{44}{https://scholar.google.com/scholar?cites=524162025595250705&as_sdt=2005&sciodt=0,5&hl=en} 
  \\
  84& \href{IDA-CycleGAN}{https://www.sciencedirect.com/science/article/pii/S0031320320300558?casa_token=o8emV4xZf2wAAAAA:tQeDT_1szh0tVGfLWSNOCiDfToAqqNvlb8fxFI43pMRnfovS5p59wxyz9lSB24vPxqZhYwalAWI}~\cite{fang2020identity} & PR  & 2020 & -  & CycleGAN, Identity Loss, Recognition Model &Y, CU &- & \href{41}{https://scholar.google.com/scholar?cites=9339643841620048957&as_sdt=2005&sciodt=0,5&hl=en} 
  \\
  85& \href{IPAM-GAN}{https://ieeexplore.ieee.org/stamp/stamp.jsp?arnumber=9126135&casa_token=Tf7o0BUUlRwAAAAA:qyayYHfyQeTxj2avGsOgBXSfjICTx9nupCr4MjrwSI749joQB4oQFSbSw_2dilWCqkCae0HkHCc}~\cite{lin2020identity} & SPL   & 2020 & -  &Identity-preserved Adversarial Model, U-Net & Y, CU  & - & \href{12}{https://scholar.google.com/scholar?cites=11832474522828785434&as_sdt=2005&sciodt=0,5&hl=en}
  \\
  86& \href{MvDT}{https://ieeexplore.ieee.org/stamp/stamp.jsp?arnumber=9171460}~\cite{peng2020universal} & TIP   & 2020 & \href{Github}{https://github.com/clpeng/UniversalFPSS}  &CNN~\cite{simonyan2015very} Features, Hand-crafted Features  & Y, E  & - & \href{10}{https://scholar.google.com/scholar?cites=2069260044810875767&as_sdt=2005&sciodt=0,5&hl=en} 
  \\
  87& \href{MSG-SARL}{https://ieeexplore.ieee.org/document/9225019}~\cite{duan2020multi} & TIFS   & 2021 & -  & Self-attention Residual Learning, Multi-scale Gradients  & Y, CU  & - & \href{6}{https://scholar.google.com/scholar?cites=12968086217493244654&as_sdt=2005&sciodt=0,5&hl=en} 
  \\
  88& \href{GAN Sketching}{https://arxiv.org/pdf/2108.02774.pdf}~\cite{wang2021sketch} & ICCV   & 2021 & \href{Project}{https://peterwang512.github.io/GANSketching/}  & Weight Adjusting, Cross-domain Fine-tuning  & CH, AL  & - & \href{8}{https://scholar.google.com/scholar?cites=32634490776494372&as_sdt=2005&sciodt=0,5&hl=en}
  \\
  89& \href{DoodleFormer}{https://arxiv.org/pdf/2112.03258.pdf}~\cite{bhunia2021doodleformer} & Arxiv   & 2021 & - & Transformer, Part Locator and Part Sketcher Networks  & CK  & - & \href{1}{https://scholar.google.com/scholar?cites=3936607397967978494&as_sdt=2005&sciodt=0,5&hl=en}
  \\
  \bottomrule
  \end{tabular}
\end{table*}

\Rev{Recently, several researchers have begun using a small number of parameters to characterize each style, \ie, changing the parameters in the normalization layer for style transfer.
Dumoulin \etal~\cite{dumoulin2016learned} made the exciting observation that normalization layers can reflect the statistical properties of different styles. Therefore, they scaled and shifted the parameters in these layers while keeping the convolutional parameters unchanged to obtain better NST.
Further, they introduced flexible conditional instance normalization, enabling style transfer by simply changing the normalization parameters online.
Ulyanov \etal~\cite{ulyanov2017improved} improved their previous TextureNet \cite{ulyanov2016texture} by simply applying normalization to each image rather than a batch of images, which they called instance normalization. 
Moreover, they also demonstrated that the style transfer network with instance normalization could converge faster than that with batch normalization while achieving visually better results. 
Later, Huang and Belongie \cite{huang2017arbitrary}, following a similar idea, introduced adaptive instance normalization into the GAN model, aligning the content and style features. 
Li \etal~\cite{WCT-NIPS-2017} further used the first few layers of a pre-trained VGGNet \cite{simonyan2015very} to extract the feature representation. 
However, they replaced the AdaIN layer with whitening and coloring transformations, enabling the universal style transfer. 
Similar to I2SGAN \cite{abdal2019image2stylegan}, Richardson \etal~\cite{richardson2020encoding} improved the classical StyleGAN with a novel encoder network that learns many style vectors that are fed into a pre-trained generator, forming an extended $\mathcal{W}+$ latent space.}

\subsection{Deep Photo-Sketch Synthesis}
Deep photo-sketch synthesis is a recent branch of the FSS task, in which deep learning is used to improve performance and quality. The related works can be divided into three categories. The first aims to translate any sketch images into their corresponding RGB images. The second tries to convert any RGB images into sketch images. The last mainly focuses on facial-sketch synthesis.

\textbf{General S2I.}
Xian \etal~\cite{xian2018texturegan} proposed the TextureGAN model to synthesize an image under the supervision of a sketch, color, and texture. 
TextureGAN consists of a ground-truth pre-training module and an external texture fine-tuning part.
Then, Lu \cite{lu2018image} \etal~introduced a two-stage contextual GAN to achieve sketch-to-image generation. This framework trains a classical GAN model with a newly defined loss, representing the joint distribution and capturing the inherent relation between a sketch and its corresponding image.
Inspired by image in-painting \cite{song2018spg}, You \etal~\cite{you2019pirec} proposed the PI-REC model, which contains three phases: an imitation phase, generating phase, and refinement phase. PI-REC is progressively trained using only one generator and one discriminator.
The ISF introduced in \cite{ghosh2019interactive} is a gating-based approach, which allows a single generator to be used to generate distinct classes without feature mixing.
Recently, Gao \etal~\cite{gao2020sketchycoco} proposed EdgeGAN for object-level image synchronization given freehand scene sketches. 
This framework contains two sequential modules: foreground generation and background generation.
Yang \etal~\cite{yang2020deep} presented a deep plastic surgery model to simulate the coarse-to-fine painting process of human artists.
Chen \etal~\cite{chen2020deepfacedrawing} proposed a local-to-global framework to allow any user to produce high-quality face images. Their model consists of three modules: component embedding, feature mapping, and image synthesis. 

\textbf{General I2S.}
Song \etal~\cite{song2018learning} proposed the first deep stroke-level photo-to-sketch synthesis method, which is a hybrid model with a shortcut cycle consistency constrained by a VAE-style reconstruction loss. 
As the default settings of I2I and NST, both can synthesize artistic portrait drawing (APD) images. However, they do not meet practical requirements because APD images usually have a highly abstract style and graphic elements.
Therefore, Yi \etal~\cite{yi2019apdrawinggan} proposed APDrawing to transform an input face image into its corresponding APD image, in which a hierarchical GAN model is built by combining both a global and a local network.
Then, they further proposed an APDrawing++ \cite{yi2020line}, in which they used an auto-encoder to refine subtle facial features and presented a novel line continuity loss to enhance the line continuity of APDrawing.
However, both of these APDrawing methods require pair-wise data for training. To handle this problem, Yi \etal~thus proposed an asymmetric cycle-structure GAN  \cite{YiLLR20}, which contains a relaxed forward cycle consistency loss (\emph{a.k.a.} truncation loss) to prevent the reconstructed photo from being noisy, and a strict cycle consistency loss to enhance the performance. This method also uses multiple local discriminators to ensure the quality of the facial portrait drawings.
Different from portrait drawing, Wang \etal~\cite{wang2020learning} observed the behavior and properties of cartoon paintings and proposed three different representations considering surface, texture, and shape information, respectively.
In addition, they also released the new SketchyCOCO dataset to better train and evaluate the performance of their model.
Based on Pix2pix, Li \etal~\cite{li2019im2pencil} designed a two-branch network (called im2Pencil) to implement photo-pencil translation, which can simulate sketch outlines and shadows.
Wang \etal~\cite{wang2021sketch} presented a GAN sketching method to rewrite a GAN with one or more sketches. This new method uses regularizations to preserve the original GAN's diversity and image quality while matching the generated sketch images with users' needs through a cross-domain adversarial loss.
Bhunia \etal~\cite{bhunia2021doodleformer} introduced a new transformer architecture to generate various yet realistic creative sketches consisting of two networks. The first part of locator networks aims to capture the coarse structure by observing the relationship between local patterns. The second part of the sketcher network, follows the standard GAN, which aims to synthesize high-quality sketches.

\begin{figure*}[t!]
\centering
\includegraphics[width=\linewidth]{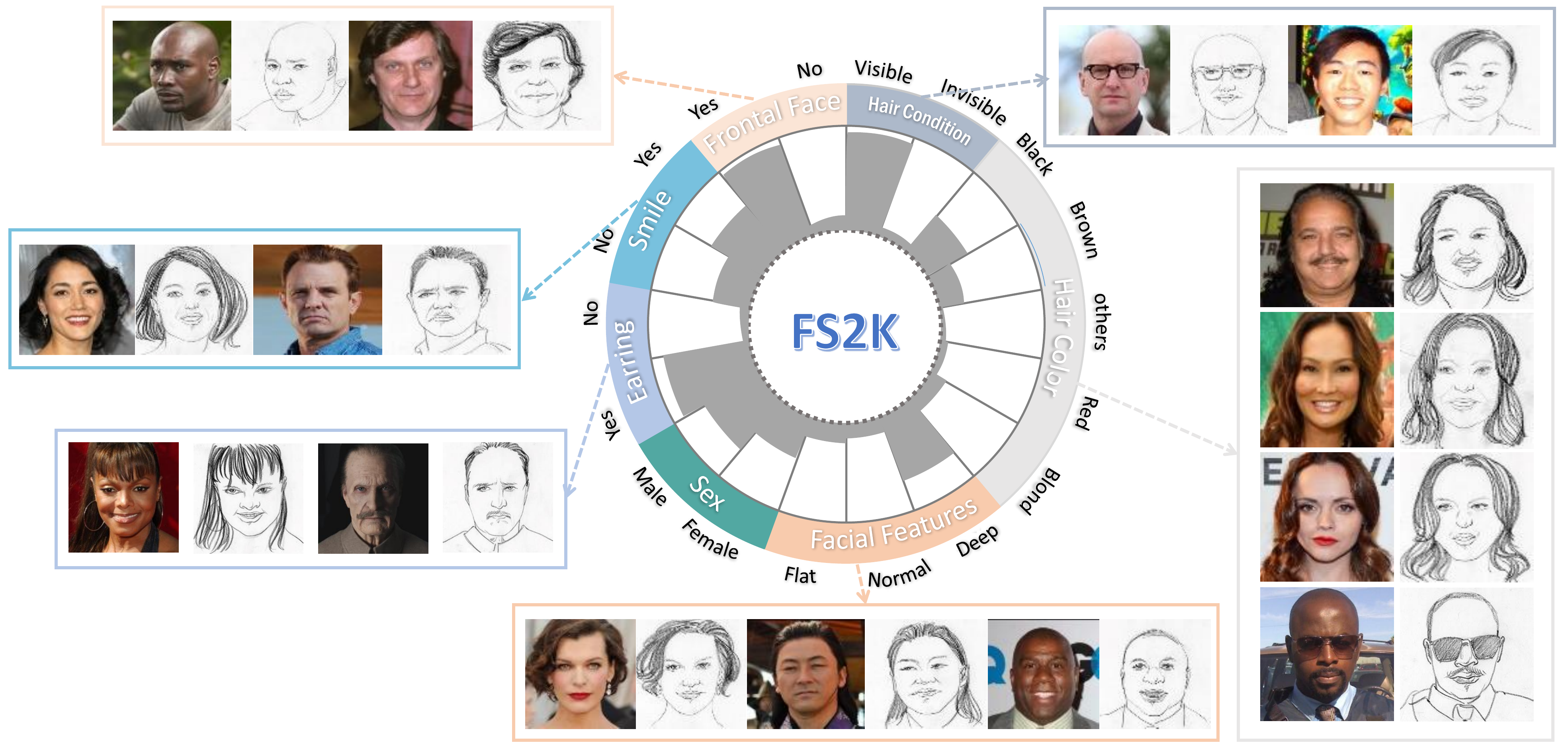}
\caption{Statistics and examples from the \ourdataset~dataset. Please refer to \secref{sec:FS2K-Dataset} for details.}\label{fig:FS2KDataset}
\end{figure*}

\textbf{Photo-Sketch Synthesis.} 
Zhang \etal~\cite{zhang2015end} were the first to use a fully convolutional neural network (FCNN) to build a deep photo-to-sketch synthesis model. 
Then,  the works \cite{zhu2016generative,zhang2018face,zhang2018markov} integrated deep features into probabilistic graph model learning, achieving better performance than traditional models \cite{wang2008face,zhou2012markov}. 
To make the network more flexible, Zhang \etal~\cite{zhang2018robust} took the key idea of CycleGAN and proposed a novel pGAN, which uses a special parametric Sigmoid activation function to reduce the effects of photo priors and illumination variations.
To improve the quality of generated photo/sketch, Wang \etal~\cite{wang2018high} introduced a synthesis method using multi-adversarial networks (PS$^2$MAN). Their model uses two U-Nets to generate high-quality images from low to high resolution.
To achieve the same goal, Zhang \etal~\cite{zhang2018multidomain} further proposed a facial-sketch synthesis by multi-domain adversarial learning (MDAL), which overcomes the defects of blur and deformation. 
The basic idea behind MDAL is the concept of ``interpretation through synthesis'', which is built upon two diverse generators.
Kazemi \etal~\cite{kazemi2018facial,kazemi2018unsupervised} proposed an improved version of CycleGAN, which focuses on the facial attributes during the portrait synthesis process.
Zhang \etal~\cite{zhang2019deep,zhang2019cascaded} introduced two methods by combining an auto-encoder and traditional subspace learning, which is more effective than the traditional FSS methods.
Besides, Zhu \etal~\cite{zhu2019deep} proposed a collaborative framework that exploits the interaction information of two opposite generators by introducing a collaborative loss.
However, it is difficult to train a good model due to the lack of large-scale training data. 
Therefore, Zhu \etal~\cite{zhu2019face} proposed using classical knowledge distillation to learn two well-defined student mapping networks via two strong teacher networks.
More recently, the works in \cite{fang2020identity,lin2020identity} introduced identity-aware models, which use a new perceptual loss to train a better image generative model, and thus consider the downstream task, \eg, face recognition, as the final goal.
Yu \etal~\cite{yu2020toward} proposed a new composition-assisted generative adversarial network, which helps synthesize realistic facial sketches/photos by using facial composition information.
By leveraging the relationships between features, \cite{duan2020multi} implemented a multi-scale self-attention residual learning framework for face photo-sketch conversions.
Finally, the method proposed in \cite{peng2020universal} does not need any images from the source domain for training, enabling it to leverage both deep features (extracted from the CNN) and handcrafted features flexibly.

\section{Proposed \ourdataset~Dataset}\label{sec:FS2K-Dataset}
In this section, we introduce the proposed \ourdataset. Some example images are shown in \figref{fig:RepresentativeExample}. We describe \ourdataset~in terms of two key aspects, namely dataset collection, and data annotation.
Overall, \ourdataset~includes \datasetNum~photo-sketch pairs, which are split into $\datasetNumTrain$ for training and $\datasetNumTest$ for testing. The complete dataset is available at \url{https://github.com/DengPingFan/FS2K}.

\subsection{Data Collection}
To establish a long-lasting benchmark, the data should be carefully selected to cover diverse scenes from different views, such as lighting conditions, skin colors, sketch styles, and image backgrounds.  
To this end, we introduce \ourdataset, a new high-quality dataset\footnote{This dataset is for scholarly communication only.} for the FSS task.

Our \ourdataset~includes \datasetNum~photos from real scenes, the Internet, and other datasets. 
The majority, however, come from CASIA-WebFace~\cite{yi2014learning}, which is a large-scale (\ie, $500K$ images) labelled dataset of faces in the wild. 
CASIA-WebFace was collected from the IMDb\footnote{\url{http://www.imdb.com}} website and contained well-organized information, such as name, gender, and birthday. 
Thanks to the rich and clean open-source data from CASIA-WebFace, it could be used to build our high-quality and representative benchmark. 
We manually selected $\datasetNumSrcOne$ images to cover a large span of major challenges faced in realistic scenes, such as varying background, hairstyle (\eg, long, short), accessories (\eg, glasses, earrings), and skin information (\eg, patch image on a given face).  
Because the photos selected in CASIA-WebFace are taken from a single angle, multi-angle face images for the same person are missing. 
To this end, we invited eight actors to take $\datasetNumSrcTwo$ photos under different settings (\eg, lighting conditions, face angles). 
In addition, to further increase the diversity, we also collected some children's photos and some faces with smaller face-to-image ratios.
The remaining $\datasetNumSrcThree$ face 
photos come from other free stock photos websites, 
including Unsplash,\footnote{\url{http://www.unsplash.com}}
Pexels,\footnote{\url{http://www.pexels.com/}} 
Pngimg,\footnote{\url{http://pngimg.com/}} 
and Google. 

\begin{figure*}[t!]
  \centering
  \begin{overpic}[width=.98\linewidth]{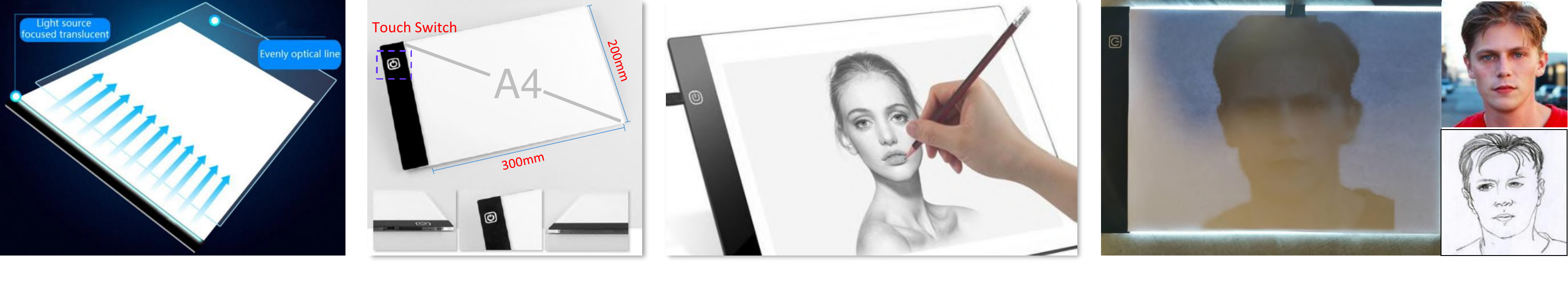}
   \put(5,0){\footnotesize{(a) Schematic}}
   \put(25,0){\footnotesize{(b) Specification}}
   \put(48,0){\footnotesize{(c) Virtual example}}
   \put(79,0){\footnotesize{(d) Our example}}
  \end{overpic}
  \caption{Use of the copy table and an example. Zoomed-in for the best view. See \secref{sec:DataAnnotation} for more details.
  }\label{fig:CopyTable}
\end{figure*}

\begin{figure}[t!]
  \centering
  \begin{overpic}[width=.98\linewidth]{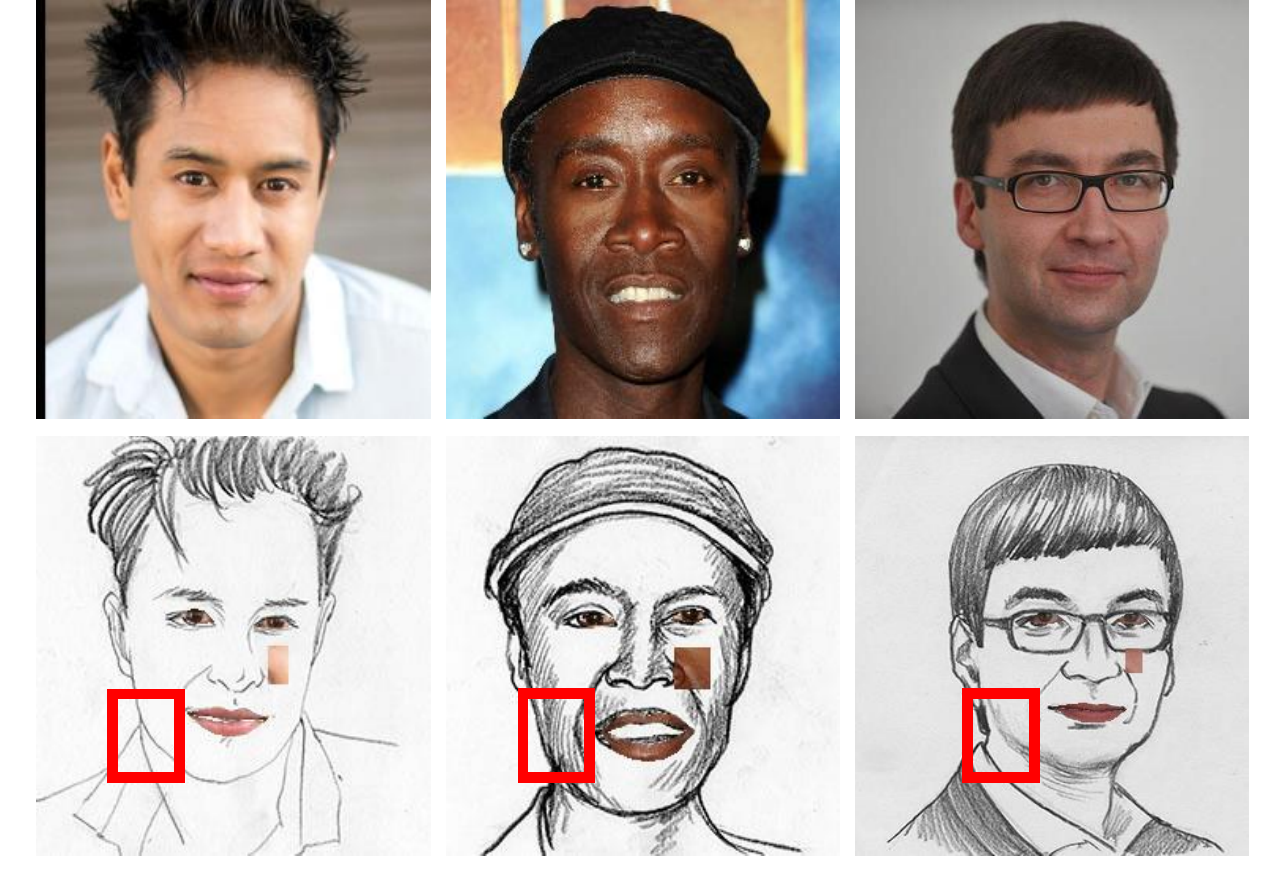}
   \put(10,-3){\footnotesize{(a) Style 1}}
   \put(42,-3){\footnotesize{(b) Style 2}}
   \put(73,-3){\footnotesize{(c) Style 3}}
  \end{overpic}
  \vspace{8pt}
  \caption{Three sketch styles in our \ourdataset. As shown in the cheek region, the styles include simple lines (style 1), long strokes (style 2) and repeated wispy details (style 3).
  }\label{fig:SketchStyle}
\end{figure}

\subsection{Data Annotation}\label{sec:DataAnnotation}
There are four types of annotations in our \ourdataset, including sketch drawing, sketch style, color, and contour feature annotations. 

\subsubsection{Sketch Drawing}
\textbf{Participants.}
Three senior artists (including two male and one female) from the Sichuan Fine Arts Institute were hired to participate in the study.\footnote{The \href{https://www.scfai.edu.cn/english/}{Sichuan Fine Arts Institute} is one of the four most prominent art academies in China. The three senior artists are all from the Design Academy.} 
All three participants had normal or corrected to normal vision. None of the participants suffered color-blindness or color-weakness. The participants ranged in age from $20$ to $23$ years, with an average of five years of professional experience in sketch drawing. 

\textbf{Apparatus.} 
The three artists drew all sketch images with the assistance of a Copy Table LED Board.\footnote{\figref{fig:CopyTable}-a presents the copy table, which has an LCD backlight. It requires a high voltage input of $100\sim240$V and $0.6$A working current.  
Its size is A4 (\ie, $300\times 200\times 3.5$mm) in \figref{fig:CopyTable}-b, and the luminous intensity is $300\sim350$LM. Therefore, it has become the most popular copy table product, after the aluminum alloy copy table, for animators (see \figref{fig:CopyTable}-c).} \figref{fig:CopyTable} shows the copy table we used and an example (\figref{fig:CopyTable}-d) of a face sketch drawn by our artists.
The touch switch region in our device supports three levels of adjustable brightness, so the artists can use the button to change the brightness they desire.
This helped them locate the contours of facial features according to the photo information from the bottom of the LED board. 
Moreover, this equipment also helped to ensure content similarity and face alignment between sketches and corresponding photos. At the same time, the drawings retain the artist's sketch style.

\subsubsection{Sketch Style Annotation}\label{sec:SketchStyleAnn}
Our \ourdataset~contains three different styles, which enrich the diversity of sketches, as shown in \figref{fig:SketchStyle}.
This enables different artists' skills to be captured while making FS2K more challenging than previous FSS datasets.

We created a balanced dataset to facilitate the comparison of different methods, \ie, the number of the images with the three different styles are equally distributed. 
Specifically, in the training set, the samples with style1, style2, and style3 are $357$, $351$, and $350$, respectively. In the test set, they are $619$, $381$, and $46$, respectively.

\begin{table*}[t!]
  \centering
  \footnotesize
  \renewcommand{\arraystretch}{1.0}
  \renewcommand{\tabcolsep}{3.5pt}
  \caption{
   Number of images for each attribute in the training and test datasets.
  }\label{tab:AttributeStatis}
  \begin{tabular}{l||cc|cccc|cc|cc|cc|cc|ccc}
  \hline
  \ourdataset~(Ours) &w/ H &w/o H &H(b) &H(bl) &H(r) &H(g) & M &F & w/ E & w/o E & w/ S &w/o S &w/ F &w/o F &S1 &S2 &S3\\
   \hline
   Train &1010 &48 &288 &423 &60 &239 &574 &484 &209 &849 &645 &413 &917 &141 &357 &351 &350 \\
   Test &994 &52 &290 &418 &44 &242 &632 &414 &187 &859 &670 &376 &872 &174 &619 &381 &46 \\
   \hline     
  \end{tabular}
\end{table*}

\subsubsection{Facial Feature Annotation}\label{sec:ContourFeature}
Sketches are rapidly executed freehand drawings, which have less attribute information than the original images, \eg, facial texture, facial expressions~\cite{wang2014feature}, and facial posture. 
Therefore, it is challenging to restore real images (\ie, S2I task) based on a single sketch image. 
Meanwhile, in real-world applications, we can use auxiliary facial information (such as gender, accessories, and hairstyle) to narrow down a suspect in a database.
Following \cite{zheng2020survey}, we added some additional facial feature annotations, including \textit{gender, smile, face pose, hair condition, hair color, earring,} and \textit{skin texture}. 
We hired two data annotators to label all photos and performed cross-checking to ensure the accuracy of the final annotations.
Overall labels can be found in \tabref{tab:AttributeStatis}, while the details of each are described below.

\textbf{Gender.} 
Gender is a high-level human attribute commonly used in traditional face databases such as CelebA \cite{liu2015deep} and LFW \cite{huang2008labeled}. 
It has been extensively studied in face detection and recognition \cite{ranjan2017hyperface,hand2017attributes,han2017heterogeneous}. Therefore, we carefully labelled all photos in \ourdataset~with gender attributes. 
Specifically, there were $574$ male photos and $484$ female photos in the training set, and $632$ male photos and $414$ female photos in the test set.

\textbf{Smile.} 
Smiling is a primary human activity that represents a positive emotional state.  
As such, many studies have focused on smile detection \cite{jang2017smilenet,ranjan2017all} or used smile as an attribute for recognition \cite{li2020deep}. 
Therefore, we also consider a smile a key attribute in our dataset. Specifically, the training set contains $645$ smiling people and $413$ with no obvious expression, while the test set contains $670$ smiling people and $376$ with no expression. We ensured that the proportion of smiling people in the training and test sets was as close as possible.

\textbf{Face Pose.} 
The facial attributes may cover only a small part of the image, but the photo is usually dominated by the effects of pose \cite{zhang2014panda}. Moreover, pose will affect the performance of face recognition \cite{kan2014stacked}, tracking \cite{wu2013facial}, and synthesis \cite{tran2017disentangled}. 
Therefore, the facial pose is useful auxiliary information. 
We define a portrait with the head rotated within $30$ degrees as a frontal face pose. According to this definition, the training set has $917$ frontal photos, while the test set has $872$. The remaining have side face poses.

\textbf{Hair Status and Color.}
Hair is a saliency feature of the head that may change in different situations. 
Even if there is sufficient information in the internal features of the face for recognition, manipulating the hair can harm the performance \cite{toseeb2012significance,bartel2018know}. 
Moreover, facial synthesis and retrieval systems often use hair as an important cue \cite{kumar2008facetracer,huai2018facial} to improve the quality of generated images. 
For FSS, although the sketches contain the hair contour, the corresponding color information and hair status (with or without hair) are missing. Therefore, in \ourdataset, we provide annotations of the hair status, which includes four available colors (\ie, black, brown, red, and blond) and another status (\ie, bald or wearing a hat), as shown in \figref{fig:FS2KDataset}. In other words, for faces with hair, we mark the color information directly, while cases of thinning hair or wearing a hat are marked as separate attributes.
The statistical results of this annotation can be found in \tabref{tab:AttributeStatis}.

\textbf{Earrings.}
The simplified characteristics of sketch drawings lead to unclear earring contours. Meanwhile, as shown in \figref{fig:FS2KDataset}, earrings in real photos are visible. 
Therefore, in \ourdataset, we provide annotations for whether earrings are present, which can help the model training. 
Specifically, the training set has $209$ people with earrings, and the test set has $187$. 

\textbf{Skin Texture.}
Skin texture provides a large amount of detailed local information and is used as a vital feature for face recognition \cite{pierrard2007skin,li2009encyclopedia}. 
However, this critical information is completely lost in sketch images.
Therefore, we clip a small patch from the real photo and use it as the skin texture, as shown in \figref{fig:SketchStyle}. 
We also include the average RGB value for the corresponding lip and eyeball region to provide more information for future research.

\begin{figure*}[t!]
\centering
\includegraphics[width=.95\linewidth]{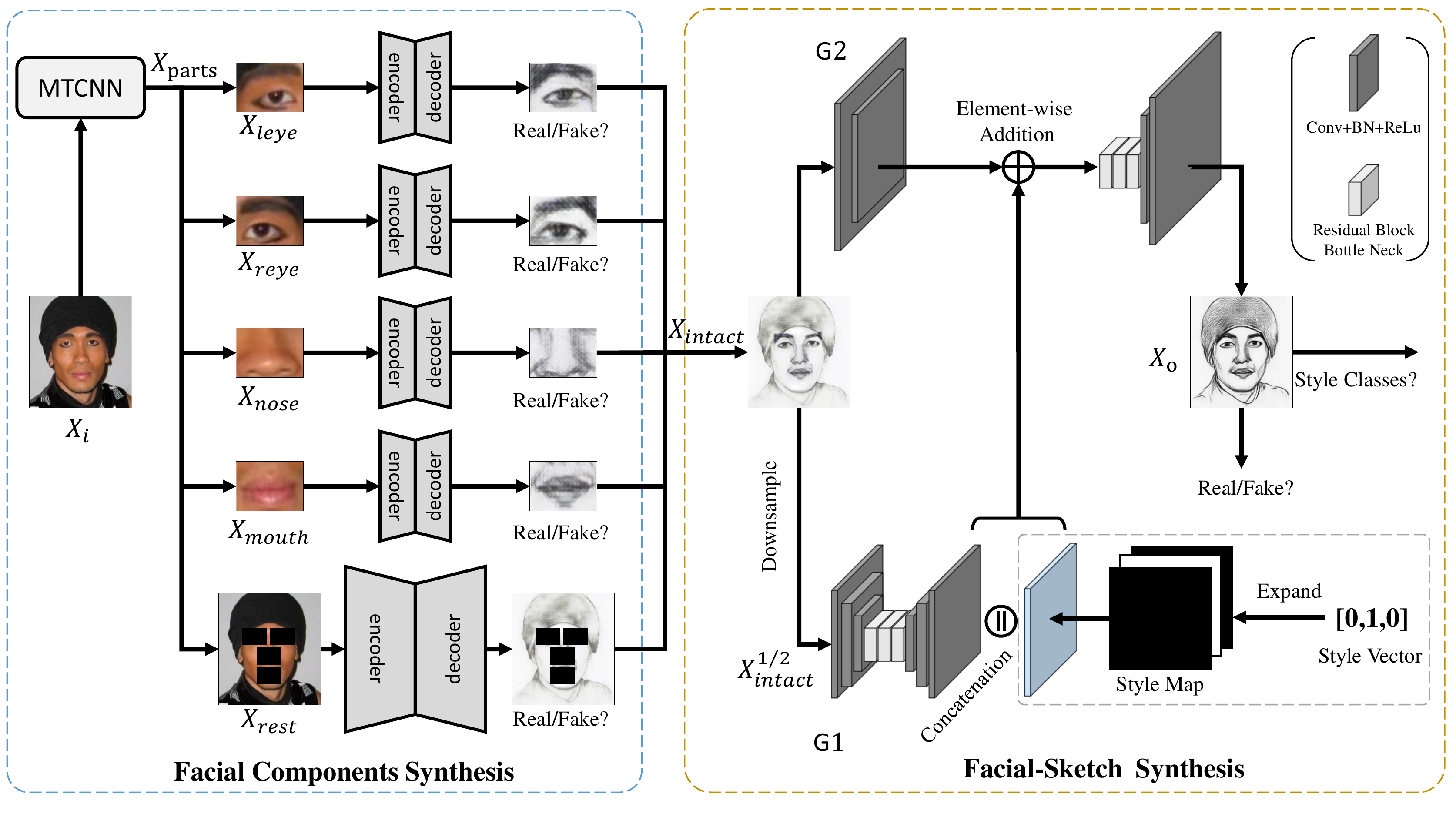}
\caption{Pipeline of our \ourmodel~baseline for the I2S task. It consists of two stages: 1) facial components synthesis and 2) facial-sketch synthesis. Please refer to \secref{sec:FeatureSyn} and \secref{sec:FacialSyn} for more details.}\label{fig:fsgan}
\end{figure*}

\section{Proposed \ourmodel~Baseline}\label{sec:Baseline}
\subsection{Problem Definition}
Facial synthesis (FS) aims to generate target representations of human faces based on the given inputs.
This process can be formulated as $X_o = F(X_i)$, where $X_i$ and $X_o$ denote the input and output (\eg, RGB images and sketches) of facial representations $F$ indicates the synthesis function. 
\Rev{In this paper, based on the overall architecture of \cite{yi2019apdrawinggan,yi2020line}, we design the baseline, 
FSGAN, for both the I2S task\footnote{\Rev{$X_{ske} = F(X_{img}, X_{style})$, where $X_{style}$ denotes the sketch style of input.}} and S2I task,\footnote{\Rev{$X_{img} = F(X_{ske})$.}} inspired by pix2pixHD~\cite{wang2018pix2pixhd}.} Instead of focusing on direct image-level facial synthesis, we propose a two-stage ``bottom-up'' facial synthesis architecture, as shown in Fig. \ref{fig:fsgan}. 
Hence, our FSGAN consists of two cascaded stages built upon multiple generative models (\ie, GANs). 

The first stage comprises of five parallel GANs, which are designed to synthesize the local facial components separately.
Given an input, four facial regions (\eg, left eye, right eye, nose, and mouth) and the rest of the inputs are cropped and fed into their corresponding GANs in the first stage to synthesize key facial features. 
These synthesized facial component patches are then stitched together to obtain the intact facial representation. 
Since the local facial patches are synthesized independently, the connecting region of the stitching, as well as their appearances, are inconsistent with each other. 
Therefore, the second stage is introduced to further refine the results by considering the global structure and texture. 
In this stage, the style vectors of the facial sketches are utilized to assist the synthesis. 

\subsection{Facial Components Synthesis}\label{sec:FeatureSyn}
Almost all human faces have the same global structure. 
The differences lie in the details of the local facial components, such as eyes, eyebrows, nose, and mouth.
To capture more details of different facial components, the first stage of our model synthesizes them separately. 
Specifically, given a facial input, the four key patterns, including the left eye, right eye, nose, and mouth, are first detected by MTCNN~\cite{zhang2016joint}. 
The input $X_i$ is then divided into five parts, $X_{parts} = $ \{$X_{leye}$, $X_{reye}$, $X_{nose}$, $X_{mouth}$, $X_{rest}$\}, based on the detection results. These include the left eye, right eye, nose, mouth, and remaining components. 
Five parallel GANs are utilized to synthesize their corresponding patches for these parts. 
Therefore, the problem can be formulated as $G_{parts} = $\{$G_{leye}$, $G_{reye}$, $G_{nose}$, $G_{mouth}$, $G_{rest}$\} and $D_{parts} = $\{$D_{leye}$, $D_{reye}$, $D_{nose}$, $D_{mouth}$, $D_{rest}$\}, where $G$ and $D$ indicate the generator and discriminator, respectively. 

First, the four GANs synthesizing the left eye, right eye, nose, and mouth have the same architecture. 
Each GAN consists of a generator and a discriminator. 
The generator is designed as an encoder-decoder, consisting of an encoder, a bottom connection, and a decoder. 
The encoder is composed of three convolutional blocks, each of which is a combination of a convolutional layer (with a kernel size of $3$ and stride of $2$), a batch normalization layer, and a ReLU activation layer. 
Meanwhile, the second bottom connection consists of nine bottleneck residual blocks that are similar to \cite{HeZRS16}. 
Finally, the decoder is built upon three deconvolutional blocks: a deconvolutional layer, a batch normalization layer, and a ReLU activation layer. 
Note that the GAN, which is used for synthesizing $X_{rest}$, is similar to the previously described ones. 
However, the encoder contains four convolutional blocks, and the decoder comprises four deconvolutional blocks to achieve larger receptive fields. 

The discriminators of the above five GANs are the same. 
Each consists of three cascaded convolutional layers (with a kernel size of $3$ and stride of $2$) followed by global average pooling. 
Then, a $1\times1$ convolutional layer and a sigmoid function are used to predict the probability of the generated results being real or fake.

Based on the above design, the first stage of FSGAN can restore details of the facial components in both the I2S and S2I tasks. 
At the end of this stage, the synthesized patches are stitched together to restore the intact facial synthesis result $X_{intact}$. 
Since different generators synthesize the patches, their overall appearances are inconsistent, which becomes even more obvious in the stitched result. 
To this end, the stitched result is then fed to the next stage to adjust and refine the global structure and appearance.

\subsection{Facial-Sketch Synthesis}\label{sec:FacialSyn}
To address the inconsistency issue of the output from the first stage, we introduce the second stage, which is designed as another GAN model inspired by Pix2pixHD~\cite{wang2018pix2pixhd}, for local detail refinement and global structure adjustment.

In this stage, we use the multi-scale discriminators $D_{fs}$ and the coarse-to-ﬁne generator $G_{fs}$ following Pix2pixHD~\cite{wang2018pix2pixhd}. 
Specifically, the generator $G_{fs}$ consists of two sub-networks $G1$ and $G2$, both of which follow encoder-decoder architecture, as shown on the right part of \figref{fig:fsgan}. 
We sample the output of the first stage using a downsampling operation with a sampling rate of $50\%$. This newly sampled image $X^{1/2}_{intact}$ ($height/2, width/2$) is then fed into the first sub-network $G1$, which is designed to capture global features.
The other sub-network $G2$ is employed to capture the local details, which takes the output of the first stage as input.
We use both concatenation and element-wise addition operations to fuse the style, local, and global information. 
Specifically, the concatenation combines the style feature map and the output of $G1$ and generates a new fused feature map. 
Then, the element-wise addition is utilized to combine this new feature map with the latent feature of the encoder part of $G2$. Finally, we use the decoder part of $G2$ to generate the final output $X_o$. 
It is worth noting that the style vector can control the style of the generated sketches, which helps improve their quality and diversity. 
Besides, the style of the real photo is often fixed, and independent from the artists' style.
Therefore, we introduce the style information in the I2S task but exclude that in the S2I task.

\subsection{Loss Function}
We use a combination of several loss functions to train our model. We denote $X$ and $Y$ as the input and its corresponding \Rev{reference}, respectively. 
For simplicity, we define $G(X)$ as the generated output of the given input $X$ and $D_k(X, Y)$ as the corresponding predicted probabilities of the $k$-th discriminator. Then, we denote the $i$-th layer feature extractor of discriminator $D_k$ as $D^i_k$, where $k$ is the index of the discriminator.

\textbf{Adversarial Loss.} 
We use the adversarial loss \cite{goodfellow2014generative} to make the generated image more visually appealing. The adversarial loss we use is defined as: 
\vspace{-5pt}
\begin{equation}
\begin{aligned}
    {L}_\mathrm{adv}(G, D) = ~&\mathbb{E}_{X,Y}[\log D(X, Y)] \\
                            &+ \mathbb{E}_{X}[1 - \log D(X, G(X))].
\label{eq:adversarial_loss}
\vspace{-5pt}
\end{aligned}
\end{equation}

\textbf{Feature Matching Loss.} 
Similar to \cite{wang2018pix2pixhd}, we use the feature matching loss to improve the adversarial loss based on the $k$-th discriminator. The feature matching loss is defined as: 
\begin{equation}
\begin{aligned}
    & {L}_\mathrm{fm}(G, D_{k}) \\
    & = \mathbb{E}_{X,Y}\sum\nolimits_{i=0}^{T}\frac{1}{N_{i}}\big[\big\|D_k^{i}(X, Y) - D_k^{i}(X, G(X))\big\|_1\big], 
\label{eq:fm_loss}
\vspace{-5pt}
\end{aligned}
\end{equation}
where $T$ denotes the total number of layers in each discriminator and $N_{i}$ is the number of feature maps in the $i$-th layer. 
This loss is used to match the intermediate feature maps of the real and synthesized images, making the generator produce multi-scale statistical information.
Besides, it stabilizes the training process and restores highly realistic outputs. 

\textbf{Perceptual Loss.} 
To maintain perceptual and semantic consistency, we use a perceptual loss \cite{johnson2016perceptual} to measure the difference between the original image and the corresponding synthesized image. 
We extract the perceptual features from the $i$-th layer activations of a pre-trained VGGNet~\cite{simonyan2015very}, which is denoted as $\phi_{i}(\cdot)$.
The perceptual loss is defined as follows:
\begin{equation}
\begin{aligned}
    {L}_\mathrm{per}\big(G(X), Y\big) = \mathbb{E}_{G(X),Y}\sum_{i=0}^{t}\big\|\phi_{i}(Y) - \phi_{i}(G(X))\big\|_1.
\label{eq:per_loss}
\end{aligned}
\end{equation}

\textbf{Pixel-Wise Loss.}
The ${L}_1$ distance between a generated image $G(X)$ and \Rev{reference} $Y$ is regarded as the pixel-wise loss, which is defined as: 
\begin{equation}
\begin{aligned}
    & {L}_1\big(G(X), Y\big) \\
    & = \frac{1}{h\times w}\sum\nolimits_{(i,j)=(0,0)}^{(h,w)}\big\| Y(i,j)-G(X)_{(i,j)}\big\|_1,
\end{aligned}
\end{equation}
where $(i,j)$ and $(h,w)$ are the pixel coordinates and the (height, width) of the output, respectively.

\textbf{Style Classification Loss.}
Similar to \cite{choi2018stargan,zhao2018modular}, we define an auxiliary classifier to predict the sketch style of the generated image. 
For any generated image $G(X)$, the style classification loss is defined as:
\begin{equation}
    {L}_\mathrm{sty}(G,S,c)= \mathbb{E}_{X,c} \big[ l_\mathrm{ce}\big( S(G(X)), c\big) \big],
\end{equation}
where $l_\mathrm{ce}(\cdot,\cdot)$ is the cross-entropy loss, $S(\cdot)$ is a CNN that outputs the probability over different styles, and $c$ is the label of a given artist's style. 
Note that we only use the style classification loss in the second stage for the I2S task.

\textbf{Overall Loss.} 
Finally, the overall loss function for the multi-scale discriminators is:
\begin{equation}
    L_{D\sim (D_{parts}, D_{fs})} = \sum\nolimits_i^K -{L}_\mathrm{adv} +  \lambda_\mathrm{fm} {L}_\mathrm{fm},
\end{equation}
and the overall loss function for generator is:
\begin{equation}
\begin{aligned}
    & L_{G\sim (G_{parts}, G_{fs})} \\
    & = {L}_\mathrm{adv} +  \lambda_\mathrm{fm} {L}_\mathrm{fm} + \lambda_1 L_1 + \lambda_\mathrm{per}L_\mathrm{per} + \lambda_\mathrm{sty}L_\mathrm{sty},
\end{aligned}
\end{equation}
where $ \lambda_\mathrm{fm}$, $\lambda_1$, $\lambda_\mathrm{per}$, and $\lambda_\mathrm{sty}$ are hyperparameters that control the importance of the feature matching loss, \Rev{pixel-wise loss, perceptual loss,} and style classification loss, respectively.

\subsection{Implementation Details}\label{sec:ImpDetails}
We use PyTorch~\cite{paszke2017automatic} to implement the baseline \ourmodel. The experiments are conducted on an NVIDIA V100S.

For the I2S task, we set $\lambda_{fm} = 25.0$, $\lambda_{1} = 25.0$, and $\lambda_{per} = 12.5$ to train the model in the facial components synthesis stage, and set $\lambda_{fm} = 100.0$, $\lambda_{1} = 100.0$, $\lambda_{per} = 50.0$, and $\lambda_{sty} = 100.0$ for facial synthesis.
The Adam optimizer~\cite{kingma2014adam} is used for training the whole network. 
The initial learning rates for the generator and discriminator are $2e-4$ and $1e-5$, respectively. 
The other hyperparameters of the optimizer are set to the default values as recommended in PyTorch. 
We set the number of epochs to $50$. All generators and discriminators are trained iteratively.

For the S2I task, we set $\lambda_{fm} = 50.0$, $\lambda_{1} = 50.0$, and $\lambda_{per} = 0.2$ to train the neural network for the facial component synthesis stage, and set $\lambda_{fm} = 100.0$, $\lambda_{1} = 100.0$, and $\lambda_{per} = 0.2$ for facial synchronization. 
We again use the Adam optimizer, with initial learning rates of $2e-4$ for both the generators and discriminators. 
The training strategy is almost the same as that for the I2S task. 
However, we set the number of epochs to $400$,\footnote{Because the S2I task needs to restore more detailed information of the RGB images, more training epochs are needed.} freezing the weights of the facial components synthesis module after $250$ epochs and further training the facial synthesis module for the remaining epochs.

\section{Benchmark}\label{sec:Benchmark}
\Rev{This section provides comprehensive comparisons and analyses of the existing models on \ourdataset, in terms of both the I2S and S2I tasks.}

\subsection{Experimental Settings}
\subsubsection{Evaluation Metrics}
For the I2S task, the most popular facial sketch metric is the structural similarity index metric (SSIM)~\cite{wang2004image,Wang2013IJCV}. 
However, it ignores the perceptual similarity between a prediction and the \Rev{reference}. 
Therefore, we further adopt the recently proposed structure co-occurrence texture (SCOOT) 
metric~\cite{fan2019scoot}, which provides a unified evaluation for both structure and texture. 
For the S2I task, we still adopt the widely used SSIM metric to evaluate the synthesized faces. 
Our evaluation toolbox is available at \url{https://github.com/DengPingFan/FS2KToolbox}.

\subsubsection{Comparison of the Models}
To evaluate the performance on the I2S task and S2I task,
we present the empirical results of $19$ representative approaches and the \ourmodel~baseline.

\begin{table}[t!]
  \centering
  \footnotesize
  \renewcommand{\arraystretch}{1.0}
  \renewcommand{\tabcolsep}{1.0pt}
  \caption{
  Quantitative results of popular models on the I2S task. 
  ``$\uparrow$'' means the higher, the better. {Publ.}: Publication information.
  }\label{tab:I2S-Result}
  \vspace{-5pt}
  \begin{tabular}{c|r|l|cc}
  \hline
   \#  & Model & Publ. & SCOOT$\uparrow$ & SSIM$\uparrow$ \\
   \hline
    1 & DualGAN~\cite{yi2017dualgan} & Yi \etal~ICCV 
    & 0.261 & 0.324 \\
    2 & FPST~\cite{chen2016fast} & Chen \etal~NeurIPSW
    & 0.271 & 0.460 \\
    3 & NST~\cite{gatys2015neural,gatys2016image} &  Gatys \etal~CVPR
    & 0.273 & 0.326 \\
    4 & Pix2pix~\cite{isola2017image} &  Isola \etal~CVPR
    & 0.275 & 0.438 \\
    5 & ACL-GAN~\cite{zhao2020unpaired} &  Zhao \etal~ECCV 
    & 0.278 & 0.404 \\
    6 & WCT~\cite{WCT-NIPS-2017} &   Li \etal~NeurIPS 
    & 0.282 & 0.369 \\ \hline
    7 & AdaIN~\cite{huang2017arbitrary} &  Huang \etal~ICCV 
    & 0.303 & 0.365 \\
    8 & UNIT~\cite{liu2017unsupervised} & Liu \etal~NeurIPS 
    & 0.304 & 0.504 \\
    9 & TSIT~\cite{jiang2020tsit} &  Jiang \etal~ECCV 
    & 0.307 & 0.441 \\
    10 & DRIT++~\cite{DRIT_plus}  &  Lee \etal~IJCV
    & 0.308 & 0.492 \\
    11 & CartoonGAN~\cite{chen2018cartoongan} & Chen \etal~CVPR 
    & 0.319 & 0.400 \\ 
    12 & UGATIT~\cite{kim2019u} & Kim \etal~ICLR
    & 0.323 & 0.457 \\
    13 & NICE-GAN~\cite{Chen_2020_CVPR} &  Chen \etal~CVPR 
    & 0.327 & 0.473 \\
    14 & CycleGAN~\cite{zhu2017unpaired} & Zhu \etal~ICCV
    & 0.348 & 0.435 \\ \hline
    15 & MDAL~\cite{zhang2018multidomain} &  Zhang \etal~TNNLS 
    & 0.355 & 0.466 \\
    16 & UPDG~\cite{YiLLR20} &   Yi \etal~CVPR 
    & 0.364 & 0.471 \\
    17 & Pix2pixHD~\cite{wang2018pix2pixhd} &  Wang \etal~CVPR
    & 0.374 & 0.492 \\
    18 & APDrawing~\cite{yi2019apdrawinggan} & Yi \etal~CVPR
    & 0.375 & 0.464 \\
    19 & DSMAP~\cite{chang2020domain}  & Chang \etal~ECCV
    & 0.378 & 0.493 \\ \hline 
    20 &\textbf{\ourmodel}&  \Rev{Fan \etal~MIR}  &\textbf{0.405}  & \textbf{0.510}\\
    \hline
  \end{tabular}
\end{table}

\subsubsection{Training/Testing Protocols}
All compared methods are selected on three criteria: a) widely regarded technology, b) open-source code, and c) state-of-the-art performance. 
The models are trained and tested on our \ourdataset~with the image sizes specified in their papers. 
If the size setting is not provided in their paper, $512\times512$ is utilized as the default.

\begin{figure*}[t!]
  \centering
  \begin{overpic}[width=\linewidth]{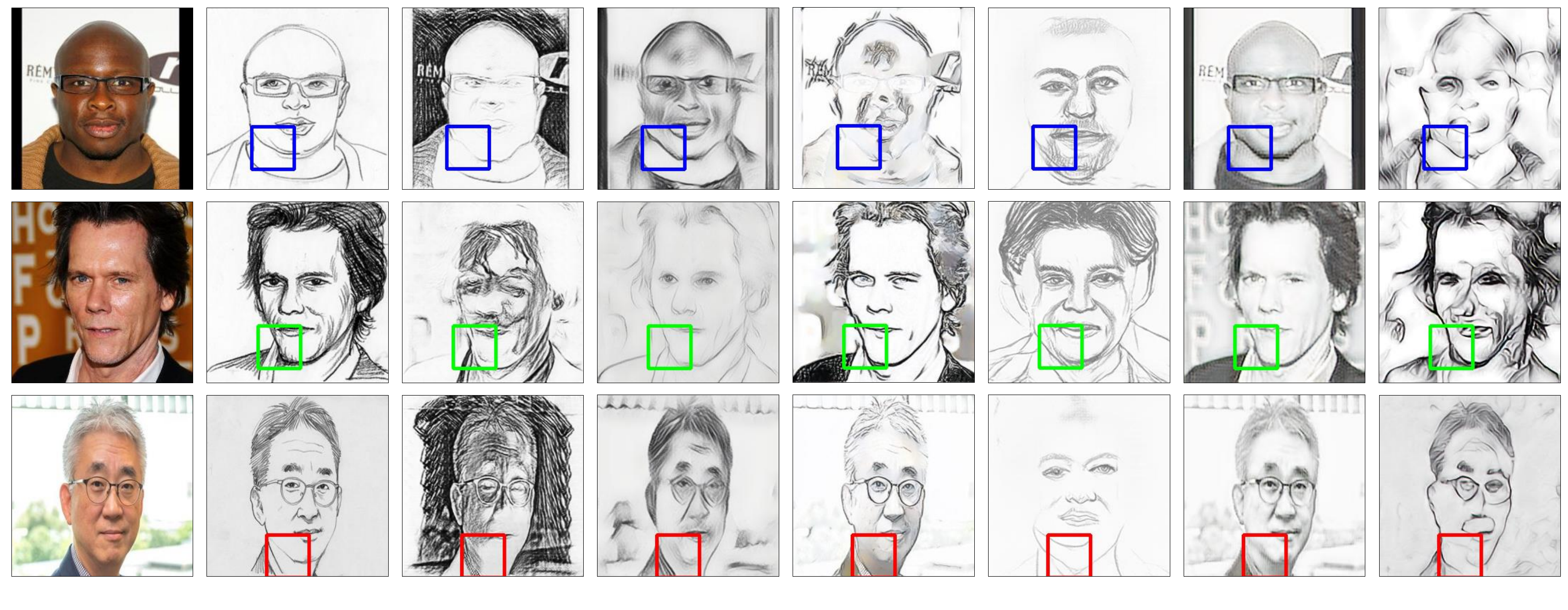}
   \put(3,0){\footnotesize{(a) Input}}
   \put(13.5,0){\footnotesize{(b) Reference}}
   \put(26,0){\footnotesize{(c) DualGAN}}
   \put(40,0){\footnotesize{(d) FPST}}
   \put(53,0){\footnotesize{(e) NST}}
   \put(64,0){\footnotesize{(f) Pix2pix}}
   \put(76,0){\footnotesize{(g) ACL-GAN}}
   \put(90,0){\footnotesize{(h) WCT}}
  \end{overpic}
  \caption{From left to right: input face, reference,
  DualGAN~\cite{yi2017dualgan}, FPST~\cite{chen2016fast}, NST~\cite{gatys2015neural,gatys2016image}, Pix2pix~\cite{isola2017image}, ACL-GAN~\cite{zhao2020unpaired}, and WCT~\cite{WCT-NIPS-2017}.
  We mark the three styles with blue, green, and red boxes for each result. Zoom-in for details.
  }\label{fig:Bench-1}
\end{figure*}

\begin{figure}[t!]
  \centering
  \begin{overpic}[width=\linewidth]{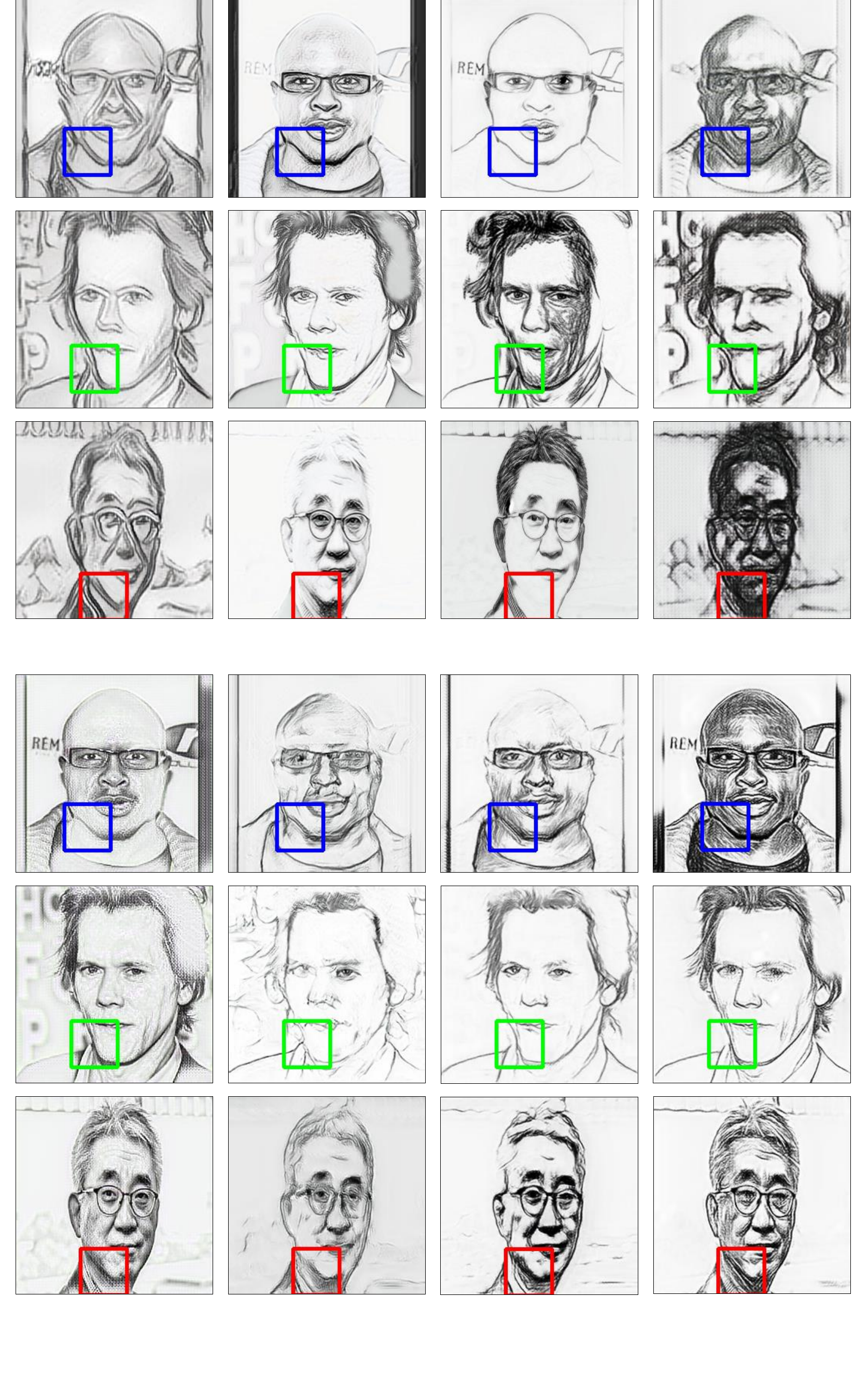}
   \put(3,53){\footnotesize{(a) AdaIN}}
   \put(19,53){\footnotesize{(b) UNIT}}
   \put(35,53){\footnotesize{(c) TSIT}}
   \put(49,53){\footnotesize{(d) DRIT++}}
   \put(-0.5,3.5){\footnotesize{(e) CartoonGAN}}
   \put(17,3.5){\footnotesize{(f) UGATIT}}
   \put(31,3.5){\footnotesize{(g) NICE-GAN}}
   \put(47,3.5){\footnotesize{(h) CycleGAN}}
  \end{overpic}
  \caption{Comparison of AdaIN~\cite{huang2017arbitrary}, UNIT~\cite{liu2017unsupervised}, 
  TSIT~\cite{jiang2020tsit}, DRIT++~\cite{DRIT_plus}, CartoonGAN~\cite{chen2018cartoongan}, UGATIT~\cite{kim2019u}, NICE-GAN~\cite{Chen_2020_CVPR}, and CycleGAN~\cite{zhu2017unpaired}. Their inputs and references are shown in \figref{fig:Bench-1}.
  }\label{fig:Bench-2}
\end{figure}
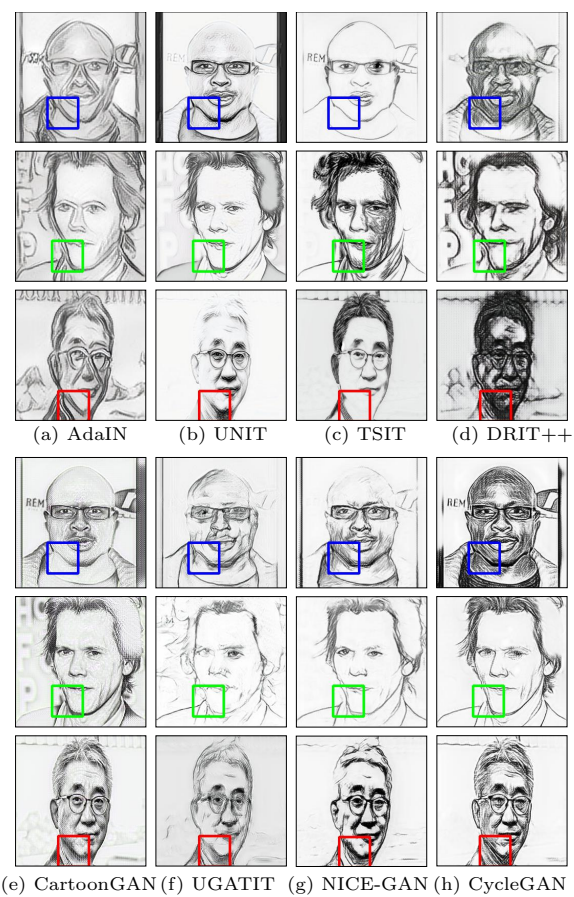

\subsection{Overall Results and Analysis}\subsubsection{I2S Task}
\begin{figure*}[t!]
  \centering
  \begin{overpic}[width=.98\linewidth]{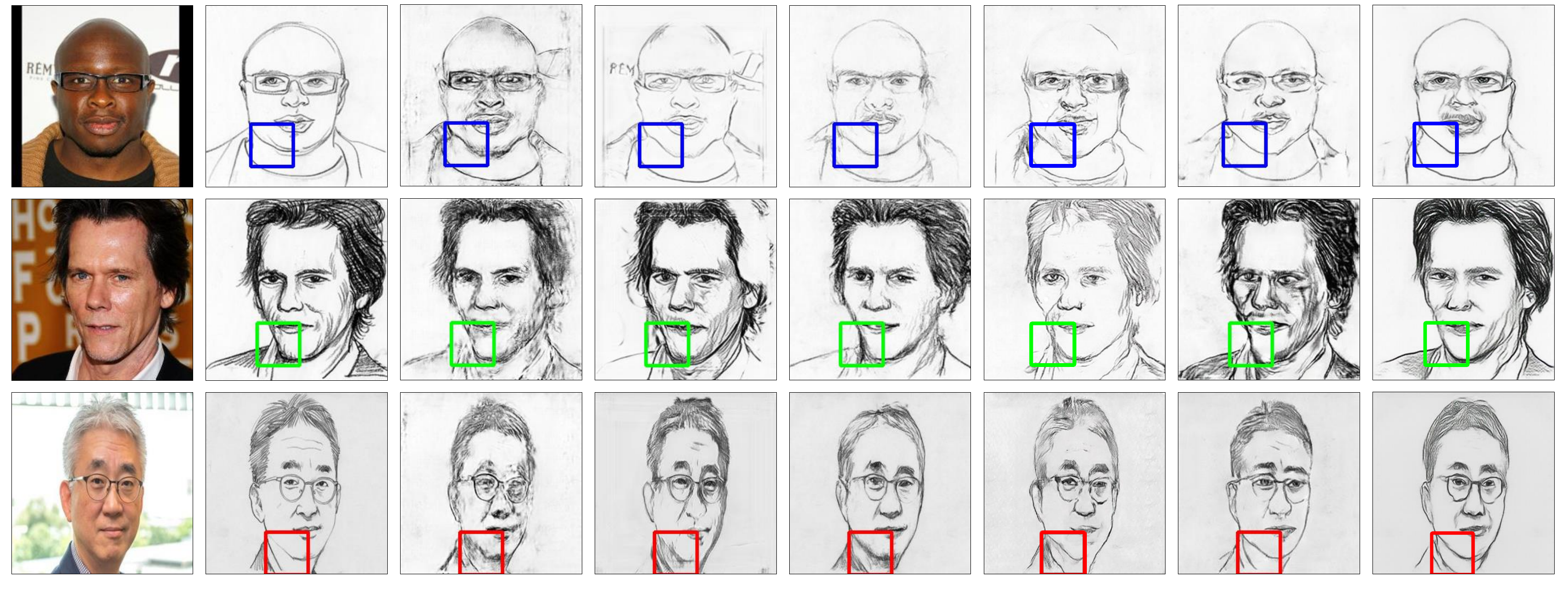}
   \put(2.8,0){\footnotesize{(a) Input}}
   \put(13.5,0){\footnotesize{(b) Reference}}
   \put(27,0){\footnotesize{(c) MDAL}}
   \put(39.5,0){\footnotesize{(d) UPDG}}
   \put(50.5,0){\footnotesize{(e) Pix2pixHD}}
   \put(63,0){\footnotesize{(f) APDrawing}}
   \put(76.5,0){\footnotesize{(g) DSMAP}}
   \put(89,0){\footnotesize{(h) \ourmodel}}
  \end{overpic}
  \caption{Comparison results with 
  MDAL~\cite{zhang2018multidomain}, 
  UPDG~\cite{YiLLR20}, 
  Pix2pixHD~\cite{wang2018pix2pixhd}, 
  APDrawing~\cite{yi2019apdrawinggan}, and 
  DSMAP~\cite{chang2020domain}.
  }\label{fig:Bench-3}
\end{figure*}

We first provide a performance summary of the I2S task regarding both SCOOT and SSIM scores. 
Quantitative results and qualitative comparisons are shown in \tabref{tab:I2S-Result} and \figref{fig:Bench-1}-\ref{fig:Bench-3}, respectively. 
The experimental observations indicate that the \ourmodel~baseline achieves better results.
For further analysis, we divide all compared methods into three categories based on their SCOOT score: 
\begin{itemize}
    \item score $<= 0.3$;
    \item $0.3 <$ score $<= 0.35$;
    \item $0.35 <$ score.
\end{itemize}

\begin{table}[b!]
  \centering
  \footnotesize
  \renewcommand{\arraystretch}{1.0}
  \renewcommand{\tabcolsep}{4.5pt}
  \caption{
  Quantitative results of popular models on the S2I task. 
  ``$\uparrow$'' means the higher, the better.
  }\label{tab:S2I-results}
  \begin{tabular}{c|r|l|c}
  \hline
   \#  & Model & Publication & SSIM$\uparrow$ \\
   \hline
    1 & DualGAN~\cite{yi2017dualgan} & Yi \etal~ICCV 
    & 0.241  \\
    2 & WCT~\cite{WCT-NIPS-2017} &  Li \etal~NeurIPS
    & 0.311  \\
    3 & ACL-GAN~\cite{zhao2020unpaired} &  Zhao \etal~ECCV 
    & 0.314 \\    
    4 & TSIT~\cite{jiang2020tsit} &  Jiang \etal~ECCV 
    & 0.316 \\
    5 & UGATIT~\cite{kim2019u} & Kim \etal~ICLR 
    & 0.317  \\
    6 & NST~\cite{gatys2015neural,gatys2016image} &  Gatys \etal~CVPR 
    & 0.335 \\
    7 & CycleGAN~\cite{zhu2017unpaired} & Zhu \etal~ICCV 
    & 0.339 \\
    8 & Pix2pix~\cite{isola2017image} &  Isola \etal~CVPR 
    & 0.346  \\
    9 & SPADE~\cite{park2019semantic} & Park \etal~CVPR 
    & 0.361 \\
    10 & UNIT~\cite{liu2017unsupervised} & Liu \etal~NeurIPS 
    & 0.362 \\
    11 & AdaIN~\cite{huang2017arbitrary} &  Huang \etal~ICCV 
    & 0.373 \\
    12 & DRIT++~\cite{DRIT_plus}  &  Lee \etal~IJCV 
    & 0.381 \\
    13 & FNS~\cite{johnson2016perceptual} &  Johnson \etal~ECCV 
    & 0.391 \\
    14 & NICE-GAN~\cite{Chen_2020_CVPR} &  Chen \etal~CVPR 
    & 0.397 \\
    15 & FPST~\cite{chen2016fast} & Chen \etal~NeurIPSW 
    & 0.400 \\
    16 & pSp~\cite{richardson2020encoding} & Richardson \etal~CVPR 
    & 0.428  \\
    17 & Pix2pixHD~\cite{wang2018pix2pixhd} & Wang \etal~CVPR 
    & 0.433  \\
    18 & DSMAP~\cite{chang2020domain}  & Chang \etal~ECCV 
    & 0.471 \\
    19 & DeepPS~\cite{yang2020deep} &  Yang \etal~ECCV 
    & 0.487 \\
    \hline
    20    &\textbf{\ourmodel}&  \Rev{Fan \etal~MIR} &\textbf{0.503} \\
    \hline
  \end{tabular}
\end{table}

\textbf{Analysis.} 
Methods in the first group achieve a SCOOT below $0.3$. 
These include DualGAN~\cite{yi2017dualgan}, FPST~\cite{chen2016fast}, 
NST~\cite{gatys2015neural,gatys2016image}, Pix2pix~\cite{isola2017image}, 
ACL-GAN~\cite{zhao2020unpaired}, and WCT~\cite{WCT-NIPS-2017}. 
As shown in \figref{fig:Bench-1}, DualGAN, NST, and WCT suffer from structural distortion, where many local facial details are lost.
The images produced by the DualGAN are poor, and it is challenging to detect facial components in them. This explains why it has lower SSIM and SCOOT scores. In addition, compared with other results, Pix2pix and FPST generate blurred results.
ACL-GAN seems to achieve satisfactory results in visual appeal, yielding a higher SSIM score. However, ACL-GAN reproduces the original facial structure almost exactly, lacking artistic style.

The second group includes AdaIN~\cite{huang2017arbitrary}, UNIT~\cite{liu2017unsupervised}, TSIT~\cite{jiang2020tsit}, 
DRIT++~\cite{DRIT_plus}, CartoonGAN~\cite{chen2018cartoongan}, UGATIT~\cite{kim2019u}, NICE-GAN~\cite{Chen_2020_CVPR}, and CycleGAN~\cite{zhu2017unpaired}, whose SCOOT scores range from $0.3$ to $0.35$.
As shown in \figref{fig:Bench-2}, the synthesized sketch images are better in terms of structure-preservation compared to the first group. 
However, except for AdaIN, all models are thrown off by the complex backgrounds (see the hair region in the second row). 
Besides, the results of CartoonGAN seem to alter the color of the input images, leading to lower SSIM scores. 

\begin{figure*}[t!]
  \centering
  \begin{overpic}[width=.98\linewidth]{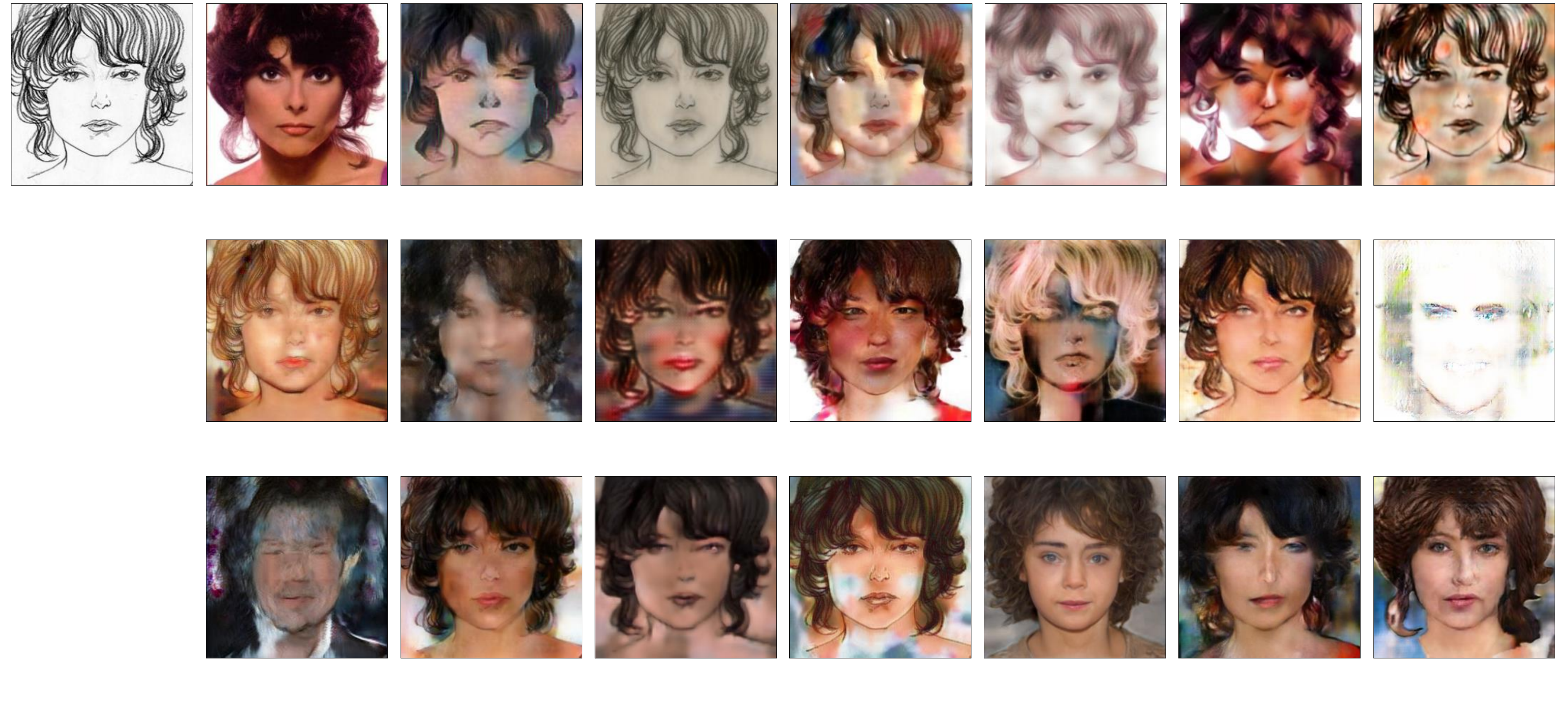}
   \put(2.5,32){\footnotesize{(a) Input}}
   \put(13.5,32){\footnotesize{(b) Reference}}
   \put(27,32){\footnotesize{(c) AdaIN}}
   \put(40,32){\footnotesize{(d) FNS}}
   \put(53,32){\footnotesize{(e) NST}}
   \put(65,32){\footnotesize{(f) FPST}}
   \put(77,32){\footnotesize{(g) WCT}}
   \put(87,32){\footnotesize{(h) ACL-GAN}}
  
   \put(13,17){\footnotesize{(i) CycleGAN}}
   \put(26.5,17){\footnotesize{(j) DeepPS}}
   \put(39,17){\footnotesize{(k) DRIT++}}
   \put(51.5,17){\footnotesize{(l) DSMAP}}
   \put(63,17){\footnotesize{(m) DualGAN}}
   \put(75,17){\footnotesize{(n) NICE-GAN}}
   \put(89,17){\footnotesize{(o) Pix2pix}}
   
   \put(14.5,2){\footnotesize{(p) SPADE}}
   \put(27.5,2){\footnotesize{(q) TSIT}}
   \put(38.5,2){\footnotesize{(r) UGATIT}}
   \put(51.5,2){\footnotesize{(s) UNIT}}
   \put(65.5,2){\footnotesize{(t) pSp}}
   \put(74.5,2){\footnotesize{(u) Pix2pixHD}}
   \put(89,2){\footnotesize{(v) \ourmodel}}
  \end{overpic}
  \vspace{-10pt}
  \caption{We select 19 classical models, including AdaIN~\cite{huang2017arbitrary}, FNS~\cite{johnson2016perceptual}, FPST~\cite{chen2016fast}, WCT~\cite{WCT-NIPS-2017}, 
  ACL-GAN~\cite{zhao2020unpaired}, CycleGAN~\cite{zhu2017unpaired}, DeepPS~\cite{yang2020deep}, DRIT++~\cite{DRIT_plus}, DSMAP~\cite{chang2020domain}, DualGAN~\cite{yi2017dualgan}, NICE-GAN~\cite{Chen_2020_CVPR},  Pix2pix~\cite{isola2017image}, SPADE~\cite{park2019semantic}, TSIT~\cite{jiang2020tsit}, UGATIT~\cite{kim2019u}, UNIT~\cite{liu2017unsupervised}, 
  pSp~\cite{richardson2020encoding}, and Pix2pixHD~\cite{wang2018pix2pixhd}, for qualitative comparison.
  }\label{fig:Bench-4}
\end{figure*}

MDAL~\cite{zhang2018multidomain}, UPDG~\cite{YiLLR20}, Pix2pixHD~\cite{wang2018pix2pixhd}, APDrawing~\cite{yi2019apdrawinggan}, DSMAP~\cite{chang2020domain}, and our \ourmodel~are categorized into the third group, which can generate sketches without distortion or losing too much of the global details. However, UPDG and APDrawing miss some details in the hair region, leading to poor visual effects. APDrawing introduces many extra strokes, especially for the first sketch style. 
Besides, APDrawing usually results in a lack and distortion of the local structure, as seen in the hair region.
Meanwhile, the sketches generated by UPDG have better style elements, but the model cannot handle complex backgrounds.
Pix2pixHD generates relatively good sketches with global structure and clean background, but it does not generate the best facial components. 
For example, in \figref{fig:Bench-3}-e, the region around the eyes is unclear, and many details are lost. Take the third row, for instance; the eyeglasses are partially lost, while the eyeball is entirely black. 
We further observe that DSMAP and MDAL tend to achieve better sketch images but with distortions in local facial information.
Finally, the baseline can synthesize high-quality sketches that focus on the global structure and local details while considering diverse styles.
Moreover, as shown in the highlighted boxes (with green, blue and red), we find that the outputs of the \ourmodel~are more similar to the \Rev{reference} compared to other state-of-the-arts methods.

\subsubsection{S2I Task}
We report our experimental results in \tabref{tab:S2I-results} and \figref{fig:Bench-4}. 
We find that \ourmodel~achieves the best results on our challenging \ourdataset~compared to the existing state-of-the-art models.

\textbf{Analysis.} 
\Rev{As seen} in \figref{fig:Bench-4}, we observe that most compared methods are unable to successfully recover accurate images, revealing that the S2I task is more complicated than I2S. 
We argue that this is because the sketches are highly abstract, and the loss of valuable information makes it difficult for neural networks to restore the original image. 
We also observe that the high-resolution models, such as Pix2pixHD and \ourmodel, tend to output more visually appealing results. 

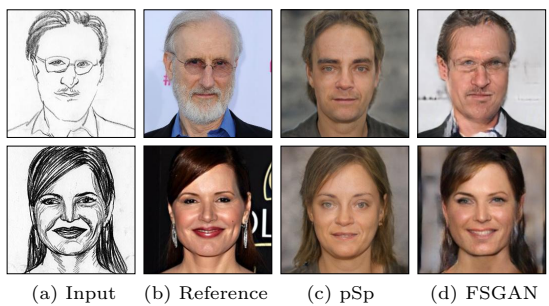
\begin{figure}[t!]
  \centering
  \begin{overpic}[width=.98\linewidth]{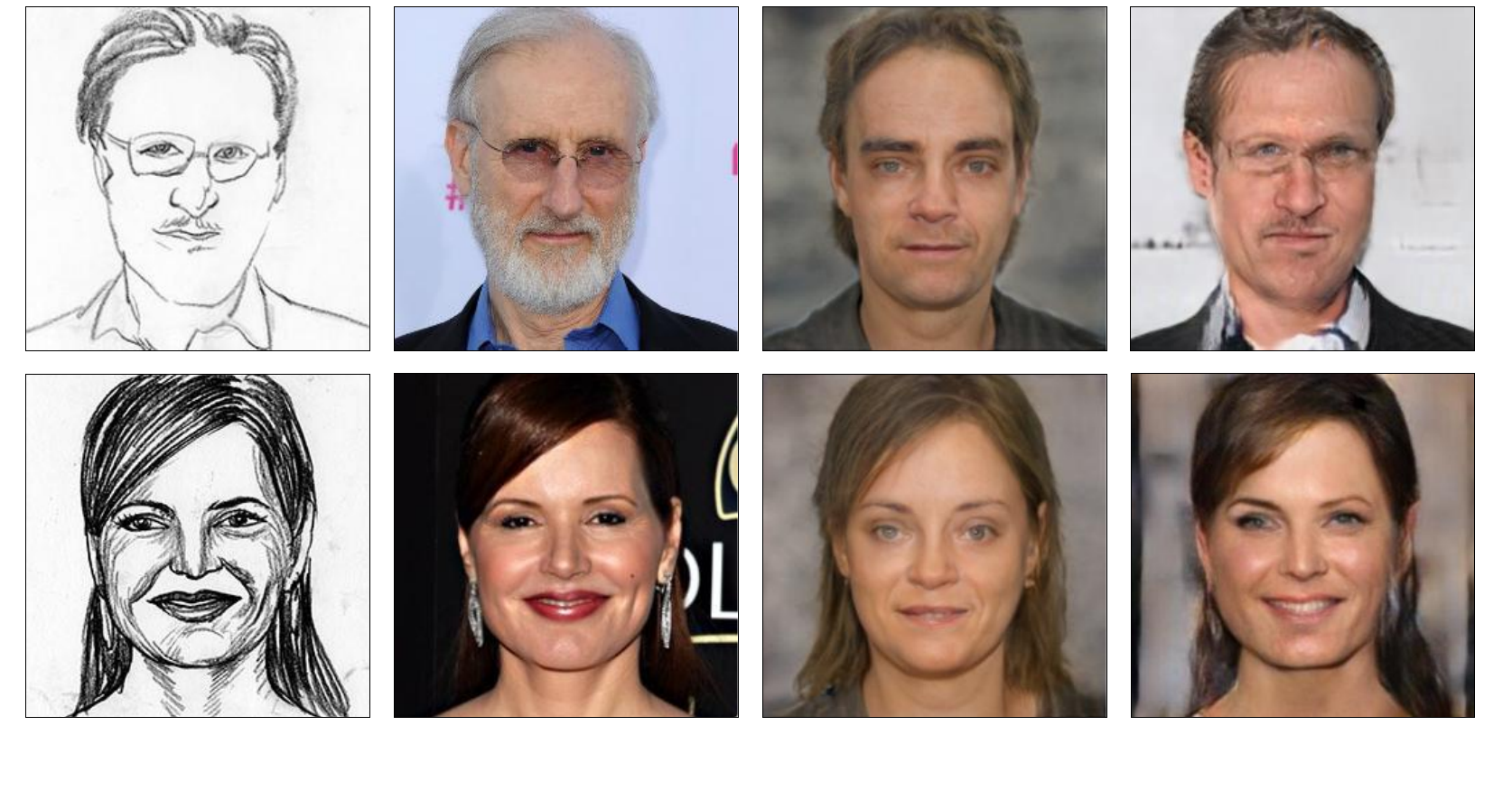}
   \put(6,1){\footnotesize{(a) Input}}
   \put(26,1){\footnotesize{(b) Reference}}
   \put(55,1){\footnotesize{(c) pSp}}
   \put(77,1){\footnotesize{(d) \ourmodel}}
  \end{overpic}
  \caption{Visual diversity of the data generated for S2I task. 
  }\label{fig:Bench-5}
\end{figure}

The results presented in \figref{fig:Bench-4} show that FNS and FPST fail to transfer the sketches into colored images.
SPADE and Pix2pix generate poor results with facial outlines (\eg, Pix2pix) or black background (\eg, SPADE).
Five models (\ie, NST, WCT, DeepPS, DSMAP, and UNIT) produce noise patches in salient regions, which corrupt the global facial structure.
Meanwhile, AdaIN, ACL-GAN, DualGAN, and UGATIT perform better than the models mentioned above, resulting in unrealistic cartoon-style images.    
Only CycleGAN, NICE-GAN, TSIT, pSp, and Pix2pixHD overcome various challenges and achieve good results in terms of facial completeness. 
In particular, the eye regions from Pix2pixHD~\cite{wang2018pix2pixhd} and pSp~\cite{richardson2020encoding} are better than those from the other models.
However, compared with the results of the \ourmodel, the facial features of Pix2pixHD are relatively inferior because a pixel-wise rather than block-wise strategy learns.
Although pSp~\cite{richardson2020encoding} can generate high-quality results, its results lack diversity compared with the \ourmodel~baseline. 
For example, pSp generates similar facial expressions under two different sketch styles, while the baseline can synthesize diverse contents, as shown in \figref{fig:Bench-5}. 

\subsection{Attribute-Based Analysis}
\subsubsection{SCOOT Metric Results}
To provide a deeper understanding of the models, we present an attribute-based performance evaluation in \tabref{tab:Image-Sketch-Attri-Result-SCOOT}. 

\begin{table*}[t!]
  \centering
  \footnotesize
  \renewcommand{\arraystretch}{1.0}
  \renewcommand{\tabcolsep}{1.0pt}
  \caption{
  Comparison of 19 state-of-the-art models in terms of attribute-based performance on the I2S task. 
  }\label{tab:Image-Sketch-Attri-Result-SCOOT}
  \begin{tabular}{r|cc|cccc|cc|cc|cc|cc|ccc}
  \hline
   \multirow{2}*{Model} & \multicolumn{17}{c}{SCOOT$\uparrow$} \\
   \cline{2-18}
   &w/ H &w/o H &H(b) &H(bl) &H(r) &H(g) & M &F & w/ E & w/o E & w/ S &w/o S &w/ F &w/o F &S1 &S2 &S3\\ \hline
   DualGAN~\cite{yi2017dualgan} & 0.260 & 0.279 & 0.250 & 0.267 & 0.216 & 0.279 & 0.275 & 0.240 & 0.239 & 0.266 & 0.255 & 0.271 & 0.261 & 0.262 & 0.298 & 0.194 & 0.319 \\
   FPST~\cite{chen2016fast} & 0.269 & 0.304 & 0.254 & 0.294 & 0.214 & 0.304 & 0.288 & 0.245 & 0.246 & 0.276 & 0.262 & 0.286 & 0.269 & 0.278 & 0.329 & 0.168 & 0.332 \\
   NST~\cite{gatys2015neural,gatys2016image} & 0.272 & 0.283 & 0.268 & 0.287 & 0.236 & 0.283 & 0.280 & 0.262 & 0.258 & 0.276 & 0.268 & 0.282 & 0.272 & 0.276 & 0.310 & 0.205 & 0.332 \\ 
   Pix2pix~\cite{isola2017image} & 0.272 & 0.335 & 0.255 & 0.300 & 0.217 & 0.335 & 0.298 & 0.240 & 0.250 & 0.281 & 0.267 & 0.290 & 0.276 & 0.272 & 0.333 & 0.178 & 0.302 \\
   ACL-GAN~\cite{zhao2020unpaired} & 0.276 & 0.309 & 0.265 & 0.298 & 0.226 & 0.309 & 0.292 & 0.256 & 0.254 & 0.283 & 0.270 & 0.291 & 0.276 & 0.284 & 0.330 & 0.183 & 0.355 \\ 
   WCT~\cite{WCT-NIPS-2017} & 0.281 & 0.315 & 0.271 & 0.302 & 0.229 & 0.315 & 0.296 & 0.261 & 0.262 & 0.287 & 0.277 & 0.292 & 0.281 & 0.290 & 0.332 & 0.195 & 0.346 \\
   AdaIN~\cite{huang2017arbitrary} & 0.303 & 0.295 & 0.307 & 0.317 & 0.258 & 0.295 & 0.306 & 0.298 & 0.283 & 0.307 & 0.298 & 0.310 & 0.300 & 0.314 & 0.348 & 0.215 & 0.419 \\
   UNIT~\cite{liu2017unsupervised} & 0.301 & 0.364 & 0.292 & 0.328 & 0.225 & 0.364 & 0.330 & 0.265 & 0.261 & 0.313 & 0.293 & 0.324 & 0.301 & 0.319 & 0.376 & 0.175 & 0.411 \\
   TSIT~\cite{jiang2020tsit} & 0.307 & 0.307 & 0.308 & 0.320 & 0.259 & 0.307 & 0.320 & 0.288 & 0.283 & 0.313 & 0.300 & 0.320 & 0.306 & 0.316 & 0.359 & 0.208 & 0.432 \\ 
   DRIT++~\cite{DRIT_plus}  & 0.305 & 0.348 & 0.291 & 0.336 & 0.248 & 0.348 & 0.329 & 0.276 & 0.279 & 0.314 & 0.299 & 0.323 & 0.305 & 0.323 & 0.380 & 0.181 & 0.378 \\
   CartoonGAN~\cite{chen2018cartoongan} & 0.319 & 0.318 & 0.320 & 0.337 & 0.262 & 0.318 & 0.329 & 0.304 & 0.291 & 0.325 & 0.314 & 0.329 & 0.317 & 0.332 & 0.382 & 0.204 & 0.428 \\
   UGATIT~\cite{kim2019u} & 0.321 & 0.365 & 0.315 & 0.347 & 0.265 & 0.365 & 0.339 & 0.298 & 0.298 & 0.328 & 0.314 & 0.338 & 0.322 & 0.325 & 0.391 & 0.204 & 0.400 \\
   NICE-GAN~\cite{Chen_2020_CVPR} & 0.325 & 0.355 & 0.320 & 0.357 & 0.262 & 0.355 & 0.342 & 0.303 & 0.302 & 0.332 & 0.317 & 0.343 & 0.325 & 0.333 & 0.398 & 0.201 & 0.401 \\
   CycleGAN~\cite{zhu2017unpaired} & 0.348 & 0.343 & 0.358 & 0.362 & 0.287 & 0.343 & 0.351 & 0.343 & 0.326 & 0.353 & 0.341 & 0.360 & 0.346 & 0.357 & 0.397 & 0.252 & 0.483 \\
   MDAL~\cite{zhang2018multidomain} & 0.354 & 0.363 & 0.348 & 0.380 & 0.292 & 0.363 & 0.369 & 0.333 & 0.329 & 0.360 & 0.345 & 0.372 & 0.352 & 0.365 & 0.436 & 0.211 & 0.446 \\
   UPDG~\cite{YiLLR20} & 0.362 & 0.411 & 0.349 & 0.390 & 0.290 & 0.411 & 0.390 & 0.325 & 0.336 & 0.371 & 0.356 & 0.379 & 0.363 & 0.370 & 0.423 & 0.259 & 0.448 \\
   APDrawing~\cite{yi2019apdrawinggan} & 0.374 & 0.395 & 0.372 & 0.399 & 0.322 & 0.395 & 0.380 & 0.369 & 0.356 & 0.380 & 0.370 & 0.385 & 0.373 & 0.390 & 0.456 & 0.227 & 0.524 \\
   Pix2pixHD~\cite{wang2018pix2pixhd} & 0.374 & 0.392 & 0.365 & 0.403 & 0.307 & 0.385 & 0.392 & 0.351 & 0.343 & 0.378 & 0.371 & 0.392 & 0.371 & 0.381 & 0.462 & 0.212 & 0.508 \\ 
   DSMAP~\cite{chang2020domain}  & 0.375 & 0.431 & 0.357 & 0.405 & 0.322 & 0.431 & 0.400 & 0.343 & 0.354 & 0.383 & 0.369 & 0.393 & 0.377 & 0.381 & 0.437 & \textbf{0.276} & 0.423 \\
   \hline
   \textbf{\ourmodel} & \textbf{0.403} & \textbf{0.435} & \textbf{0.389} & \textbf{0.435} & \textbf{0.335} & \textbf{0.435} & \textbf{0.423} & \textbf{0.377} & \textbf{0.381} & \textbf{0.410} & \textbf{0.395} & \textbf{0.422} & \textbf{0.403} & \textbf{0.414} & \textbf{0.481} & 0.268 & \textbf{0.509} \\
   \hline    
  \end{tabular}
  \begin{tablenotes}
      Here, w/ H = hair visible, w/o H = hair invisible, H(b) = brown hair, H(bl) = black hair, H(r) = red hair, H(g) = golden hair, M = male, F = female, w/ E = with earring, w/o E = without earring, w/ S = with smile, w/o S = without smile, w/ F = frontal face, w/o F = non-frontal face, S1 = style1, S2 = style2, and S3 = style3.
  \end{tablenotes}
\end{table*}

\begin{table*}[t!]
  \centering
  \footnotesize
  \renewcommand{\arraystretch}{1.0}
  \renewcommand{\tabcolsep}{1.0pt}
  \caption{
  Comparison of 19 top models in terms of attribute-based performance on the I2S task.
  }\label{tab:Image-Sketch-Attri-Result-SSIM}
  \begin{tabular}{r|cc|cccc|cc|cc|cc|cc|ccc}
  \hline
   \multirow{2}*{Model} & \multicolumn{17}{c}{SSIM$\uparrow$} \\
   \cline{2-18}
   &w/ H &w/o H &H(b) &H(bl) &H(r) &H(g) & M &F & w/ E & w/o E & w/ S &w/o S &w/ F &w/o F &S1 &S2 &S3\\ 
   \hline
   DualGAN~\cite{yi2017dualgan} & 0.320 & 0.393 & 0.310 & 0.342 & 0.276 & 0.393 & 0.352 & 0.282 & 0.292 & 0.331 & 0.313 & 0.343 & 0.318 & 0.354 & 0.364 & 0.247 & 0.424 \\
   FPST~\cite{chen2016fast} & 0.459 & 0.481 & 0.442 & 0.492 & 0.383 & 0.481 & 0.492 & 0.411 & 0.416 & 0.469 & 0.448 & 0.481 & 0.455 & 0.486 & 0.517 & 0.351 & 0.597 \\
   NST~\cite{gatys2015neural,gatys2016image} & 0.325 & 0.347 & 0.317 & 0.349 & 0.256 & 0.347 & 0.339 & 0.306 & 0.305 & 0.330 & 0.316 & 0.344 & 0.324 & 0.338 & 0.372 & 0.241 & 0.417 \\
   Pix2pix~\cite{isola2017image} & 0.434 & 0.526 & 0.410 & 0.470 & 0.332 & 0.526 & 0.478 & 0.377 & 0.391 & 0.449 & 0.425 & 0.461 & 0.438 & 0.439 & 0.503 & 0.319 & 0.558 \\
   ACL-GAN~\cite{zhao2020unpaired} & 0.402 & 0.432 & 0.392 & 0.430 & 0.334 & 0.432 & 0.427 & 0.369 & 0.363 & 0.413 & 0.393 & 0.423 & 0.398 & 0.434 & 0.445 & 0.316 & 0.583 \\
   WCT~\cite{WCT-NIPS-2017} & 0.368 & 0.389 & 0.368 & 0.387 & 0.316 & 0.389 & 0.389 & 0.339 & 0.334 & 0.377 & 0.362 & 0.381 & 0.367 & 0.380 & 0.407 & 0.297 & 0.461 \\
   AdaIN~\cite{huang2017arbitrary} & 0.364 & 0.367 & 0.364 & 0.382 & 0.319 & 0.367 & 0.378 & 0.343 & 0.340 & 0.370 & 0.359 & 0.375 & 0.362 & 0.379 & 0.399 & 0.297 & 0.460 \\
   UNIT~\cite{liu2017unsupervised} & 0.501 & 0.556 & 0.488 & 0.528 & 0.421 & 0.556 & 0.539 & 0.450 & 0.460 & 0.514 & 0.492 & 0.526 & 0.498 & 0.532 & 0.563 & 0.395 & 0.616 \\
   TSIT~\cite{jiang2020tsit} & 0.439 & 0.465 & 0.430 & 0.461 & 0.371 & 0.465 & 0.465 & 0.404 & 0.408 & 0.448 & 0.431 & 0.458 & 0.435 & 0.468 & 0.485 & 0.351 & 0.587 \\
   DRIT++~\cite{DRIT_plus}  & 0.490 & 0.534 & 0.479 & 0.519 & 0.411 & 0.534 & 0.524 & 0.444 & 0.451 & 0.501 & 0.480 & 0.512 & 0.487 & 0.515 & 0.547 & 0.387 & 0.617 \\
   CartoonGAN~\cite{chen2018cartoongan} & 0.399 & 0.420 & 0.397 & 0.421 & 0.345 & 0.420 & 0.419 & 0.372 & 0.368 & 0.407 & 0.392 & 0.416 & 0.395 & 0.425 & 0.438 & 0.321 & 0.552 \\
   UGATIT~\cite{kim2019u} & 0.455 & 0.497 & 0.445 & 0.476 & 0.386 & 0.497 & 0.489 & 0.409 & 0.416 & 0.466 & 0.447 & 0.476 & 0.451 & 0.491 & 0.499 & 0.373 & 0.593 \\
   NICE-GAN~\cite{Chen_2020_CVPR} & 0.472 & 0.497 & 0.463 & 0.492 & 0.398 & 0.497 & 0.505 & 0.424 & 0.429 & 0.483 & 0.464 & 0.490 & 0.468 & 0.498 & 0.518 & 0.384 & 0.603 \\
   CycleGAN~\cite{zhu2017unpaired} & 0.433 & 0.461 & 0.429 & 0.455 & 0.374 & 0.461 & 0.460 & 0.395 & 0.401 & 0.442 & 0.425 & 0.452 & 0.429 & 0.463 & 0.471 & 0.358 & 0.580 \\
   MDAL~\cite{zhang2018multidomain} & 0.465 & 0.487 & 0.457 & 0.491 & 0.399 & 0.487 & 0.496 & 0.420 & 0.426 & 0.475 & 0.458 & 0.481 & 0.462 & 0.488 & 0.506 & 0.386 & 0.593 \\
   UPDG~\cite{YiLLR20} & 0.468 & 0.507 & 0.456 & 0.500 & 0.391 & 0.507 & 0.501 & 0.424 & 0.431 & 0.479 & 0.459 & 0.493 & 0.465 & 0.501 & 0.534 & 0.355 & 0.584 \\
   APDrawing~\cite{yi2019apdrawinggan} & 0.461 & 0.522 & 0.441 & 0.497 & 0.373 & 0.522 & 0.504 & 0.402 & 0.419 & 0.473 & 0.452 & 0.484 & 0.458 & 0.492 & 0.512 & 0.371 & 0.582 \\
   Pix2pixHD~\cite{wang2018pix2pixhd} & 0.492 & 0.552 & 0.473 & 0.523 & 0.419 & 0.546 & 0.531 & 0.431 & 0.457 & 0.505 & 0.481 & 0.513 & 0.488 & 0.524 & 0.537 & 0.402 & 0.618 \\
   DSMAP~\cite{chang2020domain}  & 0.490 & 0.551 & 0.472 & 0.527 & 0.405 & 0.551 & 0.532 & 0.433 & 0.447 & 0.503 & 0.481 & 0.515 & 0.488 & 0.518 & 0.557 & 0.373 & 0.622 \\
   \hline
   \textbf{\ourmodel} & \textbf{0.507} & \textbf{0.565} & \textbf{0.491} & \textbf{0.539} & \textbf{0.424} & \textbf{0.565} & \textbf{0.549} & \textbf{0.451} & \textbf{0.466} & \textbf{0.520} & \textbf{0.498} & \textbf{0.531} & \textbf{0.505} & \textbf{0.534} & \textbf{0.568} & \textbf{0.403} & \textbf{0.629} \\
   \hline     
  \end{tabular}
\end{table*}

\textbf{Analysis.} 
Hair is one of the dominant features of the head. 
In \tabref{tab:Image-Sketch-Attri-Result-SCOOT}, we find that most models achieve slightly better or comparable performance on images without hair than with, except for three models, such as AdaIN, CartoonGAN, and CycleGAN. 
Meanwhile, we find that red and black hair are the most challenging and easiest to detect/reconstruct, respectively.  
We argue that this is because images with red and black hair make up the lowest and largest ($>$40\%)  proportion of all data, respectively. 
Thus, the models are unfamiliar/familiar with these attributes. 

In addition, we also notice that females (F) are more challenging than males (M) for almost all models since women usually have various accessories and hairstyles. 
For example, the models perform worse on images with earrings (w/ E) than those without earrings.
Additionally, facial images with smiles are more challenging than those without smiles. 
Interestingly, existing models achieve diverse performance irrespective of the color of hair (\eg, H(b), H(bl), H(r), and H(g)). 
Finally, compared to style $1$ (simple lines) and style $3$ (\ie, repeated wispy details), we see that style $2$ (long strokes) is the most challenging for all models. 

\subsubsection{SSIM Metric Results}
In addition to the SCOOT metric, we also provide the SSIM metric for the I2S task in \tabref{tab:Image-Sketch-Attri-Result-SSIM}. 

\textbf{Analysis.}
We find that the overall performance tends to be similar to the SCOOT metric results in several key attributes, such as hair, gender, accessories, and style. 
We note that the performance on ``w/ F'' is 
lower than on ``w/o F'', as shown in \tabref{tab:Image-Sketch-Attri-Result-SSIM}. 
One possible reason is that frontal faces preserve more structural features than non-frontal faces. Therefore, in the I2S task, images with attributes such as ``w/ F'' are more challenging than ``w/o F''. 

\begin{figure}[t!]
  \centering
  \begin{overpic}[width=.98\linewidth]{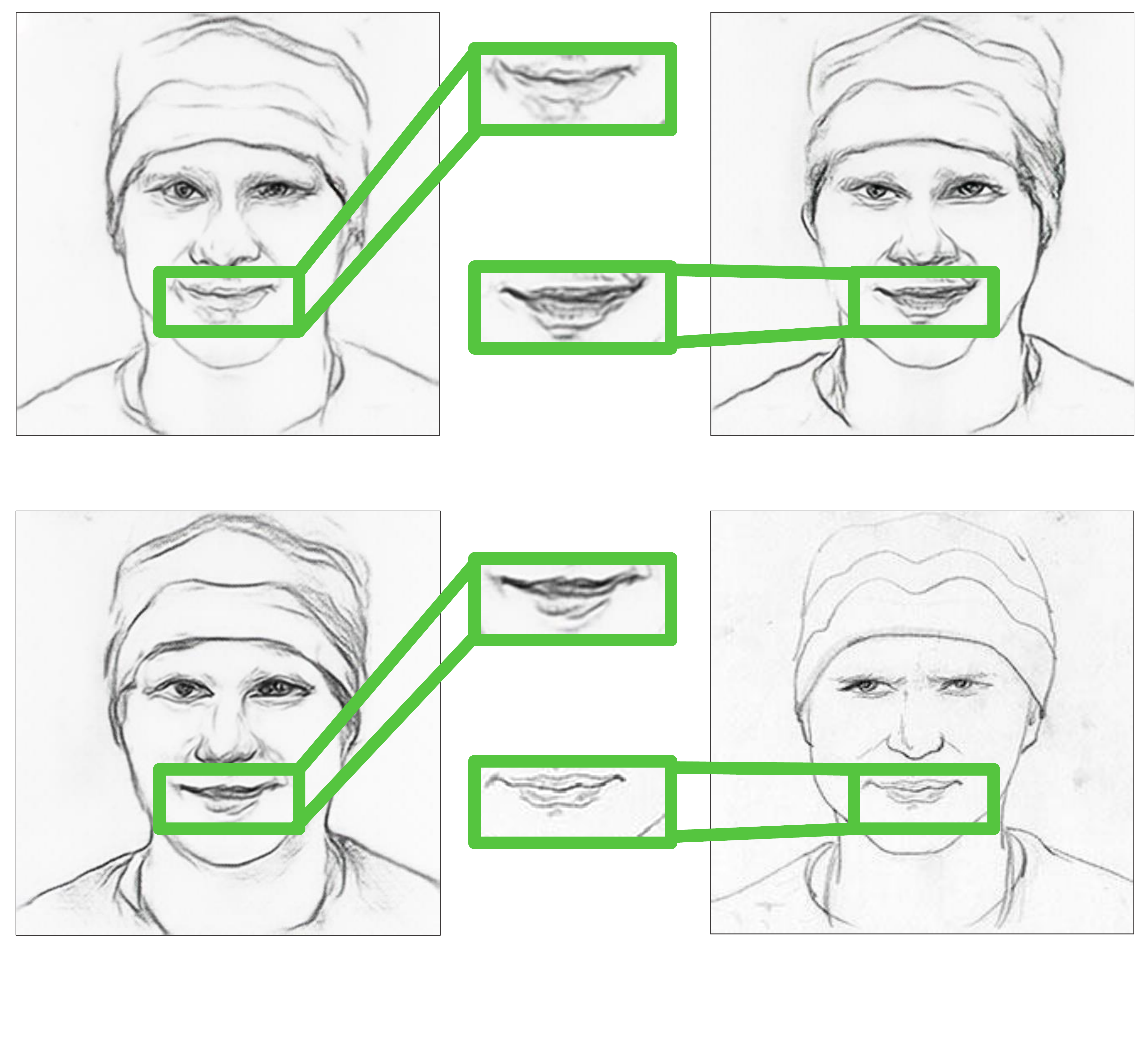}
   \put(-0.5,49.5){(a) w/o Multi-Patch}
   \put(70,49.5){(b) \ourmodel}
   \put(-0.5,5){(c) w/o Style Vector}
   \put(67,5){(d) Reference}
  \end{overpic}
  \vspace{-5pt}
  \caption{Ablation study for the I2S task.
  }\label{fig:Ablation-I2S}
\end{figure}
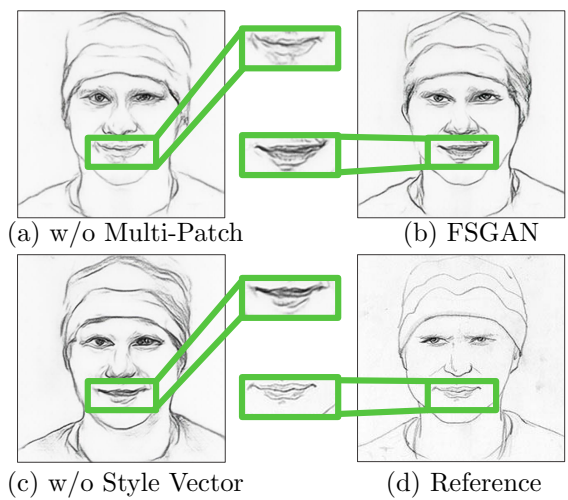

\subsection{Ablation Study}\label{sec:ablation}
This section provides a detailed analysis of \ourmodel~on the proposed \ourdataset~dataset. 
Unlike most existing facial synthesis models \cite{wang2018pix2pixhd}, our model has a two-stage GAN architecture for both I2S and S2I tasks. 
Besides, a sketch style vector is introduced to enable diversified style synthesis in the second stage of the I2S task. 
Therefore, the ablation studies on the I2S task are conducted on the following two key components: (1) the facial components synthesis stage and (2) the style vector assisted generation. Note that we adopt the same hyperparameters described in \secref{sec:ImpDetails} during our ablation experiments. 

\tabref{tab:ablation_i2s} shows the ablation results for the I2S task. 
We find that the facial components synthesis stage increases the SCOOT and SSIM scores by $1.31\%$ (relative) and $2.67\%$, respectively, while the style vector increases them by $6.30\%$ and $4.72\%$. 
As illustrated in \figref{fig:Ablation-I2S}, without the multi-patch strategy, the lines in the synthesized lips are often missing structural details. 
Meanwhile, with the multi-patch stage, the lines become smoother. 
Moreover, the synthesized drawings are messier without the style vector component and may introduce shadows in the lip regions.

\begin{figure}[t!]
  \centering
  \begin{overpic}[width=.98\linewidth]{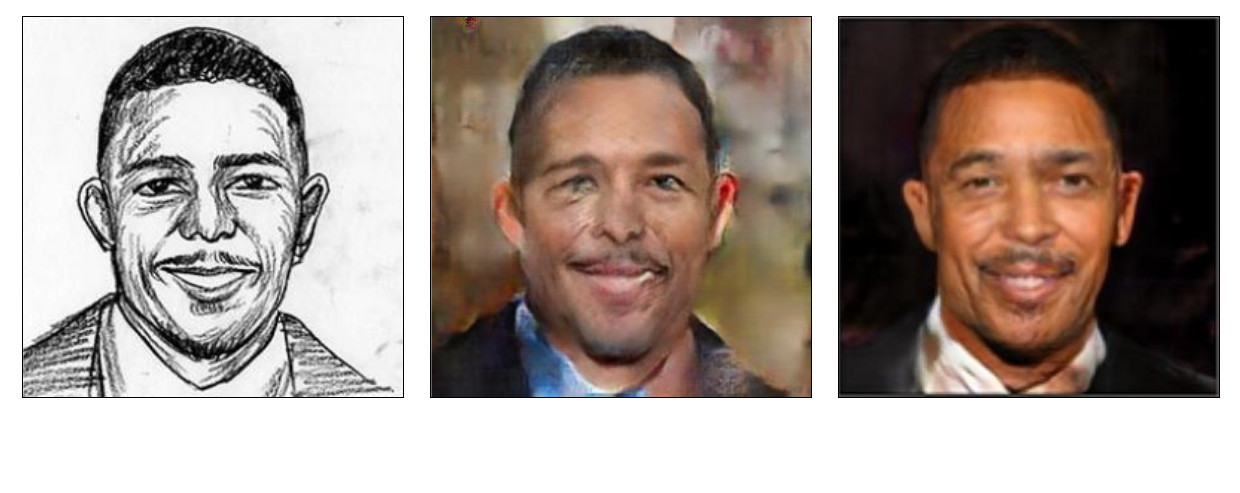}
   \put(8,2){(a) Input}
   \put(30,2){(b) w/o multi-patch}
   \put(73,2){(c) \ourmodel}
  \end{overpic}
  \caption{Ablation study for the S2I task.
  }\label{fig:abla_s2i}
\end{figure}

For the S2I task, an ablation study is conducted to validate the effectiveness of the facial component synthesis stage, as shown in \tabref{tab:ablation_s2i}. 
Similar to the I2S task, the multi-patch component achieves a significant performance gain (\ie, $3.3\%$) over the baseline model. 
\figref{fig:abla_s2i} provides examples of the results produced by our model and the model without the facial components synthesis stage. 
Our model with facial component synthesis captures more details and ensures a more realistic overall appearance (see \figref{fig:abla_s2i}-c).

\section{Discussion}\label{sec:Discussion}
Although FSS has achieved significant progress, there is still a large room for improvement. This section summarizes the possible future research directions related to FSS.

\textbf{(1) Datasets.} Due to the relative shortage of professional sketch artists, achieving large numbers of images remains an open problem, impeding the development of FSS. 
Furthermore, more diversified sketch (or drawing) styles are needed to build more attractive models and achieve better synthesis results. 
To address these issues, we believe novel data augmentation techniques \cite{yu2017sketchx,yu2016sketch,shorten2019survey} and transfer learning strategies \cite{wang2018transferring,yu2020deepi2i,shocher2020semantic} designed for FSS are promising directions of study.  

\begin{table}[t!]
    \centering
    \footnotesize
    \renewcommand{\arraystretch}{1.0}
    \renewcommand{\tabcolsep}{0.05pt}
    \caption{Ablation study of \ourmodel~on the I2S task.}
    \label{tab:ablation_i2s}
    \begin{tabular}{l|c|c|ll}
     \hline
     Setting & multi-patch & style vec.  & SCOOT$\uparrow$ & SSIM$\uparrow$ \\
     \hline
     Baseline & & & 0.381 & 0.487 \\  
     & \checkmark & & 0.386 (+1.31\%) & 0.500 (+2.67\%)\\  
     \textbf{\ourmodel} & \checkmark & \checkmark  &\textbf{0.405} (+6.30\%)  &\textbf{0.510} (+4.72\%)    \\\hline      
  \end{tabular}
\end{table}

\begin{table}[t!]
    \centering
    \footnotesize
    \renewcommand{\arraystretch}{1.0}
    \renewcommand{\tabcolsep}{15.0pt}
    \caption{Ablation study of our model on the S2I task.}
    \label{tab:ablation_s2i}
    \begin{tabular}{l|c|l}
    \hline
     Setting & multi-patch &  SSIM$\uparrow$ \\
    \hline
     Baseline &  & 0.487  \\  
   \textbf{\ourmodel} & \checkmark   &  \textbf{0.503} (+3.3\%)  \\
   \hline      
  \end{tabular}
\end{table}

\textbf{(2) Models.} 
Currently, most state-of-the-art models are trained with a large number of paired images, and sketches \cite{wang2018pix2pixhd,YiLLR20} to overcome data shortages.
However, more attention could be paid to techniques such as few-shot \cite{RaviL17}, semi-supervised \cite{chapelle2009semi}, weakly-supervised \cite{OquabBLS15}, self-supervised  \cite{wang2017transitive}, and \Rev{non-pairwise unsupervised \cite{sofgan}} learning to achieve style transfer with limited datasets. 
Besides, developing novel, human-in-the-loop \cite{PintoMT13} models is another promising direction, that would provide more interactive options to users for generating and editing personalized styles. 
Interactive models \Rev{that utilize the attributes in our \ourdataset}~could also serve as drawing tools provided to professional artists for facilitating the creation of sketches and other styles of drawing. 
Furthermore, FSS in the wild is still challenging because the image quality, including resolution, noise, and background, varies drastically. 
In addition to the techniques mentioned above, basic model units could also be focused on to develop new strategies. 
For example, most current models are built upon CNN \cite{lecun1989generalization} units. 
Therefore, more exploration of other frameworks, such as MLPs \cite{tolstikhin2021mlp} and Transformers \cite{vaswani2017attention, lee2021vitgan}, could also be conducted.  

\textbf{(3) Evaluation.} 
Evaluation metrics are essential for the development of new models and the benchmarking of existing models. 
Currently, several quantitative evaluation metrics \cite{zhang2011fsim, wang2004image} and human visual ranking methods \cite{song2014real} are used. 
However, as these aim to provide relatively objective and fair comparisons between all models, the different applications of FSS are not considered. 
This may lead to biased or unreliable evaluation of specific tasks. 
Therefore, more task-specific evaluation metrics and methods could be another important direction for future research. 

\textbf{(4) Applications.} 
Currently, the only direct applications of FSS (I2S and S2I) are entertainment, and law enforcement \cite{wang2008face,Wang2013IJCV}.
With the development of FSS techniques, many other promising applications could also be implicitly or explicitly facilitated by FSS research, such as art design and animation production. 
In addition to these industrial applications, we believe that FSS methods and ideas could also benefit other research fields. For example, sketches could be used to assist image resizing \cite{AvidanS07}, super-resolution \cite{dong2015image}, \etc. 
Further, the sketches usually contain the most conspicuous information of an image and can therefore be considered compressed versions of RGB images~\cite{hu2020towards}. 
This characteristic makes sketches useful for the image compression task. 
Besides, the S2I task can be considered a specific case of image super-resolution in a broad sense because both tasks aim to reconstruct detailed RGB images from the given inputs. 
The difference is that the input of S2I is high-frequency information, while that of the standard super-resolution task is the low-frequency information of the original image.

\section{Conclusion}\label{sec:Conclusion}
We have presented a complete review of the facial-sketch synthesis problem. To the best of our knowledge, this is the first systematic study on deep FSS in sketch-to-image and image-to-sketch tasks. 
To achieve this, we established a new challenging dataset, named \ourdataset. 
We also introduced a copy table for the proposed \ourdataset~to address the alignment issue between the sketches drawn by artists and the original images. 
The proposed simple baseline, \ourmodel,~achieves the new state-of-the-art performance with a two-stage architecture. 
Finally, as the most extensive survey (\ie, \Rev{$89$} literature methods) and benchmark (\ie, $19$ cutting-edge models), we have revealed that the development of this field is still in its infancy. 
Therefore, the main goal of this paper is to spark novel ideas rather than rank all existing models. It is not easy to benchmark all of the existing models due to the prosperity of the field.
We hope this \Rev{investigation} will attract the community's attention and yield exciting follow-up directions, such as generating vivid sketches with music, developing cartoons from sketches, synthesizing sketch videos, and fake faces~\cite{wood2021fake}.






\bibliographystyle{IEEEtran}
\bibliography{sketch_order}

\begin{thebibliography}{100}
\providecommand{\url}[1]{#1}
\csname url@samestyle\endcsname
\providecommand{\newblock}{\relax}
\providecommand{\bibinfo}[2]{#2}
\providecommand{\BIBentrySTDinterwordspacing}{\spaceskip=0pt\relax}
\providecommand{\BIBentryALTinterwordstretchfactor}{4}
\providecommand{\BIBentryALTinterwordspacing}{\spaceskip=\fontdimen2\font plus
\BIBentryALTinterwordstretchfactor\fontdimen3\font minus
  \fontdimen4\font\relax}
\providecommand{\BIBforeignlanguage}[2]{{%
\expandafter\ifx\csname l@#1\endcsname\relax
\typeout{** WARNING: IEEEtran.bst: No hyphenation pattern has been}%
\typeout{** loaded for the language `#1'. Using the pattern for}%
\typeout{** the default language instead.}%
\else
\language=\csname l@#1\endcsname
\fi
#2}}
\providecommand{\BIBdecl}{\relax}
\BIBdecl

\bibitem{wang2008face}
X.~Wang and X.~Tang, ``Face photo-sketch synthesis and recognition,''
  \emph{{IEEE Transactions on pattern analysis and machine intelligence}},
  vol.~31, no.~11, pp. 1955--1967, 2008.

\bibitem{yi2019apdrawinggan}
R.~Yi, Y.-J. Liu, Y.-K. Lai, and P.~L. Rosin, ``{APDrawingGAN}: Generating
  artistic portrait drawings from face photos with hierarchical {GAN}s,'' in
  \emph{{Conference on computer vision and pattern recognition}}.\hskip 1em
  plus 0.5em minus 0.4em\relax IEEE, 2019, pp. 10\,743--10\,752.

\bibitem{koshimizu1999kansei}
H.~Koshimizu, M.~Tominaga, T.~Fujiwara, and K.~Murakami, ``On kansei facial
  image processing for computerized facial caricaturing system picasso,'' in
  \emph{International Conference on Systems, Man, and Cybernetics}.\hskip 1em
  plus 0.5em minus 0.4em\relax IEEE, 1999, pp. 294--299.

\bibitem{kumar2009attribute}
N.~Kumar, A.~C. Berg, P.~N. Belhumeur, and S.~K. Nayar, ``Attribute and simile
  classifiers for face verification,'' in \emph{{International conference on
  computer vision}}.\hskip 1em plus 0.5em minus 0.4em\relax IEEE, 2009, pp.
  365--372.

\bibitem{du2015robust}
H.-S. Du, Q.-P. Hu, D.-F. Qiao, and I.~Pitas, ``Robust face recognition via
  low-rank sparse representation-based classification,'' \emph{International
  Journal of Automation and Computing}, vol.~12, no.~6, pp. 579--587, 2015.

\bibitem{lu2008novel}
Y.-Z. Lu, ``A novel face recognition algorithm for distinguishing faces with
  various angles,'' \emph{International Journal of Automation and Computing},
  vol.~5, no.~2, pp. 193--197, 2008.

\bibitem{jain2010fddb}
V.~Jain and E.~Learned-Miller, ``Fddb: A benchmark for face detection in
  unconstrained settings,'' UMass Amherst technical report, Tech. Rep., 2010.

\bibitem{zhang2014facial}
Z.~Zhang, P.~Luo, C.~C. Loy, and X.~Tang, ``Facial landmark detection by deep
  multi-task learning,'' in \emph{{European conference on computer
  vision}}.\hskip 1em plus 0.5em minus 0.4em\relax Springer, 2014, pp. 94--108.

\bibitem{bulat2017far}
A.~Bulat and G.~Tzimiropoulos, ``How far are we from solving the 2d \& 3d face
  alignment problem?(and a dataset of 230,000 3d facial landmarks),'' in
  \emph{{International conference on computer vision}}.\hskip 1em plus 0.5em
  minus 0.4em\relax IEEE, 2017, pp. 1021--1030.

\bibitem{sun2021multi}
J.~Sun, Q.~Li, W.~Wang, J.~Zhao, and Z.~Sun, ``Multi-caption text-to-face
  synthesis: Dataset and algorithm,'' in \emph{{International conference on
  Multimedia}}.\hskip 1em plus 0.5em minus 0.4em\relax ACM, 2021, pp.
  2290--2298.

\bibitem{bhatt2010matching}
H.~S. Bhatt, S.~Bharadwaj, R.~Singh, and M.~Vatsa, ``On matching sketches with
  digital face images,'' in \emph{International Conference on Biometrics:
  Theory, Applications and Systems}.\hskip 1em plus 0.5em minus 0.4em\relax
  IEEE, 2010, pp. 1--7.

\bibitem{zhang2011coupled}
W.~Zhang, X.~Wang, and X.~Tang, ``Coupled information-theoretic encoding for
  face photo-sketch recognition,'' in \emph{{Conference on computer vision and
  pattern recognition}}.\hskip 1em plus 0.5em minus 0.4em\relax IEEE, 2011, pp.
  513--520.

\bibitem{wang2011face}
N.~Wang, X.~Gao, D.~Tao, and X.~Li, ``Face sketch-photo synthesis under
  multi-dictionary sparse representation framework,'' in \emph{{International
  Conference on Image and Graphics}}.\hskip 1em plus 0.5em minus 0.4em\relax
  IEEE, 2011, pp. 82--87.

\bibitem{gao2012face}
X.~Gao, N.~Wang, D.~Tao, and X.~Li, ``Face sketch-photo synthesis and retrieval
  using sparse representation,'' \emph{{IEEE Transactions on circuits and
  systems for video technology}}, vol.~22, no.~8, pp. 1213--1226, 2012.

\bibitem{berger2013style}
I.~Berger, A.~Shamir, M.~Mahler, E.~Carter, and J.~Hodgins, ``Style and
  abstraction in portrait sketching,'' \emph{{ACM Transactions on graphics}},
  vol.~32, no.~4, pp. 1--12, 2013.

\bibitem{YiLLR20}
R.~Yi, Y.-J. Liu, Y.-K. Lai, and P.~L. Rosin, ``Unpaired portrait drawing
  generation via asymmetric cycle mapping,'' in \emph{{Conference on computer
  vision and pattern recognition}}.\hskip 1em plus 0.5em minus 0.4em\relax
  IEEE, 2020, pp. 8217--8225.

\bibitem{yi2020line}
R.~Yi, M.~Xia, Y.-J. Liu, Y.-K. Lai, and P.~L. Rosin, ``Line drawings for face
  portraits from photos using global and local structure based {GAN}s,''
  \emph{{IEEE Transactions on pattern analysis and machine intelligence}},
  vol.~43, no.~10, pp. 3462--3475, 2020.

\bibitem{zhang2018multidomain}
S.~Zhang, R.~Ji, J.~Hu, X.~Lu, and X.~Li, ``Face sketch synthesis by
  multidomain adversarial learning,'' \emph{{IEEE Transactions on Neural
  Networks and Learning Systems}}, vol.~30, no.~5, pp. 1419--1428, 2018.

\bibitem{zhu2020knowledge}
M.~Zhu, J.~Li, N.~Wang, and X.~Gao, ``Knowledge distillation for face
  photo-sketch synthesis,'' \emph{{IEEE Transactions on Neural Networks and
  Learning Systems}}, vol.~33, no.~2, pp. 893--906, 2022.

\bibitem{wang2004image}
Z.~Wang, A.~C. Bovik, H.~R. Sheikh, and E.~P. Simoncelli, ``Image quality
  assessment: from error visibility to structural similarity,'' \emph{{IEEE
  Transactions on image processing}}, vol.~13, no.~4, pp. 600--612, 2004.

\bibitem{zhu2017unpaired}
J.-Y. Zhu, T.~Park, P.~Isola, and A.~A. Efros, ``Unpaired image-to-image
  translation using cycle-consistent adversarial networks,'' in
  \emph{{International conference on computer vision}}.\hskip 1em plus 0.5em
  minus 0.4em\relax IEEE, 2017, pp. 2223--2232.

\bibitem{liu2017unsupervised}
M.-Y. Liu, T.~Breuel, and J.~Kautz, ``Unsupervised image-to-image translation
  networks,'' in \emph{{Advances in neural information processing
  systems}}.\hskip 1em plus 0.5em minus 0.4em\relax Curran Associates, Inc.,
  2017.

\bibitem{wang2018pix2pixhd}
T.-C. Wang, M.-Y. Liu, J.-Y. Zhu, A.~Tao, J.~Kautz, and B.~Catanzaro,
  ``High-resolution image synthesis and semantic manipulation with conditional
  {GAN}s,'' in \emph{{Conference on computer vision and pattern
  recognition}}.\hskip 1em plus 0.5em minus 0.4em\relax IEEE, 2018, pp.
  8798--8807.

\bibitem{park2019semantic}
T.~Park, M.-Y. Liu, T.-C. Wang, and J.-Y. Zhu, ``Semantic image synthesis with
  spatially-adaptive normalization,'' in \emph{{Conference on computer vision
  and pattern recognition}}.\hskip 1em plus 0.5em minus 0.4em\relax IEEE, 2019,
  pp. 2337--2346.

\bibitem{chang2020domain}
H.-Y. Chang, Z.~Wang, and Y.-Y. Chuang, ``Domain-specific mappings for
  generative adversarial style transfer,'' in \emph{{European conference on
  computer vision}}.\hskip 1em plus 0.5em minus 0.4em\relax Springer, 2020, pp.
  573--589.

\bibitem{Chen_2020_CVPR}
R.~Chen, W.~Huang, B.~Huang, F.~Sun, and B.~Fang, ``Reusing discriminators for
  encoding: Towards unsupervised image-to-image translation,'' in
  \emph{{Conference on computer vision and pattern recognition}}.\hskip 1em
  plus 0.5em minus 0.4em\relax IEEE, 2020, pp. 8168--8177.

\bibitem{DRIT_plus}
H.-Y. Lee, H.-Y. Tseng, Q.~Mao, J.-B. Huang, Y.-D. Lu, M.~Singh, and M.-H.
  Yang, ``Drit++: Diverse image-to-image translation via disentangled
  representations,'' \emph{{International journal of computer vision}}, vol.
  128, no.~10, pp. 2402--2417, 2020.

\bibitem{liu2015deep}
Z.~Liu, P.~Luo, X.~Wang, and X.~Tang, ``Deep learning face attributes in the
  wild,'' in \emph{{International conference on computer vision}}.\hskip 1em
  plus 0.5em minus 0.4em\relax IEEE, 2015, pp. 3730--3738.

\bibitem{fan2019scoot}
D.-P. Fan, S.~Zhang, Y.-H. Wu, Y.~Liu, M.-M. Cheng, B.~Ren, P.~L. Rosin, and
  R.~Ji, ``Scoot: A perceptual metric for facial sketches,'' in
  \emph{{International conference on computer vision}}.\hskip 1em plus 0.5em
  minus 0.4em\relax IEEE, 2019, pp. 5612--5622.

\bibitem{kim2019u}
J.~Kim, M.~Kim, H.~Kang, and K.~Lee, ``U-gat-it: unsupervised generative
  attentional networks with adaptive layer-instance normalization for
  image-to-image translation,'' in \emph{{International Conference on Learning
  Representations}}.\hskip 1em plus 0.5em minus 0.4em\relax OpenReview.net,
  2020.

\bibitem{isola2017image}
P.~Isola, J.-Y. Zhu, T.~Zhou, and A.~A. Efros, ``Image-to-image translation
  with conditional adversarial networks,'' in \emph{{Conference on computer
  vision and pattern recognition}}.\hskip 1em plus 0.5em minus 0.4em\relax
  IEEE, 2017, pp. 1125--1134.

\bibitem{peng2018face}
C.~Peng, X.~Gao, N.~Wang, and J.~Li, ``Face recognition from multiple stylistic
  sketches: Scenarios, datasets, and evaluation,'' \emph{{Pattern
  Recognition}}, vol.~84, pp. 262--272, 2018.

\bibitem{martinez1998ar}
A.~M. Martinez, ``The ar face database,'' \emph{CVC Technical Report24}, 1998.

\bibitem{messer1999xm2vtsdb}
K.~Messer, J.~Matas, J.~Kittler, J.~Luettin, G.~Maitre \emph{et~al.},
  ``{XM2VTSDB}: The extended m2vts database,'' in \emph{International
  conference on audio and video-based biometric person authentication}, vol.
  964.\hskip 1em plus 0.5em minus 0.4em\relax Springer, 1999, pp. 965--966.

\bibitem{phillips2000feret}
P.~J. Phillips, H.~Moon, S.~A. Rizvi, and P.~J. Rauss, ``The {FERET} evaluation
  methodology for face-recognition algorithms,'' \emph{{IEEE Transactions on
  pattern analysis and machine intelligence}}, vol.~22, no.~10, pp. 1090--1104,
  2000.

\bibitem{serrano2007influence}
{\'A}.~Serrano, I.~M. de~Diego, C.~Conde, E.~Cabello, L.~Shen, and L.~Bai,
  ``Influence of wavelet frequency and orientation in an {SVM}-based parallel
  gabor {PCA} face verification system,'' in \emph{International Conference on
  Intelligent Data Engineering and Automated Learning}.\hskip 1em plus 0.5em
  minus 0.4em\relax Springer, 2007, pp. 219--228.

\bibitem{bhatt2012memetically}
H.~S. Bhatt, S.~Bharadwaj, R.~Singh, and M.~Vatsa, ``Memetically optimized
  {MCWLD} for matching sketches with digital face images,'' \emph{Transactions
  on Information Forensics and Security}, vol.~7, no.~5, pp. 1522--1535, 2012.

\bibitem{minear2004lifespan}
M.~Minear and D.~C. Park, ``A lifespan database of adult facial stimuli,''
  \emph{Behavior research methods, instruments, \& computers}, vol.~36, no.~4,
  pp. 630--633, 2004.

\bibitem{nishino1999linguistic}
J.~Nishino, T.~Kamyama, H.~Shira, T.~Odaka, and H.~Ogura, ``Linguistic
  knowledge acquisition system on facial caricature drawing system,'' in
  \emph{{International Fuzzy Systems}}.\hskip 1em plus 0.5em minus 0.4em\relax
  IEEE, 1999, pp. 1591--1596.

\bibitem{iwashita1999expressive}
S.~Iwashita, Y.~Takeda, and T.~Onisawa, ``Expressive facial caricature
  drawing,'' in \emph{{International Fuzzy Systems}}.\hskip 1em plus 0.5em
  minus 0.4em\relax IEEE, 1999, pp. 1597--1602.

\bibitem{li1997extraction}
Y.~Li and H.~Kobatake, ``Extraction of facial sketch image based on
  morphological processing,'' in \emph{{International Conference on Image
  Processing}}.\hskip 1em plus 0.5em minus 0.4em\relax IEEE, 1997, pp.
  316--319.

\bibitem{tominaga1997facial}
M.~Tominaga, S.~Fukuoka, K.~Murakami, and H.~Koshimizu, ``Facial caricaturing
  with motion caricaturing in {PICASSO} system,'' in \emph{International
  Conference on Advanced Intelligent Mechatronics}.\hskip 1em plus 0.5em minus
  0.4em\relax IEEE, 1997, p.~30.

\bibitem{brennan1982caricature}
S.~E. Brennan, ``Caricature generator,'' Ph.D. dissertation, Massachusetts
  Institute of Technology, 1982.

\bibitem{Wang2013IJCV}
N.~Wang, D.~Tao, X.~Gao, X.~Li, and J.~Li, ``A comprehensive survey to face
  hallucination,'' \emph{{International journal of computer vision}}, vol. 106,
  no.~1, pp. 9--30, 2014.

\bibitem{chen2001example}
H.~Chen, Y.-Q. Xu, H.-Y. Shum, S.-C. Zhu, and N.-N. Zheng, ``Example-based
  facial sketch generation with non-parametric sampling,'' in
  \emph{{International conference on computer vision}}.\hskip 1em plus 0.5em
  minus 0.4em\relax IEEE, 2001, pp. 433--438.

\bibitem{nefian1999face}
A.~V. Nefian and M.~H. Hayes~III, ``Face recognition using an embedded hmm,''
  in \emph{Conference on Audio and Video-based Biometric Person
  Authentication}.\hskip 1em plus 0.5em minus 0.4em\relax IEEE, 1999.

\bibitem{gao2008face}
X.~Gao, J.~Zhong, J.~Li, and C.~Tian, ``Face sketch synthesis algorithm based
  on e-hmm and selective ensemble,'' \emph{{IEEE Transactions on circuits and
  systems for video technology}}, vol.~18, no.~4, pp. 487--496, 2008.

\bibitem{xu2008hierarchical}
Z.~Xu, H.~Chen, S.-C. Zhu, and J.~Luo, ``A hierarchical compositional model for
  face representation and sketching,'' \emph{{IEEE Transactions on pattern
  analysis and machine intelligence}}, vol.~30, no.~6, pp. 955--969, 2008.

\bibitem{zhang2010lighting}
W.~Zhang, X.~Wang, and X.~Tang, ``Lighting and pose robust face sketch
  synthesis,'' in \emph{{European conference on computer vision}}.\hskip 1em
  plus 0.5em minus 0.4em\relax Springer, 2010, pp. 420--433.

\bibitem{zhou2012markov}
H.~Zhou, Z.~Kuang, and K.-Y.~K. Wong, ``Markov weight fields for face sketch
  synthesis,'' in \emph{{Conference on computer vision and pattern
  recognition}}.\hskip 1em plus 0.5em minus 0.4em\relax IEEE, 2012, pp.
  1091--1097.

\bibitem{wang2013learnable}
T.~Wang, J.~P. Collomosse, A.~Hunter, and D.~Greig, ``Learnable stroke models
  for example-based portrait painting,'' in \emph{{British Machine Vision
  Conference}}.\hskip 1em plus 0.5em minus 0.4em\relax {BMVA} Press, 2013.

\bibitem{wang2013transductive}
N.~Wang, D.~Tao, X.~Gao, X.~Li, and J.~Li, ``Transductive face sketch-photo
  synthesis,'' \emph{{IEEE Transactions on Neural Networks and Learning
  Systems}}, vol.~24, no.~9, pp. 1364--1376, 2013.

\bibitem{peng2015superpixel}
C.~Peng, X.~Gao, N.~Wang, and J.~Li, ``Superpixel-based face sketch-photo
  synthesis,'' \emph{{IEEE Transactions on circuits and systems for video
  technology}}, vol.~27, no.~2, pp. 288--299, 2015.

\bibitem{peng2015multiple}
C.~Peng, X.~Gao, N.~Wang, D.~Tao, X.~Li, and J.~Li, ``Multiple
  representations-based face sketch-photo synthesis,'' \emph{{IEEE Transactions
  on Neural Networks and Learning Systems}}, vol.~27, no.~11, pp. 2201--2215,
  2015.

\bibitem{abdi2010principal}
H.~Abdi and L.~J. Williams, ``Principal component analysis,'' \emph{Wiley
  interdisciplinary reviews: computational statistics}, vol.~2, no.~4, pp.
  433--459, 2010.

\bibitem{tang2002face}
X.~Tang and X.~Wang, ``Face photo recognition using sketch,'' in
  \emph{{International Conference on Image Processing}}.\hskip 1em plus 0.5em
  minus 0.4em\relax IEEE, 2002, pp. I--I.

\bibitem{tang2003face}
{X. Tang and X. Wang}, ``Face sketch synthesis and recognition,'' in
  \emph{{International conference on computer vision}}.\hskip 1em plus 0.5em
  minus 0.4em\relax IEEE, 2003, pp. 687--694.

\bibitem{tang2004face}
------, ``Face sketch recognition,'' \emph{{IEEE Transactions on circuits and
  systems for video technology}}, vol.~14, no.~1, pp. 50--57, 2004.

\bibitem{liu2005nonlinear}
Q.~Liu, X.~Tang, H.~Jin, H.~Lu, and S.~Ma, ``A nonlinear approach for face
  sketch synthesis and recognition,'' in \emph{{Conference on computer vision
  and pattern recognition}}.\hskip 1em plus 0.5em minus 0.4em\relax IEEE, 2005,
  pp. 1005--1010.

\bibitem{huang2013coupled}
D.-A. Huang and Y.-C.~F. Wang, ``Coupled dictionary and feature space learning
  with applications to cross-domain image synthesis and recognition,'' in
  \emph{{International conference on computer vision}}.\hskip 1em plus 0.5em
  minus 0.4em\relax IEEE, 2013, pp. 2496--2503.

\bibitem{ji2011local}
N.~Ji, X.~Chai, S.~Shan, and X.~Chen, ``Local regression model for automatic
  face sketch generation,'' in \emph{{International Conference on Image and
  Graphics}}.\hskip 1em plus 0.5em minus 0.4em\relax IEEE, 2011, pp. 412--417.

\bibitem{chang2011face}
L.~Chang, M.~Zhou, X.~Deng, Z.~Wu, and Y.~Han, ``Face sketch synthesis via
  multivariate output regression,'' in \emph{International conference on
  human-computer interaction}.\hskip 1em plus 0.5em minus 0.4em\relax Springer,
  2011, pp. 555--561.

\bibitem{zhang2011face}
J.~Zhang, N.~Wang, X.~Gao, D.~Tao, and X.~Li, ``Face sketch-photo synthesis
  based on support vector regression,'' in \emph{{International Conference on
  Image Processing}}.\hskip 1em plus 0.5em minus 0.4em\relax IEEE, 2011, pp.
  1125--1128.

\bibitem{roweis2000nonlinear}
S.~T. Roweis and L.~K. Saul, ``Nonlinear dimensionality reduction by locally
  linear embedding,'' \emph{Science}, vol. 290, no. 5500, pp. 2323--2326, 2000.

\bibitem{wang2012semi}
S.~Wang, L.~Zhang, Y.~Liang, and Q.~Pan, ``Semi-coupled dictionary learning
  with applications to image super-resolution and photo-sketch synthesis,'' in
  \emph{{Conference on computer vision and pattern recognition}}.\hskip 1em
  plus 0.5em minus 0.4em\relax IEEE, 2012, pp. 2216--2223.

\bibitem{song2014real}
Y.~Song, L.~Bao, Q.~Yang, and M.-H. Yang, ``Real-time exemplar-based face
  sketch synthesis,'' in \emph{{European conference on computer vision}}.\hskip
  1em plus 0.5em minus 0.4em\relax Springer, 2014, pp. 800--813.

\bibitem{wang2018random}
N.~Wang, X.~Gao, and J.~Li, ``Random sampling for fast face sketch synthesis,''
  \emph{{Pattern Recognition}}, vol.~76, pp. 215--227, 2018.

\bibitem{zhang2015robust}
S.~Zhang, X.~Gao, N.~Wang, and J.~Li, ``Robust face sketch style synthesis,''
  \emph{{IEEE Transactions on image processing}}, vol.~25, no.~1, pp. 220--232,
  2015.

\bibitem{li2017free}
Y.~Li, Y.-Z. Song, T.~M. Hospedales, and S.~Gong, ``Free-hand sketch synthesis
  with deformable stroke models,'' \emph{{International journal of computer
  vision}}, vol. 122, no.~1, pp. 169--190, 2017.

\bibitem{li2017adaptive}
J.~Li, X.~Yu, C.~Peng, and N.~Wang, ``Adaptive representation-based face
  sketch-photo synthesis,'' \emph{Neurocomputing}, vol. 269, pp. 152--159,
  2017.

\bibitem{men2018common}
Y.~Men, Z.~Lian, Y.~Tang, and J.~Xiao, ``A common framework for interactive
  texture transfer,'' in \emph{{Conference on computer vision and pattern
  recognition}}.\hskip 1em plus 0.5em minus 0.4em\relax IEEE, 2018, pp.
  6353--6362.

\bibitem{saxena2021comparison}
S.~Saxena and M.~N. Teli, ``Comparison and analysis of image-to-image
  generative adversarial networks: A survey,'' \emph{arXiv preprint
  arXiv:2112.12625}, 2021.

\bibitem{goodfellow2014generative}
I.~J. Goodfellow, J.~Pouget-Abadie, M.~Mirza, B.~Xu, D.~Warde-Farley, S.~Ozair,
  A.~Courville, and Y.~Bengio, ``Generative adversarial networks,'' in
  \emph{{Advances in neural information processing systems}}.\hskip 1em plus
  0.5em minus 0.4em\relax Curran Associates, Inc., 2014.

\bibitem{mirza2014conditional}
M.~Mirza and S.~Osindero, ``Conditional generative adversarial nets,'' in
  \emph{{Advances in neural information processing systems workshops}}.\hskip
  1em plus 0.5em minus 0.4em\relax Curran Associates, Inc., 2014.

\bibitem{ronneberger2015u}
O.~Ronneberger, P.~Fischer, and T.~Brox, ``U-net: Convolutional networks for
  biomedical image segmentation,'' in \emph{{Medical Image Computing and
  Computer Assisted Intervention}}.\hskip 1em plus 0.5em minus 0.4em\relax
  Springer, 2015, pp. 234--241.

\bibitem{zhu2017toward}
J.-Y. Zhu, R.~Zhang, D.~Pathak, T.~Darrell, A.~A. Efros, O.~Wang, and
  E.~Shechtman, ``Toward multimodal image-to-image translation,'' in
  \emph{{Advances in neural information processing systems}}.\hskip 1em plus
  0.5em minus 0.4em\relax Curran Associates, Inc., 2017.

\bibitem{kim2017learning}
T.~Kim, M.~Cha, H.~Kim, J.~K. Lee, and J.~Kim, ``Learning to discover
  cross-domain relations with generative adversarial networks,'' in
  \emph{{International conference on machine learning}}.\hskip 1em plus 0.5em
  minus 0.4em\relax PMLR, 2017, pp. 1857--1865.

\bibitem{yi2017dualgan}
Z.~Yi, H.~Zhang, P.~Tan, and M.~Gong, ``{DualGAN}: Unsupervised dual learning
  for image-to-image translation,'' in \emph{{International conference on
  computer vision}}.\hskip 1em plus 0.5em minus 0.4em\relax IEEE, 2017, pp.
  2849--2857.

\bibitem{zhao2020unpaired}
Y.~Zhao, R.~Wu, and H.~Dong, ``Unpaired image-to-image translation using
  adversarial consistency loss,'' in \emph{{European conference on computer
  vision}}.\hskip 1em plus 0.5em minus 0.4em\relax Springer, 2020, pp.
  800--815.

\bibitem{jiang2020tsit}
L.~Jiang, C.~Zhang, M.~Huang, C.~Liu, J.~Shi, and C.~C. Loy, ``{TSIT}: A simple
  and versatile framework for image-to-image translation,'' in \emph{{European
  conference on computer vision}}.\hskip 1em plus 0.5em minus 0.4em\relax
  Springer, 2020, pp. 206--222.

\bibitem{zhang2020cross}
P.~Zhang, B.~Zhang, D.~Chen, L.~Yuan, and F.~Wen, ``Cross-domain correspondence
  learning for exemplar-based image translation,'' in \emph{{Conference on
  computer vision and pattern recognition}}.\hskip 1em plus 0.5em minus
  0.4em\relax IEEE, 2020, pp. 5143--5153.

\bibitem{zhou2021cocosnet}
X.~Zhou, B.~Zhang, T.~Zhang, P.~Zhang, J.~Bao, D.~Chen, Z.~Zhang, and F.~Wen,
  ``{CoCosNet} v2: Full-resolution correspondence learning for image
  translation,'' in \emph{{Conference on computer vision and pattern
  recognition}}.\hskip 1em plus 0.5em minus 0.4em\relax IEEE, 2021, pp.
  11\,465--11\,475.

\bibitem{sofgan}
A.~Chen, R.~Liu, L.~Xie, Z.~Chen, H.~Su, and J.~Yu, ``Sofgan: A portrait image
  generator with dynamic styling,'' \emph{{ACM Transactions on graphics}},
  vol.~41, no.~1, pp. 1--26, 2022.

\bibitem{jing2019neural}
Y.~Jing, Y.~Yang, Z.~Feng, J.~Ye, Y.~Yu, and M.~Song, ``Neural style transfer:
  A review,'' \emph{{IEEE Transactions on visualization and computer
  graphics}}, vol.~26, no.~11, pp. 3365--3385, 2019.

\bibitem{chen2018cartoongan}
Y.~Chen, Y.-K. Lai, and Y.-J. Liu, ``Cartoongan: Generative adversarial
  networks for photo cartoonization,'' in \emph{{Conference on computer vision
  and pattern recognition}}.\hskip 1em plus 0.5em minus 0.4em\relax IEEE, 2018,
  pp. 9465--9474.

\bibitem{richardson2020encoding}
E.~Richardson, Y.~Alaluf, O.~Patashnik, Y.~Nitzan, Y.~Azar, S.~Shapiro, and
  D.~Cohen-Or, ``Encoding in style: a stylegan encoder for image-to-image
  translation,'' in \emph{{Conference on computer vision and pattern
  recognition}}.\hskip 1em plus 0.5em minus 0.4em\relax IEEE, 2021, pp.
  2287--2296.

\bibitem{gatys2015neural}
L.~A. Gatys, A.~S. Ecker, and M.~Bethge, ``A neural algorithm of artistic
  style,'' \emph{arXiv preprint arXiv:1508.06576}, 2015.

\bibitem{gatys2016image}
{L. A. Gatys, A. S. Ecker, and M. Bethge}, ``Image style transfer using
  convolutional neural networks,'' in \emph{{Conference on computer vision and
  pattern recognition}}.\hskip 1em plus 0.5em minus 0.4em\relax IEEE, 2016, pp.
  2414--2423.

\bibitem{simonyan2015very}
K.~Simonyan and A.~Zisserman, ``Very deep convolutional networks for
  large-scale image recognition,'' in \emph{{International Conference on
  Learning Representations}}.\hskip 1em plus 0.5em minus 0.4em\relax
  OpenReview.net, 2015.

\bibitem{karras2019style}
T.~Karras, S.~Laine, and T.~Aila, ``A style-based generator architecture for
  generative adversarial networks,'' in \emph{{Conference on computer vision
  and pattern recognition}}.\hskip 1em plus 0.5em minus 0.4em\relax IEEE, 2019,
  pp. 4401--4410.

\bibitem{abdal2019image2stylegan}
R.~Abdal, Y.~Qin, and P.~Wonka, ``{Image2styleGAN}: How to embed images into
  the stylegan latent space?'' in \emph{{International conference on computer
  vision}}.\hskip 1em plus 0.5em minus 0.4em\relax IEEE, 2019, pp. 4432--4441.

\bibitem{kotovenko2021rethinking}
D.~Kotovenko, M.~Wright, A.~Heimbrecht, and B.~Ommer, ``Rethinking style
  transfer: From pixels to parameterized brushstrokes,'' in \emph{{Conference
  on computer vision and pattern recognition}}.\hskip 1em plus 0.5em minus
  0.4em\relax IEEE, 2021, pp. 12\,196--12\,205.

\bibitem{johnson2016perceptual}
J.~Johnson, A.~Alahi, and L.~Fei-Fei, ``Perceptual losses for real-time style
  transfer and super-resolution,'' in \emph{{European conference on computer
  vision}}.\hskip 1em plus 0.5em minus 0.4em\relax Springer, 2016, pp.
  694--711.

\bibitem{ulyanov2016texture}
D.~Ulyanov, V.~Lebedev, A.~Vedaldi, and V.~S. Lempitsky, ``Texture networks:
  Feed-forward synthesis of textures and stylized images,'' in
  \emph{{International conference on machine learning}}.\hskip 1em plus 0.5em
  minus 0.4em\relax PMLR, 2016, p. 1349–1357.

\bibitem{chen2016fast}
T.~Q. Chen and M.~Schmidt, ``Fast patch-based style transfer of arbitrary
  style,'' in \emph{{Advances in neural information processing systems
  workshops}}.\hskip 1em plus 0.5em minus 0.4em\relax Curran Associates, Inc.,
  2016.

\bibitem{dumoulin2016learned}
V.~Dumoulin, J.~Shlens, and M.~Kudlur, ``A learned representation for artistic
  style,'' in \emph{{International Conference on Learning
  Representations}}.\hskip 1em plus 0.5em minus 0.4em\relax OpenReview.net,
  2017.

\bibitem{ulyanov2017improved}
D.~Ulyanov, A.~Vedaldi, and V.~Lempitsky, ``Improved texture networks:
  Maximizing quality and diversity in feed-forward stylization and texture
  synthesis,'' in \emph{{Conference on computer vision and pattern
  recognition}}.\hskip 1em plus 0.5em minus 0.4em\relax IEEE, 2017, pp.
  6924--6932.

\bibitem{huang2017arbitrary}
X.~Huang and S.~Belongie, ``Arbitrary style transfer in real-time with adaptive
  instance normalization,'' in \emph{{International conference on computer
  vision}}.\hskip 1em plus 0.5em minus 0.4em\relax IEEE, 2017, pp. 1501--1510.

\bibitem{WCT-NIPS-2017}
Y.~Li, C.~Fang, J.~Yang, Z.~Wang, X.~Lu, and M.-H. Yang, ``Universal style
  transfer via feature transforms,'' in \emph{{Advances in neural information
  processing systems}}.\hskip 1em plus 0.5em minus 0.4em\relax Curran
  Associates, Inc., 2017.

\bibitem{huang2018multimodal}
X.~Huang, M.-Y. Liu, S.~Belongie, and J.~Kautz, ``Multimodal unsupervised
  image-to-image translation,'' in \emph{{European conference on computer
  vision}}.\hskip 1em plus 0.5em minus 0.4em\relax Springer, 2018, pp.
  172--189.

\bibitem{eitz2012humans}
M.~Eitz, J.~Hays, and M.~Alexa, ``How do humans sketch objects?'' \emph{{ACM
  Transactions on graphics}}, vol.~31, no.~4, pp. 1--10, 2012.

\bibitem{lin2014microsoft}
T.-Y. Lin, M.~Maire, S.~Belongie, J.~Hays, P.~Perona, D.~Ramanan,
  P.~Doll{\'a}r, and C.~L. Zitnick, ``Microsoft {COCO}: Common objects in
  context,'' in \emph{{European conference on computer vision}}.\hskip 1em plus
  0.5em minus 0.4em\relax Springer, 2014, pp. 740--755.

\bibitem{ILSVRC15}
O.~Russakovsky, J.~Deng, H.~Su, J.~Krause, S.~Satheesh, S.~Ma, Z.~Huang,
  A.~Karpathy, A.~Khosla, M.~Bernstein, A.~C. Berg, and L.~Fei-Fei,
  ``{ImageNet} large scale visual recognition challenge,'' \emph{{International
  journal of computer vision}}, vol. 115, no.~3, pp. 211--252, 2015.

\bibitem{cimpoi2014describing}
M.~Cimpoi, S.~Maji, I.~Kokkinos, S.~Mohamed, and A.~Vedaldi, ``Describing
  textures in the wild,'' in \emph{{Conference on computer vision and pattern
  recognition}}.\hskip 1em plus 0.5em minus 0.4em\relax IEEE, 2014, pp.
  3606--3613.

\bibitem{wikiart}
S.~Y. Duck, ``Painter by \\numbers, wikiart.org,'' https://
  www.kaggle.com/c/painter-by-\\numbers, 2016.

\bibitem{cordts2016cityscapes}
M.~Cordts, M.~Omran, S.~Ramos, T.~Rehfeld, M.~Enzweiler, R.~Benenson,
  U.~Franke, S.~Roth, and B.~Schiele, ``The cityscapes dataset for semantic
  urban scene understanding,'' in \emph{{Conference on computer vision and
  pattern recognition}}.\hskip 1em plus 0.5em minus 0.4em\relax IEEE, 2016, pp.
  3213--3223.

\bibitem{tylevcek2013spatial}
R.~Tyle{\v{c}}ek and R.~{\v{S}}{\'a}ra, ``Spatial pattern templates for
  recognition of objects with regular structure,'' in \emph{German conference
  on pattern recognition}.\hskip 1em plus 0.5em minus 0.4em\relax Springer,
  2013, pp. 364--374.

\bibitem{zhu2016generative}
J.-Y. Zhu, P.~Kr{\"a}henb{\"u}hl, E.~Shechtman, and A.~A. Efros, ``Generative
  visual manipulation on the natural image manifold,'' in \emph{{European
  conference on computer vision}}.\hskip 1em plus 0.5em minus 0.4em\relax
  Springer, 2016, pp. 597--613.

\bibitem{yu2014fine}
A.~Yu and K.~Grauman, ``Fine-grained visual comparisons with local learning,''
  in \emph{{Conference on computer vision and pattern recognition}}.\hskip 1em
  plus 0.5em minus 0.4em\relax IEEE, 2014, pp. 192--199.

\bibitem{laffont2014transient}
P.-Y. Laffont, Z.~Ren, X.~Tao, C.~Qian, and J.~Hays, ``Transient attributes for
  high-level understanding and editing of outdoor scenes,'' \emph{{ACM
  Transactions on graphics}}, vol.~33, no.~4, pp. 1--11, 2014.

\bibitem{lecun1998gradient}
Y.~LeCun, L.~Bottou, Y.~Bengio, and P.~Haffner, ``Gradient-based learning
  applied to document recognition,'' \emph{Proceedings of the IEEE}, vol.~86,
  no.~11, pp. 2278--2324, 1998.

\bibitem{wah2011caltech}
C.~Wah, S.~Branson, P.~Welinder, P.~Perona, and S.~Belongie, ``The caltech-ucsd
  birds-200-2011 dataset,'' 2011.

\bibitem{karras2017progressive}
T.~Karras, T.~Aila, S.~Laine, and J.~Lehtinen, ``Progressive growing of {GAN}s
  for improved quality, stability, and variation,'' in \emph{{International
  Conference on Learning Representations}}.\hskip 1em plus 0.5em minus
  0.4em\relax OpenReview.net, 2018.

\bibitem{silberman2012indoor}
N.~Silberman, D.~Hoiem, P.~Kohli, and R.~Fergus, ``Indoor segmentation and
  support inference from rgbd images,'' in \emph{{European conference on
  computer vision}}.\hskip 1em plus 0.5em minus 0.4em\relax Springer, 2012, pp.
  746--760.

\bibitem{zhou2017scene}
B.~Zhou, H.~Zhao, X.~Puig, S.~Fidler, A.~Barriuso, and A.~Torralba, ``Scene
  parsing through ade20k dataset,'' in \emph{{Conference on computer vision and
  pattern recognition}}.\hskip 1em plus 0.5em minus 0.4em\relax IEEE, 2017, pp.
  633--641.

\bibitem{yu2017sketchx}
Q.~Yu, Y.-Z. Song, T.~Xiang, and T.~M. Hospedales, ``Sketchx!-shoe/chair
  fine-grained {SBIR} dataset,'' 2017.

\bibitem{ha2017neural}
D.~Ha and D.~Eck, ``A neural representation of sketch drawings,'' in
  \emph{{International Conference on Learning Representations}}.\hskip 1em plus
  0.5em minus 0.4em\relax OpenReview.net, 2018.

\bibitem{jin2017towards}
Y.~Jin, J.~Zhang, M.~Li, Y.~Tian, H.~Zhu, and Z.~Fang, ``Towards the automatic
  anime characters creation with generative adversarial networks,'' in
  \emph{{Advances in neural information processing systems workshops}}.\hskip
  1em plus 0.5em minus 0.4em\relax Curran Associates, Inc., 2017.

\bibitem{xu2017end}
H.~Xu, Y.~Gao, F.~Yu, and T.~Darrell, ``End-to-end learning of driving models
  from large-scale video datasets,'' in \emph{{Conference on computer vision
  and pattern recognition}}.\hskip 1em plus 0.5em minus 0.4em\relax IEEE, 2017,
  pp. 2174--2182.

\bibitem{ros2016synthia}
G.~Ros, L.~Sellart, J.~Materzynska, D.~Vazquez, and A.~M. Lopez, ``The synthia
  dataset: A large collection of synthetic images for semantic segmentation of
  urban scenes,'' in \emph{{Conference on computer vision and pattern
  recognition}}.\hskip 1em plus 0.5em minus 0.4em\relax IEEE, 2016, pp.
  3234--3243.

\bibitem{liu2016deepfashion}
Z.~Liu, P.~Luo, S.~Qiu, X.~Wang, and X.~Tang, ``Deepfashion: Powering robust
  clothes recognition and retrieval with rich annotations,'' in
  \emph{{Conference on computer vision and pattern recognition}}.\hskip 1em
  plus 0.5em minus 0.4em\relax IEEE, 2016, pp. 1096--1104.

\bibitem{agustsson2017ntire}
E.~Agustsson and R.~Timofte, ``Ntire 2017 challenge on single image
  super-resolution: Dataset and study,'' in \emph{{Conference on computer
  vision and pattern recognition workshops}}.\hskip 1em plus 0.5em minus
  0.4em\relax IEEE, 2017, pp. 126--135.

\bibitem{yao2007introduction}
B.~Yao, X.~Yang, and S.-C. Zhu, ``Introduction to a large-scale general purpose
  ground truth database: methodology, annotation tool and benchmarks,'' in
  \emph{CVPRW}.\hskip 1em plus 0.5em minus 0.4em\relax IEEE, 2007, pp.
  169--183.

\bibitem{krause20133d}
J.~Krause, M.~Stark, J.~Deng, and L.~Fei-Fei, ``3d object representations for
  fine-grained categorization,'' in \emph{{International conference on computer
  vision workshops}}.\hskip 1em plus 0.5em minus 0.4em\relax IEEE, 2013, pp.
  554--561.

\bibitem{yu2015lsun}
F.~Yu, A.~Seff, Y.~Zhang, S.~Song, T.~Funkhouser, and J.~Xiao, ``Lsun:
  Construction of a large-scale image dataset using deep learning with humans
  in the loop,'' \emph{arXiv preprint arXiv:1506.03365}, 2015.

\bibitem{zhang2015end}
L.~Zhang, L.~Lin, X.~Wu, S.~Ding, and L.~Zhang, ``End-to-end photo-sketch
  generation via fully convolutional representation learning,'' in
  \emph{International Conference on Multimedia Retrieval}.\hskip 1em plus 0.5em
  minus 0.4em\relax ACM, 2015, pp. 627--634.

\bibitem{zhu2017deep}
M.~Zhu, N.~Wang, X.~Gao, and J.~Li, ``Deep graphical feature learning for face
  sketch synthesis,'' in \emph{{International Joint Conference on Artificial
  Intelligence}}.\hskip 1em plus 0.5em minus 0.4em\relax IJCAI, 2017, pp.
  3574--3580.

\bibitem{sangkloy2017scribbler}
P.~Sangkloy, J.~Lu, C.~Fang, F.~Yu, and J.~Hays, ``Scribbler: Controlling deep
  image synthesis with sketch and color,'' in \emph{{Conference on computer
  vision and pattern recognition}}.\hskip 1em plus 0.5em minus 0.4em\relax
  IEEE, 2017, pp. 5400--5409.

\bibitem{zhang2018face}
M.~Zhang, N.~Wang, Y.~Li, R.~Wang, and X.~Gao, ``Face sketch synthesis from
  coarse to fine,'' in \emph{{AAAI Conference on Artificial
  Intelligence}}.\hskip 1em plus 0.5em minus 0.4em\relax AAAI Press, 2018, pp.
  7558--7565.

\bibitem{xian2018texturegan}
W.~Xian, P.~Sangkloy, V.~Agrawal, A.~Raj, J.~Lu, C.~Fang, F.~Yu, and J.~Hays,
  ``{TextureGAN}: Controlling deep image synthesis with texture patches,'' in
  \emph{{Conference on computer vision and pattern recognition}}.\hskip 1em
  plus 0.5em minus 0.4em\relax IEEE, 2018, pp. 8456--8465.

\bibitem{song2018learning}
J.~Song, K.~Pang, Y.-Z. Song, T.~Xiang, and T.~M. Hospedales, ``Learning to
  sketch with shortcut cycle consistency,'' in \emph{{Conference on computer
  vision and pattern recognition}}.\hskip 1em plus 0.5em minus 0.4em\relax
  IEEE, 2018, pp. 801--810.

\bibitem{lu2018image}
Y.~Lu, S.~Wu, Y.-W. Tai, and C.-K. Tang, ``Image generation from sketch
  constraint using contextual {GAN},'' in \emph{{European conference on
  computer vision}}.\hskip 1em plus 0.5em minus 0.4em\relax Springer, 2018, pp.
  205--220.

\bibitem{zhang2018robust}
S.~Zhang, R.~Ji, J.~Hu, Y.~Gao, and C.-W. Lin, ``Robust face sketch synthesis
  via generative adversarial fusion of priors and parametric sigmoid,'' in
  \emph{{International Joint Conference on Artificial Intelligence}}.\hskip 1em
  plus 0.5em minus 0.4em\relax IJCAI, 2018, pp. 1163--1169.

\bibitem{zhang2018markov}
M.~Zhang, N.~Wang, X.~Gao, and Y.~Li, ``Markov random neural fields for face
  sketch synthesis,'' in \emph{{International Joint Conference on Artificial
  Intelligence}}.\hskip 1em plus 0.5em minus 0.4em\relax IJCAI, 2018, pp.
  1142--1148.

\bibitem{wang2018high}
L.~Wang, V.~Sindagi, and V.~Patel, ``High-quality facial photo-sketch synthesis
  using multi-adversarial networks,'' in \emph{International conference on
  automatic face \& gesture recognition}.\hskip 1em plus 0.5em minus
  0.4em\relax IEEE, 2018, pp. 83--90.

\bibitem{zhang2018dual}
M.~Zhang, R.~Wang, X.~Gao, J.~Li, and D.~Tao, ``Dual-transfer face sketch-photo
  synthesis,'' \emph{{IEEE Transactions on image processing}}, vol.~28, no.~2,
  pp. 642--657, 2018.

\bibitem{kazemi2018facial}
H.~Kazemi, M.~Iranmanesh, A.~Dabouei, S.~Soleymani, and N.~M. Nasrabadi,
  ``Facial attributes guided deep sketch-to-photo synthesis,'' in \emph{{Winter
  Applications of Computer Vision Workshops}}.\hskip 1em plus 0.5em minus
  0.4em\relax IEEE, 2018, pp. 1--8.

\bibitem{kazemi2018unsupervised}
H.~Kazemi, F.~Taherkhani, and N.~M. Nasrabadi, ``Unsupervised facial geometry
  learning for sketch to photo synthesis,'' in \emph{{International Conference
  of the Biometrics Special Interest Group}}.\hskip 1em plus 0.5em minus
  0.4em\relax IEEE, 2018, pp. 1--5.

\bibitem{you2019pirec}
S.~You, N.~You, and M.~Pan, ``Pi-rec: Progressive image reconstruction network
  with edge and color domain,'' \emph{arXiv preprint arXiv:1903.10146}, 2019.

\bibitem{zhang2019deep}
M.~Zhang, N.~Wang, Y.~Li, and X.~Gao, ``Deep latent low-rank representation for
  face sketch synthesis,'' \emph{{IEEE Transactions on Neural Networks and
  Learning Systems}}, vol.~30, no.~10, pp. 3109--3123, 2019.

\bibitem{zhu2019deep}
M.~Zhu, J.~Li, N.~Wang, and X.~Gao, ``A deep collaborative framework for face
  photo-sketch synthesis,'' \emph{{IEEE Transactions on Neural Networks and
  Learning Systems}}, vol.~30, no.~10, pp. 3096--3108, 2019.

\bibitem{zhang2019cascaded}
M.~Zhang, Y.~Li, N.~Wang, Y.~Chi, and X.~Gao, ``Cascaded face sketch synthesis
  under various illuminations,'' \emph{{IEEE Transactions on image
  processing}}, vol.~29, pp. 1507--1521, 2019.

\bibitem{zhu2019face}
M.~Zhu, N.~Wang, X.~Gao, J.~Li, and Z.~Li, ``Face photo-sketch synthesis via
  knowledge transfer,'' in \emph{{International Joint Conference on Artificial
  Intelligence}}.\hskip 1em plus 0.5em minus 0.4em\relax IJCAI, 2019, pp.
  1048--1054.

\bibitem{li2019im2pencil}
Y.~Li, C.~Fang, A.~Hertzmann, E.~Shechtman, and M.-H. Yang, ``Im2pencil:
  Controllable pencil illustration from photographs,'' in \emph{{Conference on
  computer vision and pattern recognition}}.\hskip 1em plus 0.5em minus
  0.4em\relax IEEE, 2019, pp. 1525--1534.

\bibitem{ghosh2019interactive}
A.~Ghosh, R.~Zhang, P.~K. Dokania, O.~Wang, A.~A. Efros, P.~H. Torr, and
  E.~Shechtman, ``Interactive sketch \& fill: Multiclass sketch-to-image
  translation,'' in \emph{{International conference on computer vision}}.\hskip
  1em plus 0.5em minus 0.4em\relax IEEE, 2019, pp. 1171--1180.

\bibitem{wang2020learning}
X.~Wang and J.~Yu, ``Learning to cartoonize using white-box cartoon
  representations,'' in \emph{{Conference on computer vision and pattern
  recognition}}.\hskip 1em plus 0.5em minus 0.4em\relax IEEE, 2020, pp.
  8090--8099.

\bibitem{gao2020sketchycoco}
C.~Gao, Q.~Liu, Q.~Xu, L.~Wang, J.~Liu, and C.~Zou, ``{SketchyCOCO}: image
  generation from freehand scene sketches,'' in \emph{{Conference on computer
  vision and pattern recognition}}.\hskip 1em plus 0.5em minus 0.4em\relax
  IEEE, 2020, pp. 5174--5183.

\bibitem{yang2020deep}
S.~Yang, Z.~Wang, J.~Liu, and Z.~Guo, ``Deep plastic surgery: Robust and
  controllable image editing with human-drawn sketches,'' in \emph{{European
  conference on computer vision}}.\hskip 1em plus 0.5em minus 0.4em\relax
  Springer, 2020, pp. 601--617.

\bibitem{chen2020deepfacedrawing}
S.-Y. Chen, W.~Su, L.~Gao, S.~Xia, and H.~Fu, ``{DeepFaceDrawing}: Deep
  generation of face images from sketches,'' \emph{{ACM Transactions on
  graphics}}, vol.~39, no.~4, pp. 72--1, 2020.

\bibitem{yu2020toward}
J.~Yu, X.~Xu, F.~Gao, S.~Shi, M.~Wang, D.~Tao, and Q.~Huang, ``Toward realistic
  face photo-sketch synthesis via composition-aided gans,'' \emph{{IEEE
  Transactions on cybernetics}}, vol.~51, no.~9, pp. 4350--4362, 2020.

\bibitem{fang2020identity}
Y.~Fang, W.~Deng, J.~Du, and J.~Hu, ``Identity-aware {CycleGAN} for face
  photo-sketch synthesis and recognition,'' \emph{{Pattern Recognition}}, vol.
  102, p. 107249, 2020.

\bibitem{lin2020identity}
Y.~Lin, S.~Ling, K.~Fu, and P.~Cheng, ``An identity-preserved model for face
  sketch-photo synthesis,'' \emph{{IEEE Signal Processing Letters}}, vol.~27,
  pp. 1095--1099, 2020.

\bibitem{peng2020universal}
C.~Peng, N.~Wang, J.~Li, and X.~Gao, ``Universal face photo-sketch style
  transfer via multiview domain translation,'' \emph{{IEEE Transactions on
  image processing}}, vol.~29, pp. 8519--8534, 2020.

\bibitem{duan2020multi}
S.~Duan, Z.~Chen, Q.~J. Wu, L.~Cai, and D.~Lu, ``Multi-scale gradients
  self-attention residual learning for face photo-sketch transformation,''
  \emph{IEEE Transactions on Information Forensics and Security}, vol.~16, pp.
  1218--1230, 2020.

\bibitem{wang2021sketch}
S.-Y. Wang, D.~Bau, and J.-Y. Zhu, ``Sketch your own {GAN},'' in
  \emph{{International conference on computer vision}}.\hskip 1em plus 0.5em
  minus 0.4em\relax IEEE, 2021, pp. 14\,050--14\,060.

\bibitem{bhunia2021doodleformer}
A.~K. Bhunia, S.~Khan, H.~Cholakkal, R.~M. Anwer, F.~S. Khan, J.~Laaksonen, and
  M.~Felsberg, ``Doodleformer: Creative sketch drawing with transformers,''
  \emph{arXiv preprint arXiv:2112.03258}, 2021.

\bibitem{song2018spg}
Y.~Song, C.~Yang, Y.~Shen, P.~Wang, Q.~Huang, and C.-C.~J. Kuo, ``Spg-net:
  Segmentation prediction and guidance network for image inpainting,'' in
  \emph{{British Machine Vision Conference}}.\hskip 1em plus 0.5em minus
  0.4em\relax {BMVA} Press, 2018, p.~97.

\bibitem{yi2014learning}
D.~Yi, Z.~Lei, S.~Liao, and S.~Z. Li, ``Learning face representation from
  scratch,'' \emph{arXiv preprint arXiv:1411.7923}, 2014.

\bibitem{wang2014feature}
L.~Wang, R.-F. Li, K.~Wang, and J.~Chen, ``Feature representation for facial
  expression recognition based on facs and lbp,'' \emph{International Journal
  of Automation and Computing}, vol.~11, no.~5, pp. 459--468, 2014.

\bibitem{zheng2020survey}
X.~Zheng, Y.~Guo, H.~Huang, Y.~Li, and R.~He, ``A survey of deep facial
  attribute analysis,'' \emph{{International journal of computer vision}}, vol.
  128, no.~8, pp. 2002--2034, 2020.

\bibitem{huang2008labeled}
G.~B. Huang, M.~Mattar, T.~Berg, and E.~Learned-Miller, ``Labeled faces in the
  wild: A database forstudying face recognition in unconstrained
  environments,'' in \emph{Workshop on faces in'Real-Life'Images: detection,
  alignment, and recognition}, 2008.

\bibitem{ranjan2017hyperface}
R.~Ranjan, V.~M. Patel, and R.~Chellappa, ``Hyperface: A deep multi-task
  learning framework for face detection, landmark localization, pose
  estimation, and gender recognition,'' \emph{{IEEE Transactions on pattern
  analysis and machine intelligence}}, vol.~41, no.~1, pp. 121--135, 2017.

\bibitem{hand2017attributes}
E.~M. Hand and R.~Chellappa, ``Attributes for improved attributes: A multi-task
  network utilizing implicit and explicit relationships for facial attribute
  classification,'' in \emph{{AAAI Conference on Artificial
  Intelligence}}.\hskip 1em plus 0.5em minus 0.4em\relax AAAI Press, 2017, pp.
  4068--4074.

\bibitem{han2017heterogeneous}
H.~Han, A.~K. Jain, F.~Wang, S.~Shan, and X.~Chen, ``Heterogeneous face
  attribute estimation: A deep multi-task learning approach,'' \emph{{IEEE
  Transactions on pattern analysis and machine intelligence}}, vol.~40, no.~11,
  pp. 2597--2609, 2017.

\bibitem{jang2017smilenet}
Y.~Jang, H.~Gunes, and I.~Patras, ``Smilenet: registration-free smiling face
  detection in the wild,'' in \emph{{International conference on computer
  vision workshops}}.\hskip 1em plus 0.5em minus 0.4em\relax IEEE, 2017, pp.
  1581--1589.

\bibitem{ranjan2017all}
R.~Ranjan, S.~Sankaranarayanan, C.~D. Castillo, and R.~Chellappa, ``An
  all-in-one convolutional neural network for face analysis,'' in
  \emph{International conference on automatic face \& gesture
  recognition}.\hskip 1em plus 0.5em minus 0.4em\relax IEEE, 2017, pp. 17--24.

\bibitem{li2020deep}
S.~Li and W.~Deng, ``Deep facial expression recognition: A survey,''
  \emph{Transactions on affective computing}, 2020.

\bibitem{zhang2014panda}
N.~Zhang, M.~Paluri, M.~Ranzato, T.~Darrell, and L.~Bourdev, ``Panda: Pose
  aligned networks for deep attribute modeling,'' in \emph{{Conference on
  computer vision and pattern recognition}}.\hskip 1em plus 0.5em minus
  0.4em\relax IEEE, 2014, pp. 1637--1644.

\bibitem{kan2014stacked}
M.~Kan, S.~Shan, H.~Chang, and X.~Chen, ``Stacked progressive auto-encoders
  (spae) for face recognition across poses,'' in \emph{{Conference on computer
  vision and pattern recognition}}.\hskip 1em plus 0.5em minus 0.4em\relax
  IEEE, 2014, pp. 1883--1890.

\bibitem{wu2013facial}
Y.~Wu, Z.~Wang, and Q.~Ji, ``Facial feature tracking under varying facial
  expressions and face poses based on restricted boltzmann machines,'' in
  \emph{{Conference on computer vision and pattern recognition}}.\hskip 1em
  plus 0.5em minus 0.4em\relax IEEE, 2013, pp. 3452--3459.

\bibitem{tran2017disentangled}
L.~Tran, X.~Yin, and X.~Liu, ``Disentangled representation learning {GAN} for
  pose-invariant face recognition,'' in \emph{{Conference on computer vision
  and pattern recognition}}.\hskip 1em plus 0.5em minus 0.4em\relax IEEE, 2017,
  pp. 1415--1424.

\bibitem{toseeb2012significance}
U.~Toseeb, D.~R. Keeble, and E.~J. Bryant, ``The significance of hair for face
  recognition,'' \emph{PloS one}, vol.~7, no.~3, p. e34144, 2012.

\bibitem{bartel2018know}
S.~J. Bartel, K.~Toews, L.~Gronhovd, and S.~L. Prime, ``Do i know you? altering
  hairstyle affects facial recognition,'' \emph{Visual Cognition}, vol.~26,
  no.~3, pp. 149--155, 2018.

\bibitem{kumar2008facetracer}
N.~Kumar, P.~Belhumeur, and S.~Nayar, ``Facetracer: A search engine for large
  collections of images with faces,'' in \emph{{European conference on computer
  vision}}.\hskip 1em plus 0.5em minus 0.4em\relax Springer, 2008, pp.
  340--353.

\bibitem{huai2018facial}
L.~Huai-Yu, D.~Wei-Ming, and B.-G. Hu, ``Facial image attributes transformation
  via conditional recycle generative adversarial networks,'' \emph{Journal of
  Computer Science and Technology}, vol.~33, no.~3, pp. 511--521, 2018.

\bibitem{pierrard2007skin}
J.-S. Pierrard and T.~Vetter, ``Skin detail analysis for face recognition,'' in
  \emph{{Conference on computer vision and pattern recognition}}.\hskip 1em
  plus 0.5em minus 0.4em\relax IEEE, 2007, pp. 1--8.

\bibitem{li2009encyclopedia}
S.~Z. Li, \emph{Encyclopedia of Biometrics: I-Z.}\hskip 1em plus 0.5em minus
  0.4em\relax Springer Science \& Business Media, 2009, vol.~2.

\bibitem{zhang2016joint}
K.~Zhang, Z.~Zhang, Z.~Li, and Y.~Qiao, ``Joint face detection and alignment
  using multitask cascaded convolutional networks,'' \emph{{IEEE Signal
  Processing Letters}}, vol.~23, no.~10, pp. 1499--1503, 2016.

\bibitem{HeZRS16}
K.~He, X.~Zhang, S.~Ren, and J.~Sun, ``Deep residual learning for image
  recognition,'' in \emph{{Conference on computer vision and pattern
  recognition}}.\hskip 1em plus 0.5em minus 0.4em\relax IEEE, 2016, pp.
  770--778.

\bibitem{choi2018stargan}
Y.~Choi, M.~Choi, M.~Kim, J.-W. Ha, S.~Kim, and J.~Choo, ``{StarGAN}: Unified
  generative adversarial networks for multi-domain image-to-image
  translation,'' in \emph{{Conference on computer vision and pattern
  recognition}}.\hskip 1em plus 0.5em minus 0.4em\relax IEEE, 2018, pp.
  8789--8797.

\bibitem{zhao2018modular}
B.~Zhao, B.~Chang, Z.~Jie, and L.~Sigal, ``Modular generative adversarial
  networks,'' in \emph{{European conference on computer vision}}.\hskip 1em
  plus 0.5em minus 0.4em\relax Springer, 2018, pp. 150--165.

\bibitem{paszke2017automatic}
A.~Paszke, S.~Gross, S.~Chintala, G.~Chanan, E.~Yang, Z.~DeVito, Z.~Lin,
  A.~Desmaison, L.~Antiga, and A.~Lerer, ``Automatic differentiation in
  pytorch,'' in \emph{{Advances in neural information processing systems
  workshops}}.\hskip 1em plus 0.5em minus 0.4em\relax Curran Associates, Inc.,
  2017.

\bibitem{kingma2014adam}
D.~P. Kingma and J.~Ba, ``Adam: A method for stochastic optimization,'' in
  \emph{{International Conference on Learning Representations}}.\hskip 1em plus
  0.5em minus 0.4em\relax OpenReview.net, 2014.

\bibitem{yu2016sketch}
Q.~Yu, F.~Liu, Y.-Z. Song, T.~Xiang, T.~M. Hospedales, and C.-C. Loy, ``Sketch
  me that shoe,'' in \emph{{Conference on computer vision and pattern
  recognition}}.\hskip 1em plus 0.5em minus 0.4em\relax IEEE, 2016, pp.
  799--807.

\bibitem{shorten2019survey}
C.~Shorten and T.~M. Khoshgoftaar, ``A survey on image data augmentation for
  deep learning,'' \emph{Journal of big data}, vol.~6, no.~1, pp. 1--48, 2019.

\bibitem{wang2018transferring}
Y.~Wang, C.~Wu, L.~Herranz, J.~van~de Weijer, A.~Gonzalez-Garcia, and
  B.~Raducanu, ``Transferring gans: generating images from limited data,'' in
  \emph{ECCV}.\hskip 1em plus 0.5em minus 0.4em\relax Springer, 2018, pp.
  218--234.

\bibitem{yu2020deepi2i}
y.~wang, L.~Yu, and J.~van~de Weijer, ``Deepi2i: Enabling deep hierarchical
  image-to-image translation by transferring from gans,'' in \emph{{Advances in
  neural information processing systems}}.\hskip 1em plus 0.5em minus
  0.4em\relax Curran Associates, Inc., 2020, pp. 11\,803--11\,815.

\bibitem{shocher2020semantic}
A.~Shocher, Y.~Gandelsman, I.~Mosseri, M.~Yarom, M.~Irani, W.~T. Freeman, and
  T.~Dekel, ``Semantic pyramid for image generation,'' in \emph{{International
  conference on computer vision}}.\hskip 1em plus 0.5em minus 0.4em\relax IEEE,
  2020, pp. 7457--7466.

\bibitem{RaviL17}
S.~Ravi and H.~Larochelle, ``Optimization as a model for few-shot learning,''
  in \emph{{International Conference on Learning Representations}}.\hskip 1em
  plus 0.5em minus 0.4em\relax OpenReview.net, 2017.

\bibitem{chapelle2009semi}
O.~Chapelle, B.~Scholkopf, and A.~Zien, ``Semi-supervised learning (chapelle,
  o. et al., eds.; 2006)[book reviews],'' \emph{IEEE Transactions on Neural
  Networks}, vol.~20, no.~3, pp. 542--542, 2009.

\bibitem{OquabBLS15}
M.~Oquab, L.~Bottou, I.~Laptev, and J.~Sivic, ``Is object localization for
  free? - weakly-supervised learning with convolutional neural networks,'' in
  \emph{{Conference on computer vision and pattern recognition}}.\hskip 1em
  plus 0.5em minus 0.4em\relax IEEE, 2015, pp. 685--694.

\bibitem{wang2017transitive}
X.~Wang, K.~He, and A.~Gupta, ``Transitive invariance for self-supervised
  visual representation learning,'' in \emph{{International conference on
  computer vision}}.\hskip 1em plus 0.5em minus 0.4em\relax IEEE, 2017, pp.
  1329--1338.

\bibitem{PintoMT13}
R.~Pinto, T.~Mettler, and M.~Taisch, ``Managing supplier delivery reliability
  risk under limited information: Foundations for a human-in-the-loop {DSS},''
  \emph{Decision support systems}, vol.~54, no.~2, pp. 1076--1084, 2013.

\bibitem{lecun1989generalization}
Y.~LeCun \emph{et~al.}, ``Generalization and network design strategies,''
  \emph{Connectionism in perspective}, vol.~19, no. 143-155, p.~18, 1989.

\bibitem{tolstikhin2021mlp}
I.~O. Tolstikhin, N.~Houlsby, A.~Kolesnikov, L.~Beyer, X.~Zhai, T.~Unterthiner,
  J.~Yung, A.~Steiner, D.~Keysers, J.~Uszkoreit \emph{et~al.}, ``Mlp-mixer: An
  all-mlp architecture for vision,'' in \emph{{Advances in neural information
  processing systems}}.\hskip 1em plus 0.5em minus 0.4em\relax Curran
  Associates, Inc., 2021.

\bibitem{vaswani2017attention}
A.~Vaswani, N.~Shazeer, N.~Parmar, J.~Uszkoreit, L.~Jones, A.~N. Gomez,
  {\L}.~Kaiser, and I.~Polosukhin, ``Attention is all you need,'' in
  \emph{{Advances in neural information processing systems}}.\hskip 1em plus
  0.5em minus 0.4em\relax Curran Associates, Inc., 2017.

\bibitem{lee2021vitgan}
K.~Lee, H.~Chang, L.~Jiang, H.~Zhang, Z.~Tu, and C.~Liu, ``{ViTGAN}: Training
  {GAN}s with vision transformers,'' in \emph{{International Conference on
  Learning Representations}}.\hskip 1em plus 0.5em minus 0.4em\relax
  OpenReview.net, 2022.

\bibitem{zhang2011fsim}
L.~Zhang, L.~Zhang, X.~Mou, and D.~Zhang, ``Fsim: A feature similarity index
  for image quality assessment,'' \emph{{IEEE Transactions on image
  processing}}, vol.~20, no.~8, pp. 2378--2386, 2011.

\bibitem{AvidanS07}
S.~Avidan and A.~Shamir, ``Seam carving for content-aware image resizing,''
  \emph{{ACM Transactions on graphics}}, vol.~26, no.~3, p. 10–es, 2007.

\bibitem{dong2015image}
C.~Dong, C.~C. Loy, K.~He, and X.~Tang, ``Image super-resolution using deep
  convolutional networks,'' \emph{{IEEE Transactions on pattern analysis and
  machine intelligence}}, vol.~38, no.~2, pp. 295--307, 2015.

\bibitem{hu2020towards}
Y.~Hu, S.~Yang, W.~Yang, L.-Y. Duan, and J.~Liu, ``Towards coding for human and
  machine vision: A scalable image coding approach,'' in \emph{IEEE
  International Conference on Multimedia and Expo}.\hskip 1em plus 0.5em minus
  0.4em\relax IEEE, 2020, pp. 1--6.

\bibitem{wood2021fake}
E.~Wood, T.~Baltru{\v{s}}aitis, C.~Hewitt, S.~Dziadzio, T.~J. Cashman, and
  J.~Shotton, ``Fake it till you make it: Face analysis in the wild using
  synthetic data alone,'' in \emph{{International conference on computer
  vision}}.\hskip 1em plus 0.5em minus 0.4em\relax IEEE, 2021, pp. 3681--3691.

\end{thebibliography}

\end{document}